\documentclass[runningheads]{llncs}

\usepackage{graphicx}
\usepackage{jhoagg}
\usepackage{natbib}
\usepackage{amsmath}
\usepackage{amsfonts}
\usepackage[english]{babel}
\usepackage{comment}
\usepackage{authblk}
\usepackage{setspace}
\usepackage{xfrac}
\usepackage{subcaption}
\usepackage{algorithm}
\usepackage{algorithmic}
\usepackage{xcolor}
\usepackage{siunitx}
\usepackage{booktabs}
\usepackage{comment}

\DeclareMathOperator*{\argmin}{arg\,min}

\graphicspath{{./Figures/}}
\begin{document}

\title{Robust AI Driving Strategy for Autonomous Vehicles}
%
%
\author{Subramanya Nageshrao \and
Yousaf Rahman \and
Vladimir Ivanovic \and
Mrdjan Jankovic \and
Eric Tseng \and
Michael Hafner \and
Dimitar Filev}

\authorrunning{S. Nageshrao et al.}
%
\institute{Ford Motor Company, Research and Advanced Engineering, Dearborn MI 48124, USA \\
\email{\{snageshr,yrahman,vivanovi,mjankov1,htseng,mhafner2,dfilev\}@ford.com}}

\maketitle              
\abstract{There has been significant progress in sensing, perception, and localization for automated driving, However, due to the wide spectrum of traffic/road structure scenarios and the long tail distribution of human driver behavior,  it has remained an open challenge for an intelligent vehicle to always know how to make and execute the best decision on road given available sensing / perception / localization information. In this chapter, we talk about how artificial intelligence and more specifically, reinforcement learning, can take advantage of operational knowledge and safety reflex to make strategical and tactical decisions. We discuss some challenging problems related to the robustness of reinforcement learning solutions and their implications to the practical design of driving strategies for autonomous vehicles. We focus on automated driving on highway and the integration of reinforcement learning, vehicle motion control, and control barrier function, leading to a  robust AI driving strategy that can learn and adapt safely \footnote{*Selected portions reprinted, with permission, from \citep{nageshrao2019autonomous}, \copyright 2019 IEEE}.}

%
%
%
\section{Introduction}

Reinforcement learning (RL) is a key branch of Artificial Intelligence (AI) at the intersection of machine learning, decision making and control.  It is a method of learning from interaction with the environment and it is inspired by the human learning process. RL gained success in the last few years as the right approach to learn how to make decisions in tasks that are too complex to explain or to learn from supervised examples.  In 2015 DeepMind pioneered a new approach to RL, Deep Q-Network (DQN) that approximates the Q-function \citep{sutton1998reinforcement} with a deep neural network \citep{mnih2015human,mnih2016asynchronous,lillicrap2015continuous,silver2017mastering}. The key idea of the DQN algorithm is to store the agent's experiences in a replay buffer and then randomly sample and replay these experiences to provide diverse and decorrelated training data for learning the Q-function. The DQN concept was further extended to the other RL methods, e.g. Deep Deterministic Policy Gradient (DDPG) and Proximal Policy Optimization (PPO), resulting in a new subfield aptly termed as Deep Reinforcement Learning (DRL). Following the success of the DQN methodology, DRL has become one of the most used AI methods in the development of autonomous vehicles (AV).

\begin{figure*}[htbp]
\centering
\includegraphics[width=6.5cm]{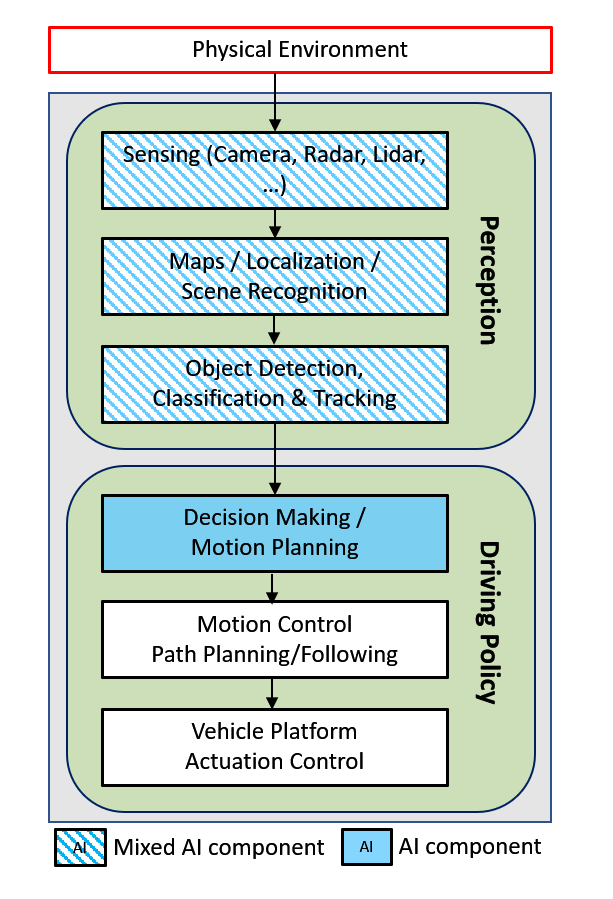}
\caption{AI for AV: A prototypical technology stack.}
\label{fig:AIforAV}
\end{figure*}

DRL can be an essential part of the prototypical AV technology stack (Sensing, Perception, Localization, Driving Policy and Actuation) as a method for decision making in the hierarchical driving policy workstream – Fig.~\ref{fig:AIforAV}.

The role of the decision making and motion planning module is to transform the information about the route and the state of the AV and surrounded traffic and environment into a high-level strategy -- modulation of the speed set-point, merging, lane changing, car following, aborting the current maneuver, etc. -- that is further implemented by the motion control system encompassing path planning, path following, and actuator control modules.  Early AV research and prototypes utilized motion planning concepts derived from classical control techniques – finite state machines, rules, heuristic strategies and Model Predictive Control (MPC) \citep{buehler20072005,buehler2009darpa,falcone2007linear,kim2014model,zhang2017finite}.  Following the massive introduction of AI methods in AV development in the last years, we can observe a considerable growth in the share of the DRL based methods for motion planning \citep{aradi2020survey}. The rationale behind this trend is the ability of the DRL to handle complex and ill-defined situations and environments that often occur in autonomous driving setting.  Presently, alternative versions and implementations of the DRL are at the core of the AV motion planning methods (e.g., detailed surveys of the DRL applications to AV can be found in \citep{aradi2020survey,ye2021survey}).

One of the challenging problems of the DRL based decision making and motion planning for AV is improving the robustness of the algorithms and preventing the possibility of unsafe actions and accidents.  The essence of this problem is the probabilistic nature of the RL concept, which does not generally imply that the maximization of the reward function can guarantee the safety and repeatability of the solution.

One common approach to improving DRL robustness, we call it the probabilistic robustness, is focused on enhancing the exploration strategy during the training of the DRL algorithms.  The final goal is enriching the training set by introducing, through adversarial disturbances and non-reward based exploration, unexpected scenarios and situations well beyond the space defined by the simulation model/distribution and the handcrafted reward function.  The methods of this group include model-based exploration \citep{pathak2017curiosity,vezzani2019learning,zhang2019discretionary}, adversarial perturbations of the state observations \citep{lin2017tactics}, infusing disturbances that are modulated by the capability of the control policy \citep{ma2018improved}, generating socially acceptable perturbations \citep{zhang2020generating}, maximization of the rewards associated with the high risk trajectories \citep{tamar2015optimizing}, etc. In general, the outcome of these techniques is improved robustness of the algorithms and consequentially reducing the level of risk, but without providing deterministic guaranties for solution.

An alternative methodology proposed by \citep{alshiekh2017safe} introduces the idea of constraining (\lq shielding\rq) the output of the RL algorithm within a safety envelope that is defined by a deterministic decision-making strategy.  Human crafted rules improving the safety of the DRL algorithm are applied in \citep{nageshrao2019autonomous, nageshrao2020vehicle} and \citep{mirchevska2018high}  in order to eliminate the unsafe actions produced by the DRL algorithm. On one side, the rules have the advantage of introducing human knowledge and experience in conjunction with the machine learning approach of developing the DRL. On the other side, the deterministic envelope defined by the rules might be too conservative and limited only to the specific situations considered by the designers.

In this chapter we are extending the `shielding' concept by combining the DRL decision making and motion planning algorithm with a generic deterministic algorithm that is aimed to define a safety boundary, called the ‘safety filter’ at the output of the DRL algorithm.  Our work started with a rule-based safety filter \citep{nageshrao2020vehicle} with simple motion actuation, described in Section~\ref{DRL+RB}, continued with representative vehicle motion control (Section~\ref{sec:Motion Control}), and extended to safety filter for wide-ranging situations (Section~\ref{Rahman2021CSD} and Section~\ref{SEC:RL_CBF}). The wide-ranging safety filter design is inspired by the recent progress in the theory of Control Barrier Functions (CBF) as a generic and reliable methodology for object avoidance in robotic and automotive applications.

\section{Decision Making: DRL Driving Strategy for Changing Lanes} \label{DRL+RB}


To ensure automated driving is capable of operating in diverse environment including varying traffic density, different driving style and norms, we develop a novel Deep Reinforcement Learning framework. In this section we first provide a basic introduction to RL and Q-learning. Following this we elaborate on our hybrid approach to solve decision-making and control problems. We introduce essential modifications to classical Q-learning and show the necessity in incorporating basic safety rules. We demonstrate the effectiveness of our approach using a comparative simulation study for the highway driving problem. Our novel hybrid architecture can effectively handle complex and possibly ill-defined situations and environments.

\subsection{Reinforcement learning and deep reinforcement learning an introduction}
In this section we will provide a brief theoretical background on decision making  using deep reinforcement learning. Additionally, we elaborate on the main motivation for using the hybrid control architecture presented in this chapter.

\subsubsection{Reinforcement learning}
In reinforcement learning (RL), an Agent learns an optimal policy for a given cost function by directly interacting with the environment. Broadly RL constitutes a set of algorithms that efficiently solve the sequential decision making problem for an underlying Markov decision process (MDP) \citep{sutton1998reinforcement}. An MDP is formally defined by a tuple $\langle \mathcal{S}, \mathcal{A}, \mathcal{T}, \mathcal{R} \rangle$. Where $\mathcal{S}\in \mathbb{R}^n$ is the state-space, $\mathcal{A}$ is the action space, $\mathcal{T}:\mathcal{S}\times \mathcal{A}\rightarrow \mathcal{S}$ is the state transformation function, and $\mathcal{R}:\mathcal{S}\times \mathcal{A} \times \mathcal{S} \rightarrow \mathbb{R}$ is the reward function.

At each discrete time step $t$ the learning Agent selects an action $a_t$ according to some policy $\pi$ for a given system state $s_t$ , i.e., $a_t = \pi(s_t)\, \in \mathcal{A}$, where $\mathcal{A}$ is a set of feasible actions. On applying this action the system transitions to a new state $s_{t+1} \in \mathcal{S}$,  and provides a scalar reward $r_{t+1}$. This process is repeated till the policy converges. It must be noted, both state transition $\mathcal{T}$ and policy $\pi$ can be stochastic.

The goal of an RL algorithm is to learn a policy $\pi$ so as to maximize the total cumulative reward termed as return $R_\mathrm{t} = \sum_{k=0}^{\infty} \gamma^k r_{t+k}$ where the scalar constant $\gamma \in \left( 0,1 \right]$ is the discount factor. For discrete action space the optimal policy is obtained by solving for the optimal $Q$ function, termed as the action-value function. An optimal $Q$ function satisfies the Bellman equation:
\begin{equation}\label{Sec2Eq1:BellmanQ}
Q^{\pi_\mathrm{opt}}(s_t,a_t) =  Q^{*}(s_t,a_t) = \mathbb{E}_{s_{t+1}} \left[ r_{t}+\gamma \max_{a_{t+1}} Q^{*}(s_{t+1},a_{t+1}) | (s_t,a_t) \right].
\end{equation}

For a given state $s_t$ any action $a_t$ that satisfies \eqref{Sec2Eq1:BellmanQ}  is the optimal action.

\subsubsection{Decision making and low-level actuation}
Inspired by an example from \citep{sutton1998reinforcement} (see Chapter 1.2), we elaborate on the difference between decision making ($\mathcal{D}$'s) and low level actuation ($\mathcal{C}$'s) :
\begin{itemize}
    \item Phil wants to have breakfast, he may chose between having cereal or a bagel ($\mathcal{D}_1$), after deciding he will either walk to cupboard or to the counter ($\mathcal{C}_1$).
    \item He will then either pick a choice of cereal ($\mathcal{D}_2$) and then walk to the fridge ($\mathcal{C}_2$) to get milk or chose a bagel ($\mathcal{D}_3$) and walk to the toaster ($\mathcal{C}_3$), etc.
\end{itemize}
This entire process involves a series of decision making followed by low-level actuation. High-level decisions (i.e., $\mathcal{D}$'s) are decided by reward function such as pleasure or getting enough nutrition etc. 

Any RL methodology that tries to solve both decision making and actuation simultaneously may require a large amount of training data. In this work we simplify this process by having a clear hierarchy between high-level decision making and low level actuation. RL is used to solve the high-level decision making problem, while the classical feedback control methods are used for low-level actuation.  Our approach has considerable overlap with hierarchical RL, where both decision making and control are learned simultaneously \citep{frans2017meta}. To learn optimal high-level decisions we use a modified Double DQN (DDQN) algorithm from \citep{van2016deep}.

\subsubsection{Need for robustness \& safety}\label{sec::NeedForSafety}
The trade-off between exploration and exploitation is a key characteristic that distinguish RL from other forms of machine learning algorithms \citep{sutton1998reinforcement}. In an unknown situation an agent needs to be curious and explore new action. Hence during the initial learning phase the Agent will invariably explore all viable actions as the entire environment is unknown. Unfortunately this curiosity can be fatal and potentially expensive \citep{garcia2012safe}. For example, during training, using unguided exploration could potentially lead to frequent collisions resulting in slow the training process. Additionally in the inference stage, due to perrception noise, function approximation etc. a trained agent may still potentially recommend a non-safe maneuver. In order to address these issues, we augment the DDQN decision maker with an explicit \emph{short-horizon safety check} that is used both during training as well as in the inference phase. Safe exploration or safe-RL is an active research topic, for a detailed survey see \citep{garcia2015comprehensive}.

\begin{figure}[!tbp]
  \begin{subfigure}[b]{0.5\textwidth}
    \includegraphics[width=\textwidth]{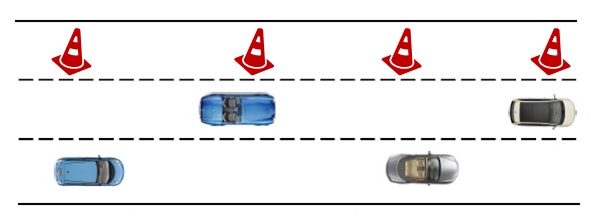}
    \caption{Highway with barricaded right lane.}
    \label{fig:BaricadedHighway}
  \end{subfigure}
  \hfill
  \begin{subfigure}[b]{0.5\textwidth}
    \includegraphics[width=\textwidth]{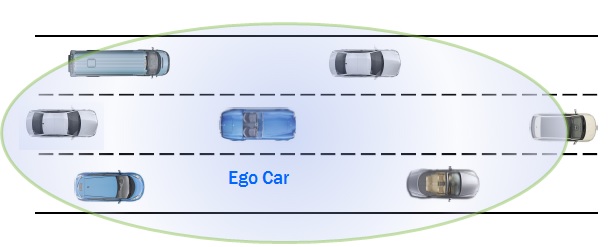}
    \caption{A three lane highway scenario.}
    \label{fig:EgoCar}
  \end{subfigure}
  \caption{An example for common sense road rule along with the ego vehicle perspective considered in this work (\copyright 2019 IEEE).
  }
\end{figure}

A standard DRL approach such as DDQN, may require a lot of samples before learning that certain action to be potentially dangerous in certain states. For example, consider a highway where one of the lane has been barricaded, see Fig.~\ref{fig:BaricadedHighway}. A standard epsilon-greedy algorithm may need to collide multiple times before learning that in certain situations going to the left-most lane, for e.g. when it is blocked, to be catastrophic. This may result in significant waste of learning effort as the Agent will spend considerable amount of time exploring irrelevant regions of the state and action space.  Additionally due to function approximation there is a small probability where the trained DDQN Agent may still chose unsafe actions leading to a catastrophic outcome. This can be avoided by including an explicit short-horizon safety check that evaluates the action choice by the learning Agent and provides an alternative safe action whenever it is feasible \citep{alshiekh2017safe}.

Our first solution to this problem called the Rule Based Safety Filter \citep{nageshrao2019autonomous} (see Section~\ref{sec:DQNtraining}) uses a simple safety check based on the common sense road rules. This forces the Agent to avoid non-safe actions in dangerous situations resulting in a faster training. Our approach is similar to a teacher who provides corrective action when it is necessary \citep{thomaz2008teachable}. Note that the safety filter may not be optimal, i.e., an expert in the task under consideration. Additionally, due to explicit safety check, new data can be obtained even in the inference phase which can then be used for continuous adaption of the learned network, this is further elaborated in Section~\ref{sec:continuous_adaptation}.

\subsection{DRL for Autonomous Driving}\label{sec:DQNtraining}

\begin{figure*}[h!]
\centering
\includegraphics[width=\textwidth, height=3cm]{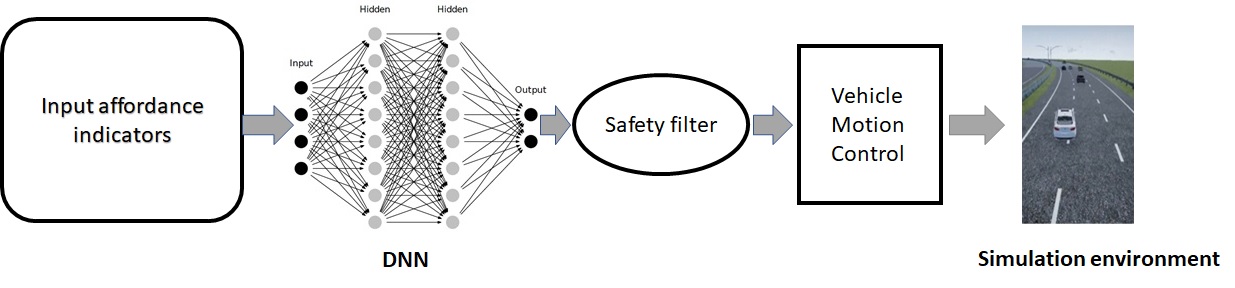}
\caption{DRL Agent control architecture.}
\label{fig:ControlDecisionMaking}
\end{figure*}

The DRL architecture used in this work is given in Fig.~\ref{fig:ControlDecisionMaking}.

Unlike the mediated perception method that relies on complete reconstruction of scene prior \citep{xu2017end,hecker2018end}, we use the concept of affordance indicators, i.e., state variables based on the direct perception approach from \citep{chen2015deepdriving}. For a three lane highway scenario, the ego vehicle (EV) can be surrounded by up to six traffic vehicles (TV), see Fig.~\ref{fig:EgoCar}.
\subsubsection{Affordance Indicators}
For Autonomous driving, the Agent uses information about its location within the road coordinate system and surrounding vehicles to select an action. This information is collectively termed as affordance indicators. The action taken by the Agent is then rewarded or penalized based on transitioned state. Cumulative reward is used by the Agent to learn safe and effective driving behavior. In this work, we use affordance indicators from 6 target vehicles in the vicinity of the ego vehicle. The 6 target vehicles are designated as Front Left, Front Center, Front Right, Rear Left, Rear Center and Rear Right.

To define the traffic vehicle's variable we use the following notation
\begin{equation}
    {\huge \mathrm{ State}}_{\huge \mathrm{Lane}_{\huge \mathrm{Location} \& \mathrm{Direction}}}
\end{equation}
where State can be distance $d$ or velocity $v$, Lane is either right $r$, center $c$, or left lane $l$. Location is either front $f$ or rear $r$ and Direction is either longitudinal $x$ or lateral $y$.

In total the following $24$ indicators are used to represent the spatio-temporal information of the six nearest traffic vehicles (see Fig.~\ref{fig:EgoCar}), they are formulated from the ego vehicle's perspective.
Here the lane occupancy of a TV and the closest car to EV are obtained using the methodology presented in \citep{zhang2017finite}. In addition to the 24 traffic vehicle states we use longitudinal velocity $v_\mathrm{e_x}$, lateral position $d_\mathrm{e_y}$, and lateral velocity $v_\mathrm{e_y}$ of the ego vehicle $e$. Since the affordance indicators are formulated w.r.t. ego vehicle we do not need a state corresponding to longitudinal position of the ego vehicle. These variables are minimal requirement for highway driving, however they are not sufficient for all highway driving tasks such as use of restricted lane, on-ramp to enter the highway, off-ramp to exit, etc.

\subsubsection{High Level Decision and Feedback Control}
A total of 27 affordance indicators, i.e., $s_t \in \mathbb{R}^{27}$ is used as an input to the deep $Q$-network. The $Q$-network is trained by using a modified double deep Q-learning algorithm from \citep{van2016deep}, see Algorithm~\ref{alg:DRL4AV}. For highway driving decision making, we consider four longitudinal action choices:
\begin{enumerate}
    \item \emph{maintain}
    \item \emph{accelerate}
    \item \emph{brake}
    \item \emph{hard brake}
\end{enumerate}
and three lateral action choices:
\begin{enumerate}
    \item \emph{keep lane}
    \item \emph{change lane to right}
    \item \emph{change lane to left}.
\end{enumerate}
The combination results in set of 12 unique actions. For each of these action choice a numerical value can be assigned \citep{li2017game}, for example for longitudinal acceleration four different discrete choices  $\{a_1,0,-a_1, -a_2 \}$ may be considered. Alternatively, one can obtain the reference for the throttle or brake controller either using the intelligent driver model (IDM) \citep{treiber2000congested} or adaptive cruise control \citep{vahidi2003research,rajamani2011vehicle}. The reference for the steering controller is self-evident, it is either stay in lane or change lane to right or left. To obtain the front wheel angle we use a simple feedback controller (see Section~\ref{sec:Lateral Motion Control} for details)
\begin{equation}\label{eq:SteerCmd}
 \kappa_\mathrm{cmd} = f(k_\mathrm{road},e_\mathrm{y_{off}},e_\mathrm{\psi},T_\mathrm{LC},v_\mathrm{e_x})
\end{equation}
where $k_\mathrm{road}$ is the road curvature, $e_\mathrm{\psi}$ is the heading angle offset, $T_\mathrm{LC}$ is the desired time to complete a  lane change, and the reference $e_\mathrm{y_{off}}$ is the lateral offset to the desired position. When the DDQN Agent decides to perform a lane change, the absolute value of the lane offset $e_\mathrm{y_{off}}$ will be set to the lane width. The inputs to the steering feedback controller \eqref{eq:SteerCmd} can be considered as additional affordance indicators which can be obtained from the perception module.

\begin{algorithm}[ht]
	\begin{algorithmic}[1]
		\STATE Initialize: $\mathrm{Buf}_\mathrm{S}$,  $\mathrm{Buf}_\mathrm{C}$, $Q(\theta)$  and target network $\hat{Q}(\hat{\theta})$ with $\hat{\theta}=\theta$ \\		
        \FOR{episode =1,$\cdots$,$N_\mathrm{e}$,}
		\STATE Initialize: $\{1,\cdots ,N_\mathrm{T}\}$ cars randomly, obtain affordance indicator $s_0$\\
		\FOR{samples $t$ =1,$\cdots$,$N_\mathrm{S}$, or Collision, }
    		\STATE With $\epsilon$ select random action $a_t$, else $a_t = \arg\max_\mathrm{a} Q\left((s_t,a,\theta_t) \right)$\\
            \STATE For ego car: \textbf{If} $a_t$ is not safe \textbf{Then} store $\left(s_t,a_t,*,r_\mathrm{col}\right)$ in $\mathrm{Buf}_\mathrm{C}$ and replace $a_t$ by safe action $a_\mathrm{s}$ \\
    		\STATE Apply action, observe $s_{t+1}$ and obtain $r_{t+1} = \rho(s_t, s_{t+1}, a_{t})$ \\
            \IF{Collision}
                \STATE Store transition $\left(s_t,a_t,*,r_\mathrm{col}\right)$ in collision buffer $\mathrm{Buf}_\mathrm{C}$ \\
            \ELSE
                \STATE Store transition $\left(s_t,a_t,s_{t+1},r_{t+1}\right)$ in safe buffer $\mathrm{Buf}_\mathrm{S}$ \\
            \ENDIF
            \STATE Sample random minibatch $\left(s_j,a_j,s_{j+1},r_{j+1}\right)$ from $\mathrm{Buf}_\mathrm{S}$ and $\mathrm{Buf}_\mathrm{C}$
            \STATE Set \begin{equation*}
                         y_j =
                                  \begin{cases}
                                    r_{j+1}       \quad \text{if sample is from } \mathrm{Buf}_\mathrm{C}\\
                                    r_{j+1}+\gamma \hat{Q} \Big( s_{j+1},  \arg\max_a Q\left(s_{j+1},a,\theta_t\right),\hat{\theta}_t \Big) \quad \text{if sample is from } \mathrm{Buf}_\mathrm{S}
                                  \end{cases}
                        \end{equation*}
            \STATE Perform gradient descent on $\|y_j-Q(s_j,a_j,\theta_t)\|_2$ w.r.t. $\theta$
            \STATE Every $N_\mathrm{C}$ episodes set $\hat{Q} = Q$
    		\ENDFOR
		\ENDFOR
	\end{algorithmic}
	\caption{A DRL based safe decision maker for autonomous highway driving}
	\label{alg:DRL4AV}
\end{algorithm}
\subsubsection{Rule Based Safety Filter}
As elaborated in Section~\ref{sec::NeedForSafety}, we use an explicit short-horizon safety check to validate the action choice by DDQN. For the current action choice, the safety filter verifies common sense but well known rules  of the road such as ensuring a minimum relative gap to a TV based on relative velocity
\begin{equation}\label{eq:sfConstraint}
    d_\mathrm{TV}-T_\mathrm{min}\times v_\mathrm{TV}>d_\mathrm{TV_{min}}
\end{equation}
where $d_\mathrm{TV}$, $v_\mathrm{TV}$ are the relative distance and velocity to a given traffic vehicle, $T_\mathrm{min}$ is the minimum time to collision, $d_\mathrm{TV_{min}}$ is the minimum gap which must be ensured before executing the action choice by the Agent. If this condition is not satisfied and when feasible, an alternate safe action will be provided by the short-horizon safety filter. In this example we use a simple variant of the intelligent driver model (IDM) \citep{treiber2000congested} to provide safe alternative longitudinal action, it is formulated as
\begin{equation}\label{eq:sfCtrl}
a_\mathrm{s} =
  \begin{cases}
    \text{Hard brake}       & \quad \text{if } T_\mathrm{C} \leq T_\mathrm{HB}\\
    \text{Brake}   & \quad \text{if } T_\mathrm{HB}<T_\mathrm{C} \leq T_\mathrm{B} \\
    \text{Maintain}   & \quad \text{if } T_\mathrm{B}<T_\mathrm{C}
  \end{cases}
\end{equation}
where $a_\mathrm{s}$ is the safe action, $T_\mathrm{C}$ is the calculated time to collision. It is defined as
\begin{equation}
    T_\mathrm{C} = \frac{d_\mathrm{TV}}{v_\mathrm{TV}}
\end{equation}
\citep{minderhoud2001extended}, $T_\mathrm{HB}$ and $T_\mathrm{B}$ are the thresholds above which the decision made by the DDQN Agent is considered to be safe.

In particular we use the following safety check prior to performing an action given by DDQN:
\begin{enumerate}
\item Instead of the in-lane longitudinal action by DDQN, chose a safe action using \eqref{eq:sfCtrl} if \eqref{eq:sfConstraint} is not satisfied and the ego vehicle is faster than the preceding vehicle.
\item If the ego vehicle is in left most lane then \emph{change lane to left} is not valid, similarly for the right lane.
\item For \emph{change lane to left} continuously monitor \eqref{eq:sfConstraint} for center-front car and to the left-front and the left-rear car in the target lane. If condition \eqref{eq:sfConstraint} fails then  lane change is either not initiated or aborted, similarly for the \emph{change lane to right}.
\end{enumerate}

After training, generally the trained Agent is frozen and used only in inference mode. However, in reality the Agent may encounter new information, additionally there can be a considerable variation between the training environment and the real-world experience. Also due of function approximation there can be a small probability of choosing an unsafe action even by the trained Agent, this can happen even after convergence and in the absence of any explicit exploration. In order to address these issues, in the implementation phase we augment the trained DDQN Agent with the short-horizon safety check that was used during learning. Any new safety violation data will be added to the collision buffer $\mathrm{Buf}_\mathrm{C}$,  by using the training part of the Algorithm~\ref{alg:DRL4AV} (line 13 to 15) the learned Agent can be re-trained or adapted in a continuous manner. In the following section we apply the developed DRL based decision making Algorithm~\ref{alg:DRL4AV} for autonomous highway driving.

\subsection{Vehicle Dynamics}
Throughout this chapter, we use different vehicle dynamics models based on the modeling requirements. In this section, we present the 3 models that are used in subsequent sections.
\subsubsection{Point Mass Model}
When using this model, each vehicle is modeled as a computationally efficient point-mass in discrete time. For longitudinal equations of motion we use a discrete-time double integrator, and for lateral motion we use a simple kinematic model.
\begin{align}
x(t+1) &= x(t)+v_{x}(t) \Delta t, \nonumber \\
y(t+1) &= y(t)+v_\mathrm{y}(t) \Delta t, \label{model:pm} \\
v_{x}(t+1) &= v_{x}(t) + a_\mathrm{x}(t) \Delta t, \notag
\end{align}
where $t$ is the time index, $\Delta t$ is the sampling time, $x\in \mathbb{R}$ is the longitudinal position, $y \in \mathbb{R}$ is the lateral position of the car, and $v_{x} \in \mathbb{R}$ is the longitudinal velocity of the vehicle. In (\ref{model:pm}), the external control inputs $a_\mathrm{x}(t)$ and $v_\mathrm{y}(t)$ represents the longitudinal acceleration and lateral velocity of the vehicle, respectively.

We assume $a_\mathrm{x}(t)$ to vary from nominal acceleration,  to hard brake and is discretized into four values, i.e., $a_\mathrm{x} = \{a_1,0,-a_1, -a_2 \}$, with $a_1 = 2\, m/s^2$ and $a_2 = 4\, m/s^2$. Only in case of emergency hard braking of $a_\mathrm{x}=-a_2$ is applied. The lateral velocity $v_\mathrm{y}(t)$ provides a reference lane for the vehicle, we assume 5 seconds to complete a lane change action \citep{toledo2007modeling}, with an option to abort at any sampling instance. In this work we use $1 \,Hz$ sampling for the driving policy.

\subsubsection{Dynamic Bicycle Model}
In subsequent sections, we also use a continuous dynamic bicycle model for vehicle dynamics \citep{rajamani2011vehicle}. This model is required when simulating vehicle dynamics at higher frequencies.
\begin{equation}
\begin{aligned}
    \dot{x} &= v \: \textrm{cos}(\phi + \beta),  \\
    \dot{y} &= v \: \textrm{sin}(\phi + \beta),  \\
    \dot{v}_{\rml\rmo\rmn} &= g\alpha - F_{\textrm{aero}} - mg \:\textrm{sin}(\theta_{r}) \\
    \dot{v}_{\rml\rma\rmt} &= \frac{F_\rmf + F_\rmr}{m} -v_{\rml\rmo\rmn}\dot{\phi}\\
    \ddot{\phi} &= \frac{F_\rmf L_\rmf - F_\rmr L_\rmr}{I_\rmz},
\end{aligned}
\label{eq:fulldyn}
\end{equation}
where $v$ is the velocity, $\phi$ is the heading angle, $\beta$ is the slip angle, $L_\rmf$ is the distance of the front wheel to the center of mass, $L_\rmr$ is the distance of the rear wheel to the center of mass, $\delta$ is the front wheel angle, $v_{\textrm{lon}}$ is the longitudinal velocity, $v_{\textrm{lat}}$ is the lateral velocity, $g$ is the acceleration due to gravity, $\alpha$ is the longitudinal acceleration request in $g$'s, $F_{\textrm{aero}}$ is the aerodynamic drag, $m$ is the mass of the vehicle, $\theta_r$ is the road grade, $F_\rmf$ is the lateral tire force on the front tire, $F_\rmf$ is the lateral tire force on the rear tire, and $I_\rmz$ is the yaw moment of inertia of vehicle. The lateral tire forces are calculated as follows
\begin{equation}
\begin{aligned}
    F_\rmf &= 2C_\rmf(\delta - \theta_{v\rmf}), \\
    F_\rmr &= 2C_\rmr(-\theta_{v\rmr}), \\
    \theta_{v\rmf} &= \arctan\Big(\frac{v_{\textrm{lat}}+L_\rmf\dot{\phi}}{v_{\textrm{lon}}}\Big), \\
    \theta_{v\rmr} &= \arctan\Big(\frac{v_{\textrm{lat}}-L_\rmr\dot{\phi}}{v_{\textrm{lon}}}\Big),
\end{aligned}
\label{eq:tire_forces}
\end{equation}
where $C_\rmf$ is the cornering stiffness of each front tire, $C_\rmr$ is the cornering stiffness of each rear tire, $\theta_{v\rmf}$ is the front tire velocity angle and $\theta_{v\rmr}$ is the rear tire velocity angle. In all simulations in Section \ref{Rahman2021CSD}, this model is used.
\subsubsection{Simplified Bicycle Model}
To aid in calculating the barrier dynamics in Sections \ref{Rahman2021CSD} and \ref{SEC:RL_CBF}, we use the simplified decoupled system dynamics shown below.
\begin{equation}
\begin{aligned}
\dot{x}_\rmT &= v_\rmT \, \textrm{cos}(\phi_\rmT) - v_\rmH\, \textrm{cos}(\phi_\rmH), \\
\dot{v}_\rmH &= g\alpha \\
\dot{y}_\rmT &= v_\rmT\, \textrm{sin}(\phi_\rmT) - v_\rmH\, \textrm{sin}(\phi_\rmH), \\
\dot{\phi}_\rmH &= \frac{v_\rmH}{L_\rmH}\delta,
\end{aligned}
\label{eq:simpdyn}
\end{equation}
where $x_\rmT$ and $y_\rmT$ are the relative $x$ and $y$ positions of the target center with respect to the ego center in road coordinates, $\phi_\rmT$ and $\phi_\rmH$ are the heading angles of the target and ego vehicles w.r.t. the road coordinate system, $v_\rmT$ and $v_\rmH$ are the absolute velocities of the target and ego respectively, $L_\rmH$ is the wheelbase of the ego vehicle, and $\delta$ is the front wheel angle of the ego vehicle.
\subsection{Simulation Results}
In this section we will show the applicability of our DRL based decision making Algorithm~\ref{alg:DRL4AV} for autonomous highway driving. First, we will elaborate on the training environment and evaluate the learned policy.
\subsubsection{Training environment}

\begin{figure}[htbp]
\centering
\includegraphics[width=12cm]{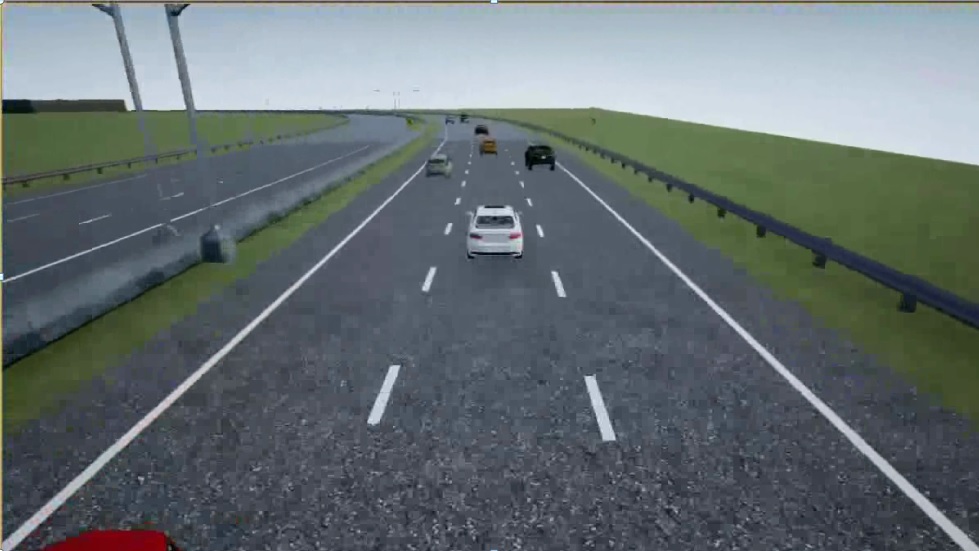}
\caption{A schematic of the simulation environment used for training.}
\label{fig:SimulationEnv}
\end{figure}

A schematic of the simulation environment used for training is given in Fig.~\ref{fig:SimulationEnv}. It is a three lane circular loop and is used to approximate an infinite stretch of straight highway. At the beginning of an episode, anywhere between $\{1,\cdots,N_\mathrm{T}\}$ number of cars are placed randomly within a distance of $250\, m$ from the ego car. In this example we chose $N_\mathrm{T}$ to be 30.

During learning stage the ego car (for e.g., white Fusion in Fig.~\ref{fig:SimulationEnv}) uses an $\epsilon$-greedy RL policy to make decisions, whereas for the traffic vehicles a combination of controllers from \citep{li2017game} and \citep{zhang2017finite} are used along with an IDM controller \citep{treiber2000congested}. Additionally the traffic vehicles can randomly chose to perform lane change. For the traffic vehicles, the system parameters such as maximum velocity are randomly chosen. This is to ensure a diverse traffic scenario in training and evaluation. We assume that all the traffic vehicles take into account the relative distance and velocity to preceding vehicle before making a decision, i.e., they will not rear end the preceding car in the same lane. We use Algorithm~\ref{alg:DRL4AV} to train an Agent for decision making.

\subsubsection{Reward components for DRL training}
In order to train the policy $\pi$ we use a reward function $\rho$ that consists of a set driving goals for the ego car. It is formulated as a function of
\begin{itemize}
\item Desired traveling speed subject to traffic condition \eqref{eq:rewv},
\item Desired lane and lane offset subject to traffic condition \eqref{eq:rewy},
\item Relative distance to the preceding car based on relative velocity \eqref{eq:rewx},
\end{itemize}

\begin{align}
r_v &= e^{-\frac{\left(v_\mathrm{e_x}-v_\mathrm{des}\right)^2}{10}}  -1, \label{eq:rewv} \\
r_y &= e^{-\frac{\left(d_\mathrm{e_y}-y_\mathrm{des}\right)^2}{10}}-1, \label{eq:rewy} \\
r_x &= \begin{cases}
    e^{-\frac{\left(d_\mathrm{lead}-d_\mathrm{safe}\right)^2}{10 d_\mathrm{safe}}}-1       & \quad \text{if } e_x < d_\mathrm{safe}\\
    0  & \quad \text{otherwise,}
  \end{cases} \label{eq:rewx}
\end{align}
where $v_\mathrm{e_x}$, $d_\mathrm{e_y}$, and $d_\mathrm{lead}$ are the ego velocity, lateral position, and the longitudinal distance to the lead vehicle respectively. Similarly, $v_\mathrm{des}$, $y_\mathrm{des}$, and $d_\mathrm{safe}$ are the desired speed, lane position, and safe longitudinal distance to the lead vehicle respectively.

\begin{figure}[h!]
\centering	
  \begin{subfigure}[htbp]{0.48\textwidth}
    \includegraphics[width=\textwidth]{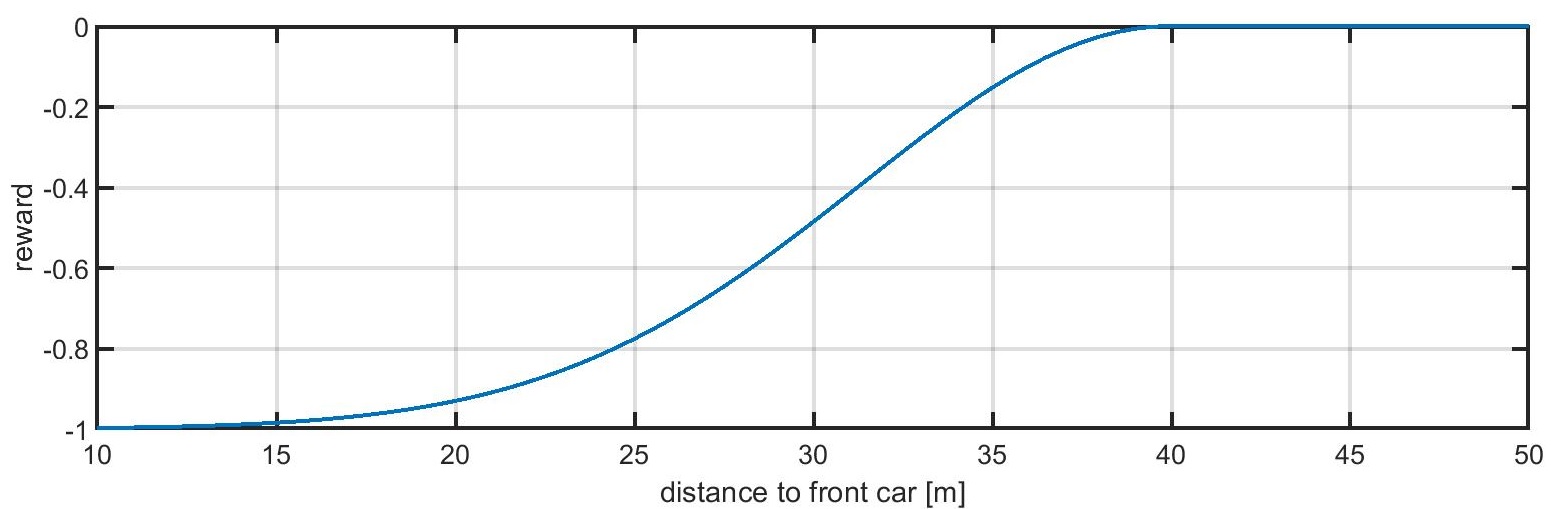}
    \caption{Reward for desired relative distance.}
    \label{fig:rewD}
  \end{subfigure}
  \quad
  \begin{subfigure}[htbp]{0.48\textwidth}
    \includegraphics[width=\textwidth]{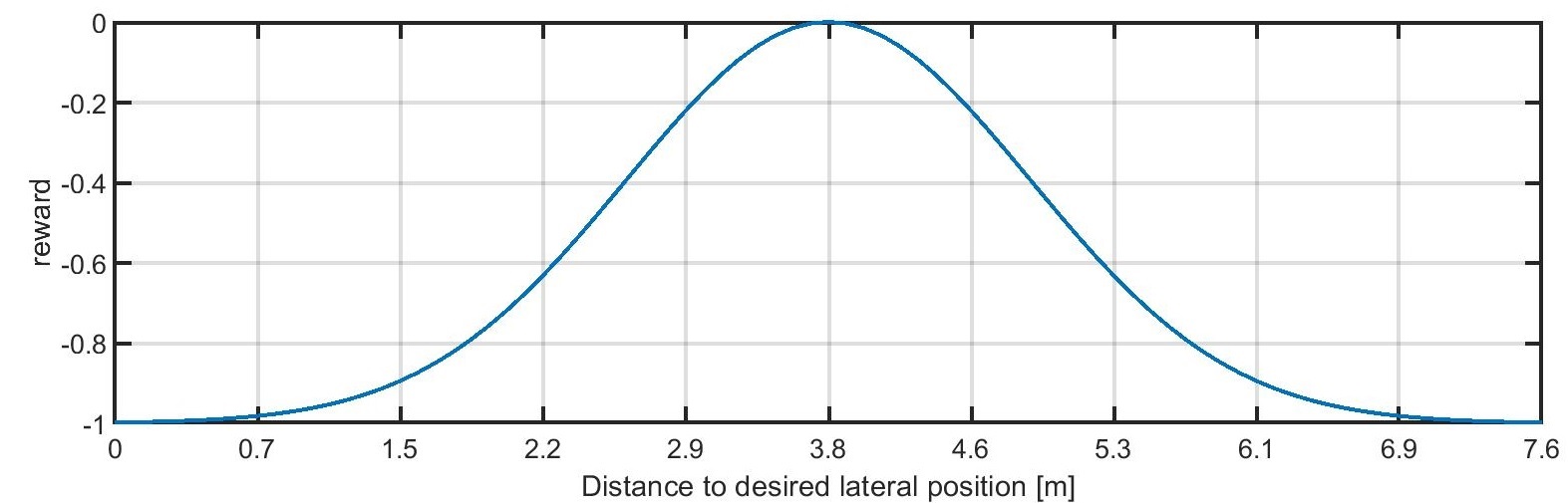}
    \caption{Reward for desired lateral position.}
    \label{fig:rewY}
  \end{subfigure}

  \begin{subfigure}[htbp]{0.48\textwidth}
    \includegraphics[width=\textwidth,height=3cm]{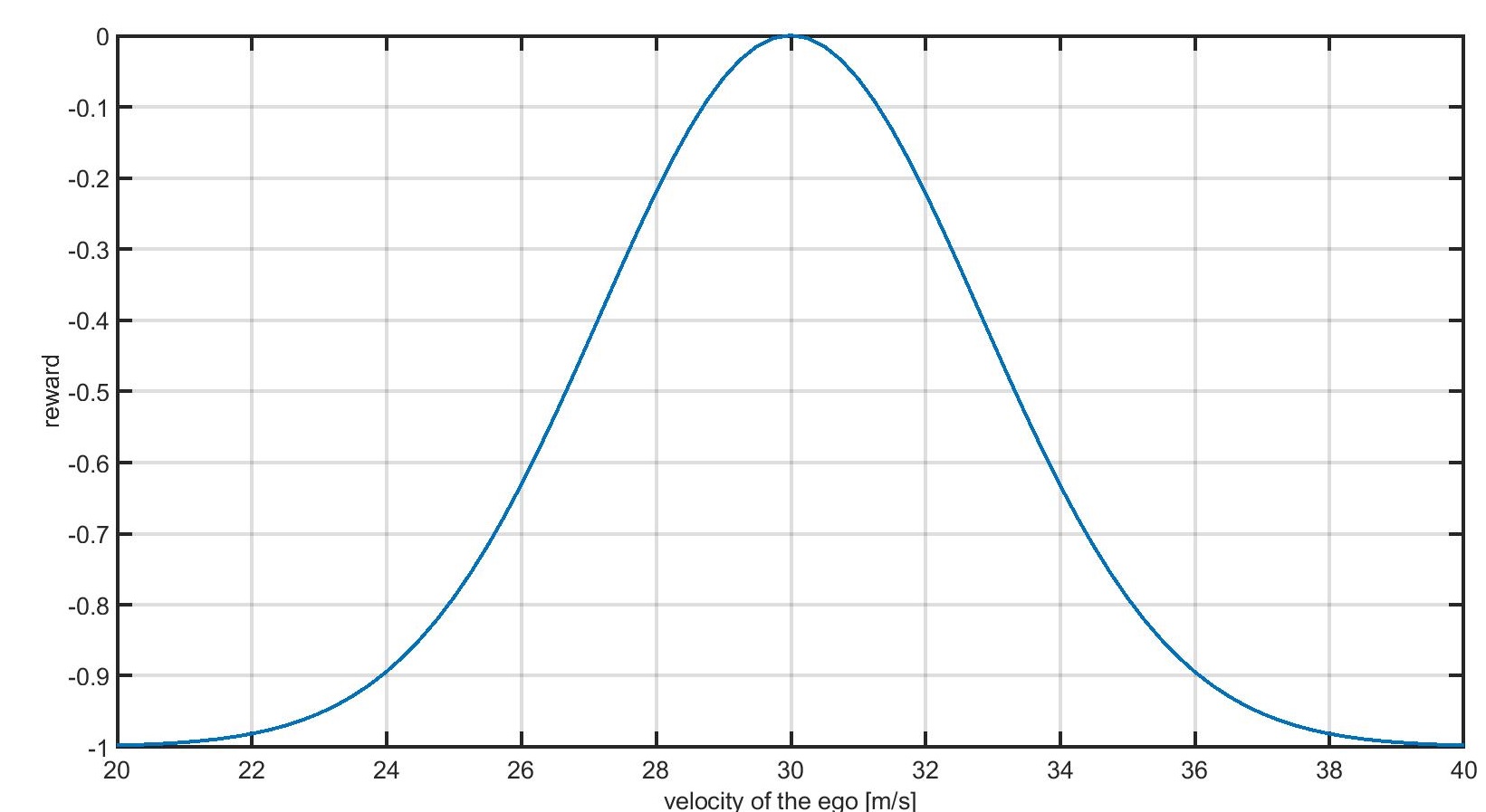}
    \caption{Reward for desired ego speed.}
    \label{fig:rewV}
  \end{subfigure}

  \caption{Reward for the ego car based on traffic condition, sub goals are weighted equally when calculating the final reward.}
  \label{fig:rewardFunc}
\end{figure}

Fig.~\ref{fig:rewardFunc} gives an indicative plot of the reward functions \eqref{eq:rewv}-\eqref{eq:rewx}, it is formulated assuming $v_\mathrm{des}\,=\,30\; m/s$ which can be achieved in the center lane i.e., $y_\mathrm{des}\,=\,3.8 m$ with a minimum safe distance $d_\mathrm{safe}\,=\,40m$. The desired values are based on the traffic condition and can change depending on the scenario. For slow/fast moving traffic the peak in Fig.~\ref{fig:rewV} will be adjusted based on the traffic condition. In this work we penalize the ego vehicle if it cannot maintain a minimum time headway of at least $1.3$ seconds.

\begin{figure}[htbp]
\centering
\includegraphics[width=12cm]{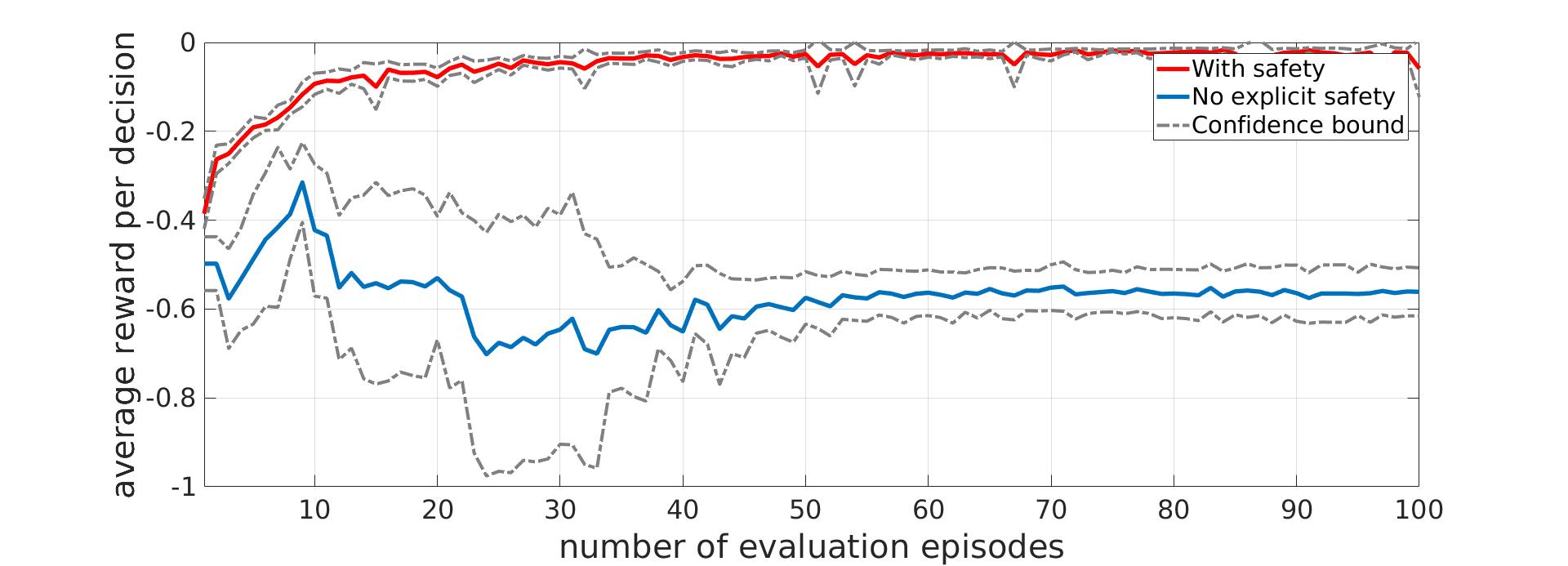}
\caption{Average learning curve with confidence bound for with and without short horizon safety check in Algorithm~\ref{alg:DRL4AV}.}
\label{fig:AvRewPerEp}
\end{figure}

During learning, we evaluate the (partially) trained DRL controller  every $100^{\text{th}}$ episode. Fig.~\ref{fig:AvRewPerEp} shows the average reward per decision during the training phase. It takes nearly $2000$ episodes for the Agent to converge. We train the DRL Agent for a total of $10000$ episodes. Where each episode lasts until $200$ samples or collision, whichever is earlier. Exploration is continuously annealed from $1$ to $0.2$  over first $7000$ episodes and then kept constant for the remaining duration of learning. The $Q$-network is a deep neural network with $2$ hidden layers each having $100$ fully connected leaky ReLU's \citep{maas2013rectifier}. We train the network using Adam optimizer \citep{kingma2014adam} with a fixed learning rate of $1e-4$.

For the highway driving task, the safety filter was found to be a key component for learning a meaningful policy. Fig.~\ref{fig:AvRewPerEp} shows the mean and confidence bound for training with and without safety filter over $200$ training iterations of Algorithm~\ref{alg:DRL4AV}. Training a standard DDQN Agent without explicit safety check could not learn a decent policy and always resulted in collision. Whereas DDQN with explicit safety check was able to converge to an optimal policy. Based on \eqref{eq:rewv} - \eqref{eq:rewx}, the maximum reward an Agent can receive is zero per decision, the average reward per decision obtained by our trained DDQN Agent with safety check is around $-0.025$.

\begin{figure}[htbp]
\centering
\includegraphics[width=12cm]{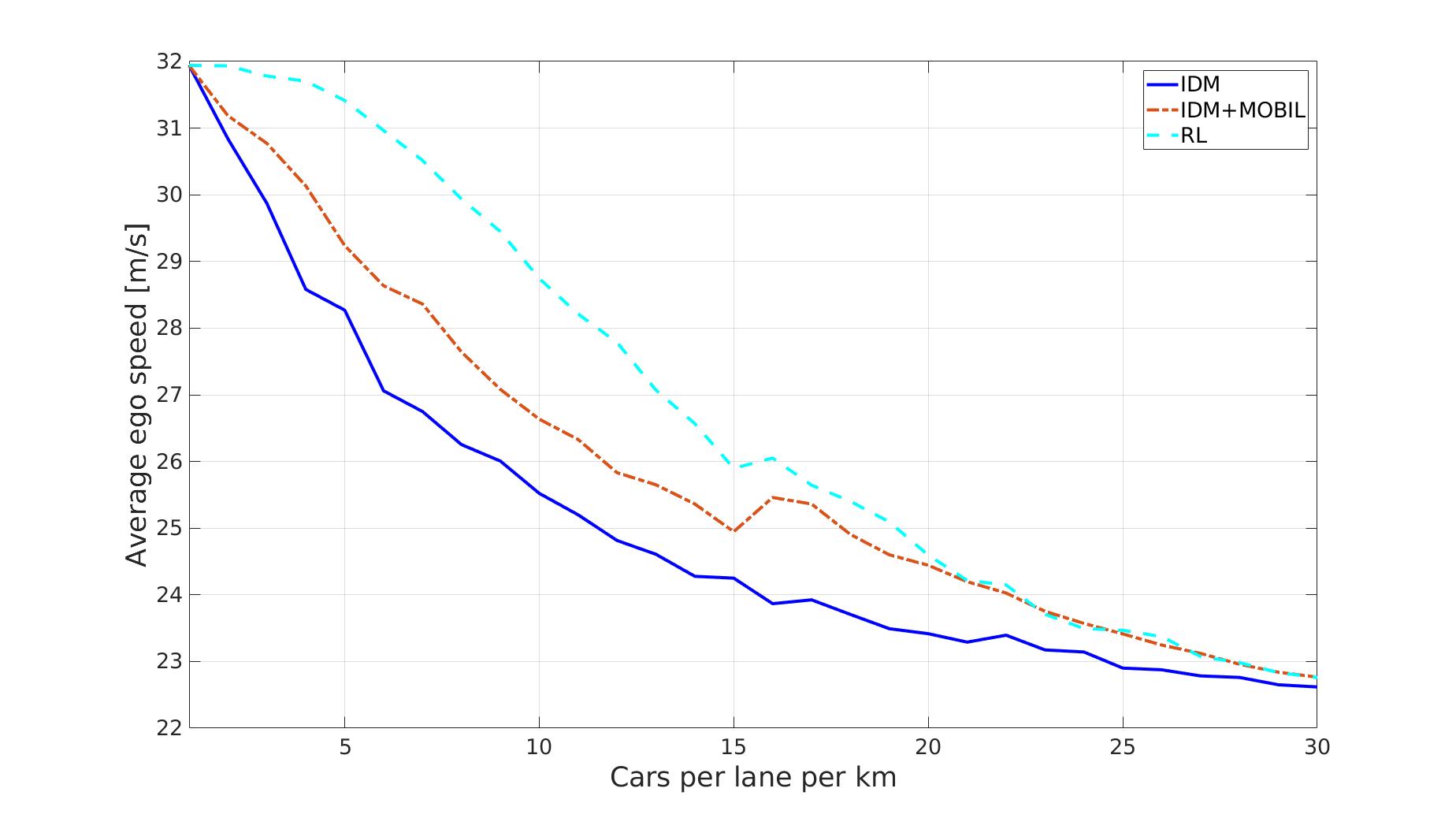}
\caption{Average speed for simple IDM controller, with lane change, and trained RL Agent.}
\label{fig:Avspd}
\end{figure}

In Fig.~\ref{fig:Avspd} We evaluate our trained DDQN Agent to obtain average velocity with increase in traffic density.  We compare this against modified safety filter from  \eqref{eq:sfCtrl}, the modification provides an acceleration command when the calculated time to collision $T_\mathrm{C}$ is higher than $T_\mathrm{A}$. This is referred as IDM in Fig.~\ref{fig:Avspd}. It must be noted IDM controller from \eqref{eq:sfCtrl} cannot initiate lane change, in order to address this
we integrate IDM with SUMO lane change decision making from \citep{kesting2007general} and \citep{erdmann2014lane}. Fig.~\ref{fig:Avspd}, clearly demonstrates advantage of RL for high level decision making when compared to model-based approaches. With the increase in traffic density both the trained DDQN Agent and the model-based lane change controller converges to IDM controller. This is anticipated since lane change is neither safe nor advantageous in higher traffic density.

\begin{figure}[htbp]
\centering
\includegraphics[width=12cm]{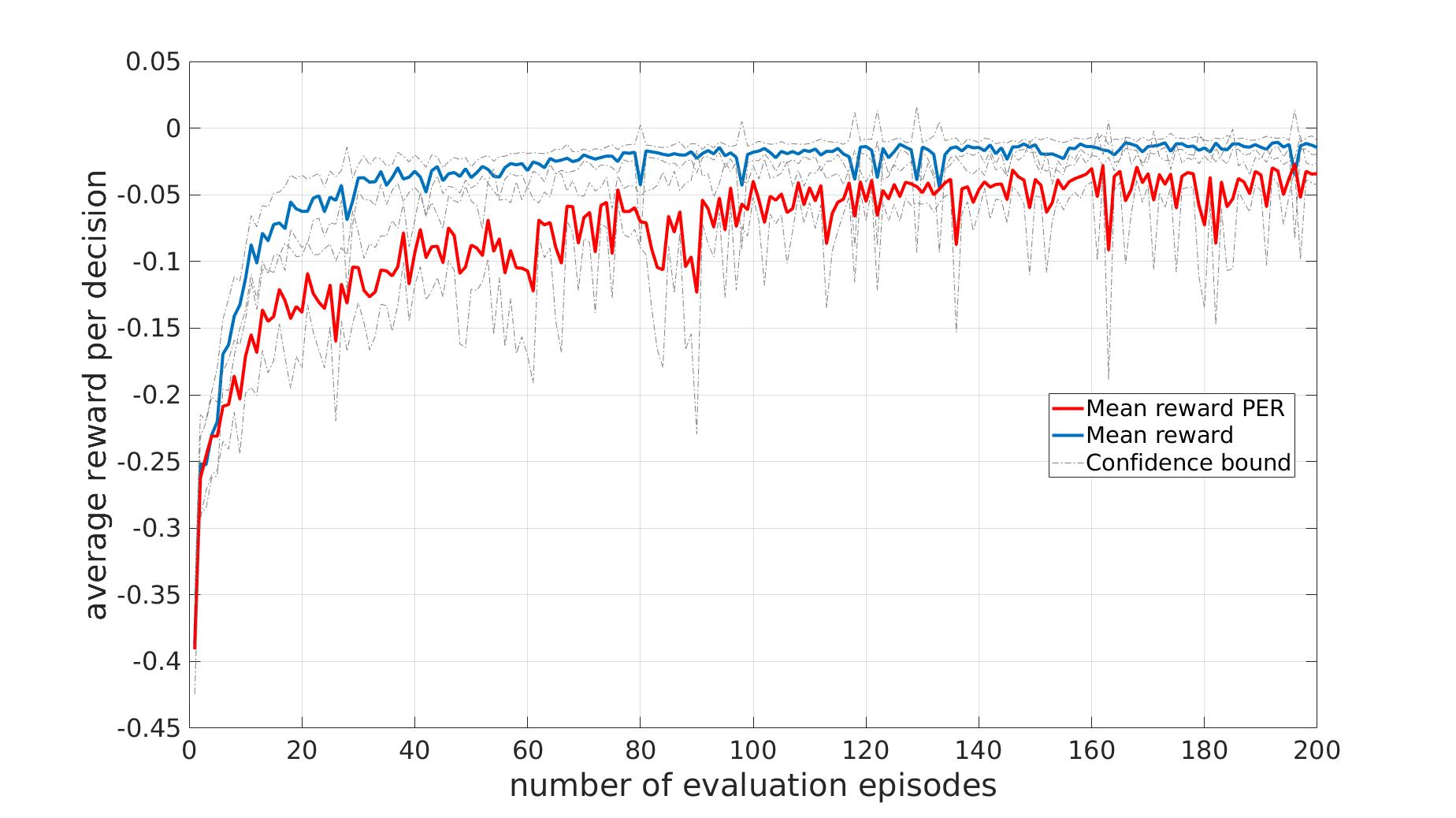}
\caption{Mean learning curve with confidence bound for Algorithm~\ref{alg:DRL4AV} and prioritized experience replay \citep{schaul2015prioritized}. In this work we used the PER implementation from \citep{stable-baselines}.}
\label{fig:CompAdap}
\end{figure}

Use of two explicit buffers namely $\mathrm{Buf}_\mathrm{S}$ and $\mathrm{Buf}_\mathrm{C}$ in Algorithm~\ref{alg:DRL4AV} to store safe and non-safe transitions is simplified version of prioritized experience replay (PER) from \citep{schaul2015prioritized}. Fig.~\ref{fig:CompAdap} shows the mean and confidence bound for training with two buffers and PER over $200$ training iterations of Algorithm~\ref{alg:DRL4AV}. For the highway driving example using two explicit buffers provides marginally better policy when compared to PER.  This can be due to clear bifurcation of safe and non-safe transitions.

\subsubsection{Continuous adaptation}\label{sec:continuous_adaptation}
\begin{figure}[htbp]
\centering
\includegraphics[width=12cm]{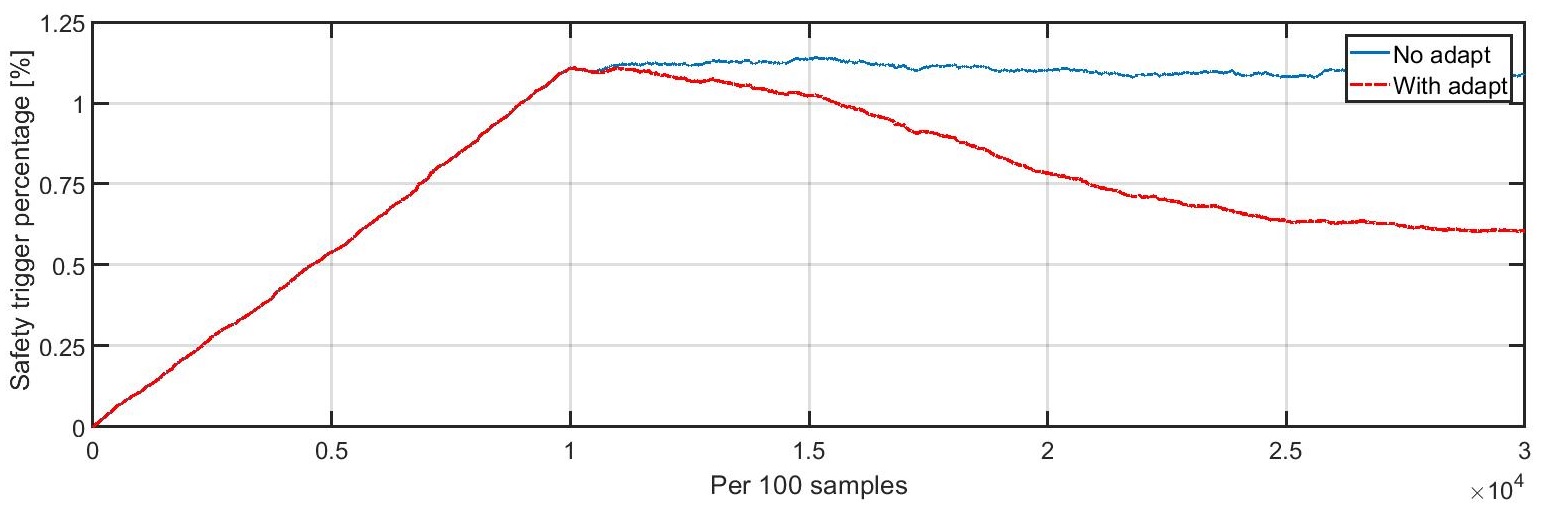}
\caption{Comparison of number of safety trigger after learning with and without continuous adaptation.}
\label{fig:CompAdap1}
\end{figure}

During the implementation phase, we replace the $\epsilon$-greedy policy, i.e., $\pi_\epsilon$ in Algorithm~\ref{alg:DRL4AV} line 5 by the learned policy $\pi$. Whenever the control decision by DDQN fails the short-horizon safety check, buffer $\mathrm{Buf}_\mathrm{C}$ is updated with additional data. Using a lower learning rate than the one used for training, $Q$-network can be retrained (line 13 until 16). Fig.~\ref{fig:CompAdap1} shows the continuous adaptation result over 30K episodes and is obtained by averaging the data over 10k episodes using a moving average filter. Because of filtering, the mean number of safety trigger increases over first 10k episodes and stays constant for no adaptation scenario whereas it monotonically decreases to a smaller value thanks to continuous adaptation. Even with continuous adaptation the mean number safety trigger never converges to zero, this may be due to
\begin{enumerate}
    \item Use of function approximation where a trained NN can potentially chose a non-safe action,
    \item Use of rigid and static safety rules.
\end{enumerate}

\subsection{Summary}
In this section we provided an introduction to basics of RL, introduced our hybrid decision making – control architecture. We explained the need for including a safety filter during RL training and inference, we also showed its effectiveness in learning a driving policy. Although the rule-based safety filter is handcrafted and predetermined it significantly improved the learning speed. Additionally the number of safety interventions reduces as the agent learns to perform optimal actions.

\section{Executing DRL Decision with Motion Control Algorithm}
\label{sec:Motion Control}
While the previous section illustrated the design methodology of Deep Reinforcement Learning (DRL) for a generic driving environment, it is best to train DRL with representative motion control that executes the DRL decision. This will ensure a higher fidelity of the vehicle - environment transition probability distribution for the DRL training. In this section, we discuss longitudinal and lateral motion control algorithms and their integration aimed at executing designated actions from AI DRL strategy.

\subsection{Longitudinal Motion Control}
\label{sec:Longitudinal Motion Control}
Longitudinal motion control systems need to provide tracking of the desired vehicle speed while maintaining a safety gap/range relative to the selected target vehicle. This functionality is provided by modern adaptive cruise control (ACC) systems which compute the longitudinal acceleration command based on the actual and set vehicle speed ($V_x$ and $V_{x.set}$), the actual and set range\footnote{Range $R$ is defined as the distance or gap between the ego and target vehicle, i.e., distance between the ego vehicle's front bumper and target vehicle's rear bumper.} ($R$ and $R_{set}$), and range rate\footnote{Range rate $\dot{R}$ is defined as relative speed between the ego and target vehicle.} ($\dot{R}$)
\begin{equation}
\label{eq:ax_cmd}
    a_{x.ACC.cmd}= f_{ACC}(V_{x.set},R_{set},V_x,R,\dot{R})
\end{equation}
The commanded acceleration is delivered by the actuation layer through modulating propulsion or brake torque. The ACC mapping function $f_{ACC}$ can be implemented in various ways. For example, \citep{rajamani2011vehicle} described an ACC mapping function that includes both car-following mode regulating steady state range/gap and speed-control mode responding to transient situations such as target vehicle cut-in or loss of its lead vehicle. The two modes were integrated through a calibrated range-range rate ($R-\dot{R}$) transition diagram. \citep{treiber2013traffic} proposed the Intelligent Driver Model (IDM) which has been widely used in traffic simulation environments and represents human car-following behavior observed on freeway and urban traffic. \citep{jin2018experimental} implemented state ([$R, \dot{R}$]) feedback controller imitating human driving behavior and experimented on connected automated vehicle. \citep{treiber2013traffic} further developed Improved IDM (IIDM) and the ACC Model to avoid unrealistically large deceleration when desired speed suddenly changes, e.g., when entering a reduced speed zone or when a vehicle cuts in.  For more variety of ACC a reader is referred to the survey \citep{xiao2010comprehensive} and references therein.

To execute DRL’s lane change decision, we leverage our production ACC design. This will avoid the need of complicated spatial-temporal optimization \citep{lee2018trajectory}. By adding scenario dependent smart ‘target selection’ to existing ACC design, the vehicle speed will change during the lane change maneuvers for a smooth transition into a new gap and a new lane speed. The elements considered in the scenarios include the lane change decision, the range and range rate of surrounding vehicles, target lane speed, and the status/prediction of lateral motion. The ‘target selection’ refers to the selected lead vehicle whose range and range-rate of the ACC mapping function (Eq. (\ref{eq:ax_cmd})) will be based on. For example, in case of staying in lane decision, which represents the base ACC functionality, the target vehicle selected is the lead vehicle of the current lane. In case of lane changing decision, the target selection is considered in two phases: before and after merging into target lane. In the first phase, both the lead vehicles in current lane and target lane will be fed into the ACC mapping function, with a calibratable desired range, and the commanded acceleration will take the minimum values of the two, thus ensuring safety. In the second phase (after merging into the target lane), the algorithm reverts to the base ACC functionality, i.e., selecting the lead vehicle of the current (new) lane as the target.

Therefore, with the scenario-based ACC always aware of the status of the lateral motion control (to be described next), smooth and human-driver-like motion control can be expected for lane change and lane keep decisions.

\subsection{Lateral Motion Control}
\label{sec:Lateral Motion Control}
\begin{figure}[t]
\centering
\includegraphics[width=8cm]{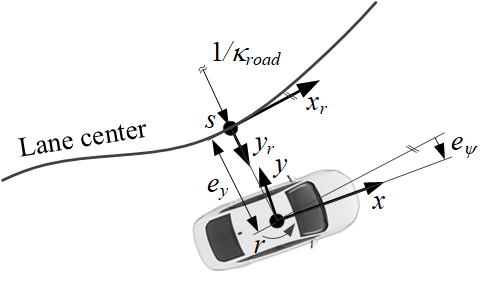}
\caption{Coordinate frames and vehicle motion dynamics variables. Legend: $(x_r-y_r)$ is the curvilinear coordinate road frame, $(x-y)$ is the Cartesian coordinate vehicle body frame, $\kappa_{road}$ is road curvature, $e_y$ is path offset, $e_{\psi}$ is heading offset, $s$ is curvilinear coordinate, $r$ is vehicle yaw rate.}
\label{fig:CoordinateFrames}
\end{figure}

For lane centering and lane changing maneuvers, we first introduce the relevant vehicle-road kinematics. Fig. \ref{fig:CoordinateFrames} shows the definition of coordinate frames and variables for vehicle road/lane level localization. The vehicle motion dynamics is considered with respect to the curvilinear Frenet-Serret coordinate frame ($x_r$-$y_r$). The states of the vehicle-road kinematics include the travel distance along the curvilinear coordinate $s$, the path offset $e_y$, and the heading offset $e_{\psi}$, while the curvature of the road is considered as an exogenous input, denoted by $\kappa_{road}$. Vehicle motion - the longitudinal velocity $V_x$, lateral velocity $V_y$, and yaw rate $r$ - are described in the Cartesian coordinate vehicle body frame $(x-y)$. Vehicle motion curvature $\kappa$ is defined as $\kappa=r/V_x$.\footnote {Alternatively, the curvature can be computed based on the front wheel angle $\delta$ as $\kappa=\delta/(L+K_uV_x)$  where $L$ is the vehicle wheelbase, and $K_u$ is the vehicle understeer gradient in units of rad-s$^2$/m. The understeer gradient is defined as: $K_u=m_f/C_f-m_r/C_r$ where $m_f$ and $m_r$ are the front and rear axle mass, $C_f$ and $C_r$ are the front and rear axle cornering coefficients.}

\begin{figure}[b]
\centering
\includegraphics[width=9cm]{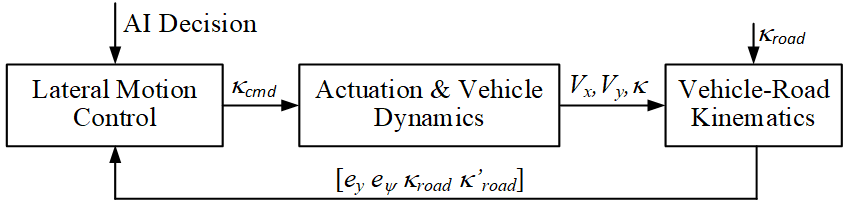}
\caption{Block diagram of lateral motion control and plant. Legend: $\kappa_{cmd}$ is vehicle motion curvature command, $V_x$ and $V_y$ are vehicle longitudinal and lateral velocity, $\kappa$ is vehicle motion curvature, $\kappa_{road}$ is road curvature.}
\label{fig:LMC_Block_Diagram}
\end{figure}

Fig. \ref{fig:LMC_Block_Diagram} shows the system block diagram. The lateral motion controller performs lane localization and computes vehicle motion curvature command. The plant includes actuator and vehicle dynamics, as well as vehicle-road kinematics/geometry.

Vehicle-Road kinematics in the Frenet-Serret coordinate frame is given by \citep{werling2010optimal}
\begin{subequations}
\begin{alignat}{3}
\dot{e}_y&=V_x\sin e_{\psi}-V_y\cos e_{\psi} \label{eq:ey_dot_general} \\
\dot{e}_{\psi} &=\kappa_{road}\dot{s}-r \label{eq:epsi_dot_general} \\
\dot{s}&=\frac1{1-\kappa_{road}e_y}(V_x\cos e_{\psi}-V_y \sin e_{\psi}). \label{eq:s_dot_general}
\end{alignat}
\end{subequations}
Since the value of heading offset, road curvature, and lateral velocity are relatively small during the lane centering and lane change maneuvers on highways, the vehicle-road kinematics can be further simplified to the following
\begin{subequations}
\begin{alignat}{3}
\centering
    \dot{e}_y &=V_xe_{\psi} \label{eq:ey_dot} \\
    \dot{e}_{\psi} &=V_x(\kappa_{road}-\kappa) \label{eq:epsi_dot} \\
    \dot{s}&=V_x. \label{eq:s_dot}
\end{alignat}
\end{subequations}
The path offset ($e_y$) dynamics from ($\kappa_{road}-\kappa$) are represented by a double integrator with crossover frequency of $|V_x|$. The vehicle motion curvature $\kappa$ is the system input, whereas the road curvature $\kappa_{road}$ is considered as a system disturbance.

In typical applications, the system states $e_y$ and $e_{\psi}$ and road curvature $\kappa_{road}$ are obtained through image processing of video frames taken from a front view camera and by fitting the lane markers to a 3$^{rd}$ order polynomial \citep{wang2000lane}. Coefficients of these lane marker polynomials are then fused to obtain the coefficients of the lane center polynomial with respect to vehicle coordinate ($x$-$y$),  referred to as the path polynomial, where
\begin{equation}
\label{eq:y_path}
    y_{path}=a_0+a_1x+a_2x^2+a_3x^3.
\end{equation}

In this format, coefficients $a_0$ and $a_1$ are equivalent to the path offset ($e_y$) and heading offset respectively ($e_{\psi}$) (see Fig. \ref{fig:CoordinateFrames}). Coefficients $a_2$ and $a_3$ are related to the road curvature $\kappa_{road}$  and curvature rate  $\kappa'_{road}=d\kappa_{road}/dx$. Based on vector differential calculus, curvature of a parametric curve $y=y(x)$ in the $x-y$ plane is given by \citep{kreyszig1993advanced}
\begin{equation}
\label{eq:param_curve_curvature1}
    \kappa(x)=|y''|/(1+y'^2)^{3/2}
\end{equation}
where $y'=dy/dx$. Computing $y'_{path}$ and $y''_{path}$ from Eq. (\ref{eq:y_path}) and replacing with $y'$ and $y''$ in (\ref{eq:param_curve_curvature1}) gives
\begin{equation}
\label{eq:param_curve_curvature2}
    \kappa(x)=\frac{2a_2+6a_3x}{[1+(a_1+2a_2x+3a_3x^2)^2]^{3/2}}.
\end{equation}
Assuming small heading offset $a_1$\footnote{For highway driving nominal value of $a_1\leq 5\, deg$ hence the error introduce by the small heading offset assumption is  $\leq1\%$.}, the curvature and curvature rate of the path (road) at the origin ($x=0$) are equal to
\begin{subequations}
\begin{alignat}{2}
    \kappa_{road} & =2a_2 \label{eq:curvature} \\
    \kappa'_{road} & =6a_3 \label{eq:curvature_rate}
\end{alignat}
\end{subequations}

\subsubsection{Lane Centering Control}
Vehicle-Road kinematics for lane centering applications is given by Eqs. (\ref{eq:ey_dot}) and (\ref{eq:epsi_dot}) where the road curvature is considered as an exogenous disturbance which can be measured. The proposed curvature command (Eq. (\ref{eq:kappa_cmd})) represent a generic lane centering algorithm which includes both a feedforward term of the road curvature and the feedback terms. As such, with different coefficient design, it can represent various well-known path following algorithms/approaches including Pure Pursuit \citep{coulter1992implementation}, Bezier curves based planning \citep{giersiefer_dornhege_klein_klein_2019}, or lookahead point design \citep{tseng2002steering}. Our contribution is to introduce a novel and heuristic calibration approach for any of the above path following approaches to achieve desired closed-loop system response. The generic control law is given by

\begin{equation}
\label{eq:kappa_cmd}
    \kappa_{cmd}=\underbrace{K_{ff}(\kappa_{road}+V_xT_{prev}\kappa'_{road})}_\text{feedforward} + \underbrace{K_ye_y+K_{\psi}e_{\psi}}_\text{feedback}
\end{equation}
where $K_{ff}$ is the feedforward gain, $T_{prev}$ is the preview time that can be effectively utilized to compensate for actuation delays, $K_y$ is the path offset gain, and $K_{\psi}$ is the heading offset gain. Substituting $\kappa$ in (\ref{eq:epsi_dot}) with $\kappa_{cmd}$ from (\ref{eq:kappa_cmd}) and combining with (\ref{eq:ey_dot}) gives the closed-loop dynamics

\begin{equation}
\label{eq:closed_loop_dynamics}
    \begin{bmatrix}
        \dot{e}_y \\
        \dot{e}_{\psi}
    \end{bmatrix}
    = V_x
    \begin{bmatrix}
        0 & 1 \\
        -K_y & -K_{\psi}
    \end{bmatrix}
    \begin{bmatrix}
        e_y \\
        e_{\psi}
    \end{bmatrix}
    + V_x(1-K_{ff})\begin{bmatrix}
        0 & 0\\
        1 & V_xT_{prev}
    \end{bmatrix}
    \begin{bmatrix}
        \kappa_{road} \\
        \kappa'_{road}
    \end{bmatrix}.
\end{equation}

Eigenvalues of this 2$^{nd}$ order system are

\begin{equation}
\label{eq:closed_loop_eigenvals}
    \lambda_{1,2}=\frac{V_xK_{\psi}}{2} \pm V_x\sqrt{K_y} \sqrt{\frac{K_{\psi}^2}{4K_y}-1}.
\end{equation}

To design the state feedback gains in terms of desired closed-loop metrics such as the natural frequency and damping ratio, we compare eigenvalues (\ref{eq:closed_loop_eigenvals}) with eigenvalues of the 2$^{nd}$ order system expressed in terms of the natural frequency $\omega_n$ and damping ratio $\zeta$

\begin{equation}
\label{eq:2nd_order_system_eigenval}
    \lambda_{1,2}^*=-\omega_n\zeta \pm \omega_n\sqrt{\zeta^2-1}
\end{equation}

Derived feedback gains are given by
\begin{subequations}
\begin{alignat}{2}
    K_y&=\frac{\omega_n^2}{V_x^2} \label{eq:Ky} \\
    K_{\psi}&=\frac{2\zeta\omega_n}{V_x}. \label{eq:K_psi}
\end{alignat}
\end{subequations}

Instead of the natural frequency as the tuning parameter, for lane centering applications it is convenient to use alternative time response metrics such as the transient time to reach 95\% of the target value or 5\% of the initial value. This time is referred to as the response time and denoted by $T_r$. For practical damping ratios in range between 0.7 and 0.9, relation between the transient time, closed-loop frequency and damping ratio can be obtained by numerical optimization (with coefficient of determination $R^2=0.99$) and is given by
\begin{equation}
\label{eq:Tr}
    T_r\approx\frac{4.3\zeta}{\omega_n}.
\end{equation}

Solving (\ref{eq:Tr}) for $\omega_n$ by using (\ref{eq:Ky}) and (\ref{eq:K_psi}) gives the feedback gains in terms of the transient time as
\begin{subequations}
\begin{alignat}{2}
    K_y&=\frac{18.5\zeta^2}{T_r^2V_x^2}\label{eq:Ky1} \\
    K_{\psi}&=\frac{8.6\zeta^2}{T_rV_x} \label{eq:K_psi1}.
\end{alignat}
\end{subequations}

\begin{figure}[h!]
\centering
\includegraphics[width=8cm]{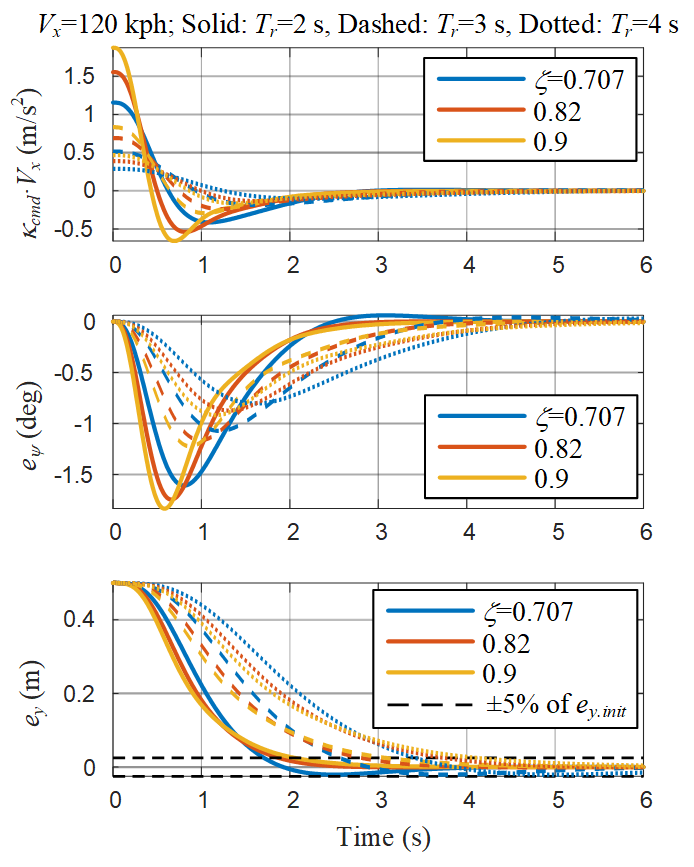}
\caption{Initial condition response Lane change control for various damping ratios $\zeta$ and response times $T_r$. The initial conditions are: $e_{y.0}=0.5\, m$ and $e_{\psi.0}=0\, deg$. Actuation dynamics is approximated by 2$^{nd}$ order term with $\omega_{n.act}=5\omega_n$ and $\zeta_{act}=0.6$.}
\label{fig:Initial_condition_response}
\end{figure}

Fig. \ref{fig:Initial_condition_response} illustrates vehicle response with non-zero initial conditions using the proposed controller for different choice of the design parameters.

\subsubsection{Lane Change Control}
Lane change control systems need to provide a smooth lateral shift of a vehicle from current to target lane within specified time and with small or zero overshoot. Typical duration of lane changes\footnote{Duration is defined as time required to reach path offset relative to the target lane center that is withing the range of nominal lane centering oscillations in range of 0.1 to 0.2 m. For nominal lane width of 3.4 m, this corresponds to approximately 5\% of the lane width or the path offset at the lane change start.} is in the range of 5 and 7 seconds with lateral accelerations amplitudes in the range of 0.4 and 0.8 $m/s^2$ (for making a lane change on straight roads). One can divide the problem into explicit path planning and path following, or alternatively integrate both components from the perspective of smooth transitioning into a new lane center. The latter eliminates the need of specifying way points and enables the designer to reuse the lane centering control methodology described in the previous section.  In the case of integrated controller, the path is switched from the current to the target lane which introduces a step change in the path offset $e_y$. The lane centering controller will follow the new set point by driving the vehicle towards the target lane. In order to achieve the desired transient time and damping ratio, the controller gains are scheduled by using (\ref{eq:Tr})-(\ref{eq:K_psi1}), where $T_r$ corresponds to the lane change duration. Benefits of this controller are in the ease of tuning, adaptation to initial conditions or changing conditions which may arise from lane change aborts or driver interactions.

Since this controller is linear, the initial curvature command in response to a step change in set point is large (see Fig. \ref{fig:Initial_condition_response}) creating a pronounced lateral acceleration and jerk. Although the actual acceleration and jerk will be partially filtered out by the actuation and vehicle dynamics, a systematic solution is proposed below.

The pronounced initial acceleration and jerk can be overcome by adding amplitude and rate limits to the state feedback part of the control input. The limits need to be sufficiently large in order not to affect the performance, e.g., not to cause oscillations or even limit cycle. Methods such as describing function analysis \citep{slotine1991applied,ackermann1999robust,duda1998flight} can be applied to ensure stability at the presence of calibrated rate and amplitude limits. We propose to utilize limits obtained from well-known quintic polynomial path planning (see, e.g., \citep{papadimitriou2003fast}).

Path in time domain described by the quintic polynomial is given by: $y(t)=a_0+a_1 t+a_2 t^2+a_3 t^3+a_4 t^4+a_5 t^5$. For given initial and final conditions ($t_f$ is final time): $y(0)=0$, $\dot{y}(0)=0$, $\ddot{y}(0)=0$, $y(t_f)=y_f$, $\dot{y}(t_f)=0$, $\ddot{y}(t_f)=0$, the coefficients are equal to:$a_0=0$, $a_1=0$, $a_2=0$, $a_3=10y_f/t_f^3$, $a_4=-15y_f/t_f^4$, and $a_5=6y_f/t_f^5$. From the second derivative of the polynomial, time instances at which lateral acceleration peaks occur are $t_{1,2}=(3 \pm \sqrt{3})/6t_f \Rightarrow t_1 \approx0.21t_f$, $t_2 \approx 0.79t_f$. Accordingly, the acceleration maximum is
\begin{equation}
\label{eq:ay_max}
    a_{y.max}=\frac{5.77y_f}{t_f^2}
\end{equation}
Maximum jerk occurs at the beginning and end with magnitude are equal to
\begin{equation}
\label{eq:j_max}
    j_{max}=\frac{60y_f}{t_f^3}
\end{equation}

\begin{figure}[h!]
\centering
\includegraphics[width=8cm]{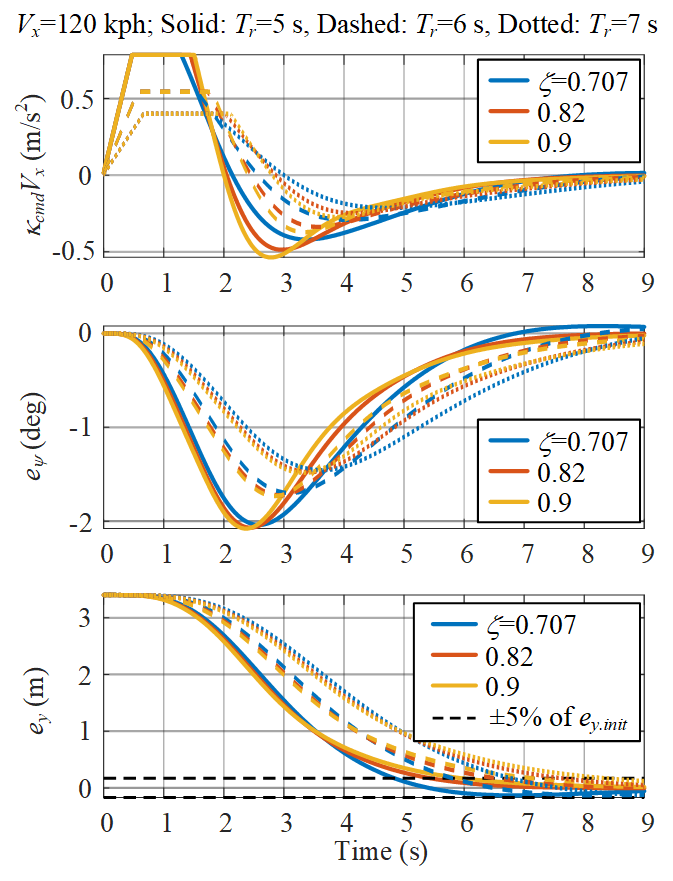}
\caption{Lane change control with control input amplitude and rate limiting. Actuation dynamics is approximated by 2$^{nd}$ order term with $\omega_{n.act}=5\omega_n$ and $\zeta_{act}=0.6$. }
\label{fig:LaneChangeSimulationResultsWithLimits}
\end{figure}

Obtained lateral acceleration and jerk limits\footnote{For lane width of 3.4 m and lane change duration of 6 s, the limits are $a_{y.max}=0.54\, m/s^2$ and $j_{max}=0.94\, m/s^3$.} provide comfortable maneuver yet are high enough to prevent oscillations or limit cycles which can be proved by nonlinear simulations or describing function analysis.  Fig. \ref{fig:LaneChangeSimulationResultsWithLimits} shows simulation results with amplitude and rate limiting of the curvature command. We see that although there is a major effect on the initial transient response, the target metrics in terms of the duration and damping ratio are preserved.

\subsection{Summary}

In this section, we presented longitudinal and lateral motion control, and how they work together to execute the decision made by higher level AI (DRL). For the longitudinal control, we introduced a smart ‘target’ selection logic to leverage production ACC design, and to facilitate the integration with lateral motion control for delivering smooth and human-driver-like lane change maneuvers. For the lateral control, we introduced a unified lane centering and lane changing algorithm with a general control algorithm formula that can be calibrated to represent various well-known path following design methodologies. We further introduced a new calibration approach that can automatically adjust the state feedback gains to achieve desired system damping, closed loop bandwidth, or 0-95\% set point response time.
In addition, for lane change maneuvers, we integrated both path planning and path following aspects. Since the integrated control does not try to regulate the vehicle to specific waypoints, no explicit waypoints designation is required. It provides implicit, smooth, and seamless re-planning in case of unexpected vehicle motion which may arise from unintended steering movement, and in case of abrupt path offset change (set point) change in the event of lane change abort (to return to the original lane). We further recommended a way to analyze and calibrate the limit of lateral acceleration and jerk – the nonlinearities purposely added to the controller for comfort and actuation constraints – for assuring both subjective and objective performance.

\section{Generic Safety Filter Design with Control Barrier Functions} \label{Rahman2021CSD}
In this section we develop a more generic safety filter than the heuristic handcrafted rules utilized in Section 2, because the latter may not be sufficient for avoid all types of collisions. The performance of an Agent depends on what it sees during training. Since the handcrafted heuristic rules are designed with certain imminent collision threats in mind, they could miss corner cases in the real-world, and even in a simulation world with long tail distribution where simulated target vehicles behaving beyond the engineers’ imagination. For this reason, the rule-based safety filter can be ‘brittle’. Hence a collision-avoidance intervention mechanism that is not specific to certain ego-target vehicle poses/situations is developed here to serve as a generic safety filter. 

Control Barrier Functions (CBF) have been used as a method to compute minimally-invasive feedback control that satisfy various prescribed system constraints. For AVs, CBFs can provide a computationally efficient approach to avoid collision by correcting ego vehicle motion based on its relative position and velocity compared with surrounding road users. The CBF framework used here assumes that a nominal control signal $u_0$ is computed by the autonomous driver, in this case an Agent. That signal is then evaluated by the CBF for safety. If $u_0$ is safe, the CBF safety filter does not modify it. If it is deemed unsafe, the safety filter overrides the nominal control and calculates a minimally invasive safe control that can avoid collision. The advantage of using a CBF based collision avoidance system is that it provides a safe steering and longitudinal acceleration command that is as close as possible (in the Euclidean sense) to the nominal control. This results in a much less intrusive safety filter compared to the rule based safety filter from Section \ref{DRL+RB}.

However, it is not straightforward to decide if the actuation of braking or steering should be used, (or how the actuation should be combined/split) to best correct vehicle motion when CBF intervenes to avoid a collision. To arbitrate between steering and/or braking,we use Contextual Selection of Decoupled CBF. This algorithm addresses the above issue by taking advantage of the structure and rules of the road, as well as the knowledge of steering/braking efficacy for different situations based on physics. Therefore, the proposed logic and resulted methodology, will determine steering and braking action required to effectively and decisively impose decoupled longitudinal and lateral CBFs in order to avoid collisions. The algorithm is designed to use the same input structure, i.e. affordance indicators as the DRL Agent. In addition to the affordance indicators used by the Agent, the CBF safety filter also uses the heading with respect to the road of the six surrounding traffic vehicles. The algorithm has been verified in extensive simulation with varying traffic and road users.

\subsection{Control Barrier Functions}
\begin{figure}[!ht]
\centering
\includegraphics[width = 0.9\linewidth]{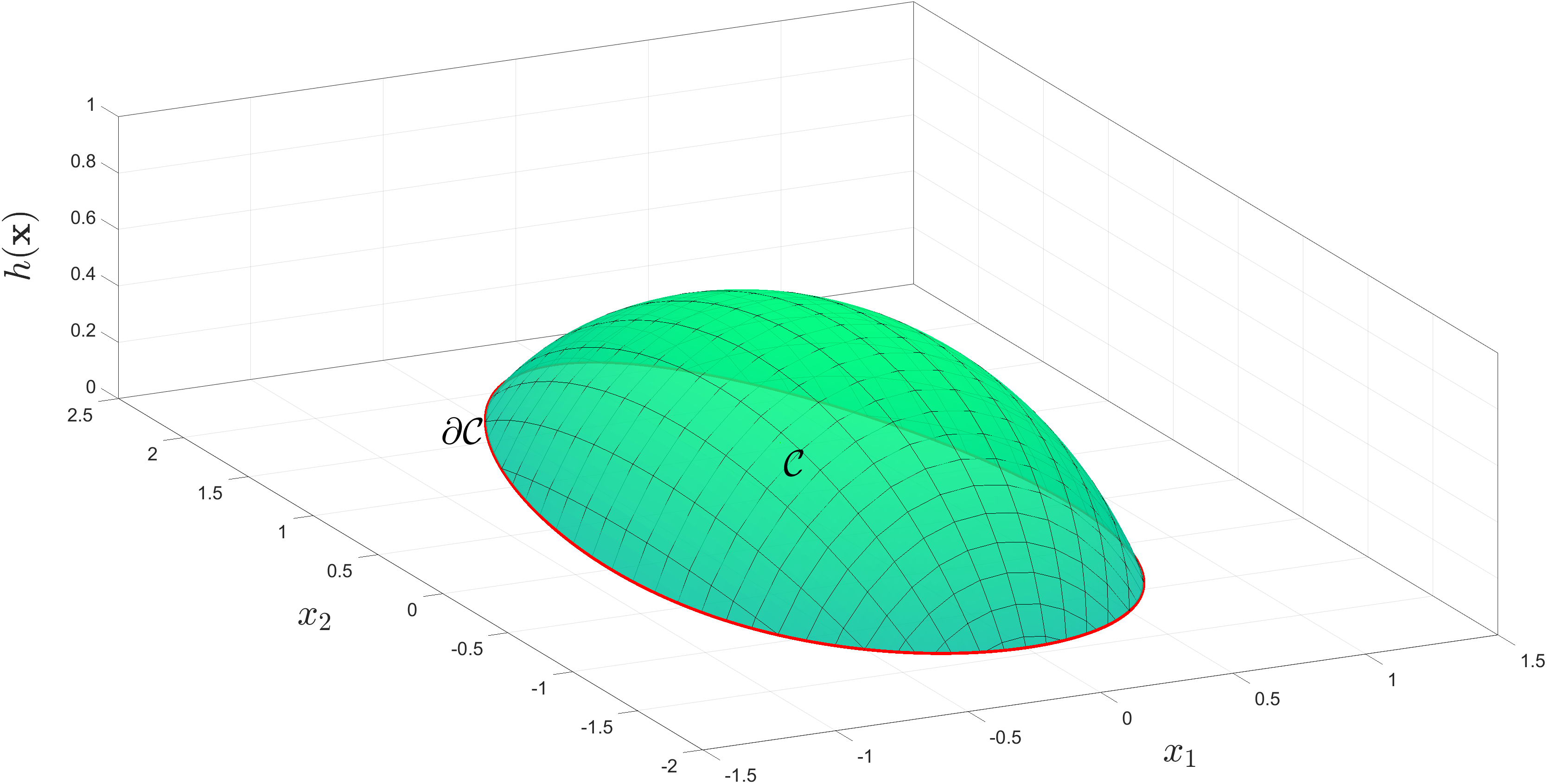}
\caption{Example Barrier Function.}
\label{fig:CBF_theory}
\end{figure}

Barrier Functions have been used to enhance system robustness \citep{prajna}, while Control Barrier Functions (CBFs) have been used as a method to provide minimally-invasive feedback control that satisfies various constraints prescribed on that system \citep{wieland2007constructive,ames2016control,jankovic2018robust,xiao2019control}. Here we mix the two concepts because we would like to apply them to systems with control inputs that are outside of the ego (the entity doing the calculations) control. Consider a control affine system
\begin{align}
    \dot{\textbf{x}} = f(\textbf{x}) + g(\textbf{x})u , \label{sys}
\end{align}
where $\textbf{x} \in \BBR^n$ and $u \in \BBR^m$ is the control input. We assume that a control input $u_0$ is computed by a driver (or an Agent/controller) to achieve some objective, and it is known. Let us also assume that an additional control objective is to keep the state of the system in a closed admissible set $\SC \subset \BBR^n $ defined as
\begin{equation}
\begin{aligned}
    \SC &\isdef \{ {\textbf{x} \in \BBR^n : h(\textbf{x}) \ge 0} \}, \\
    \partial \SC &\isdef {\textbf{x} \in \BBR^n : h(\textbf{x}) = 0}, \\
    \textrm{Int} (\SC) &\isdef {\textbf{x} \in \BBR^n : h(\textbf{x}) > 0},
    \end{aligned}
\end{equation}
where $h: \BBR^n \rightarrow \BBR$ is a twice continuously differentiable function. In addition, we assume that $\textrm{Int}(\SC) \ne \emptyset$, where $\textrm{Int}(\SC)$. Fig. \ref{fig:CBF_theory} shows an example Barrier Function $h(\textbf{x})$, the admissible set $\SC$ and the boundary $\partial \SC$.
$h(\textbf{x})$ is relative degree 1 with respect to the control inputs if $h_{\textbf{x}} \cdot g(\textbf{x}) \ne \emptyset$, where $h_{\textbf{x}} = \frac{\partial h(\textbf{x})}{\partial\textbf{x}}$. For the system (\ref{sys}) with a given control input $\bar u$ a function $h(\textbf{x})$ is a Barrier Function \citep{ames2016control} with respect to the admissible set $\SC$ if $h(\textbf{x})$ is relative degree 1 with respect to the control inputs and
\begin{align}
   \dot h(\textbf{x},\bar{u}) + l_0 h(\textbf{x}) \ge  0,
\end{align}
where $\dot h(\textbf{x},\bar{u}) = h_{\textbf{x}} \cdot f(\textbf{x}) + h_{\textbf{x}} \cdot g(\textbf{x})\bar{u}$.
If a Barrier Function is relative degree 2 with respect to the control inputs (i.e. $h_{\textbf{x}} \cdot g(\textbf{x}) \equiv 0$, as is the case later in this chapter) we can -- motivated by \citep{nguyen} -- consider the second order barrier constraint: $\ddot{h}(\textbf{x}) + l_1 \dot{h}(\textbf{x}) + l_0 h(\textbf{x}) \ge 0$. The parameters $l_1, l_0$ are selected such that the roots of the polynomial $s^2 + l_1s +l_0$ are negative real.
The forward invariant set is $\{ x: h > 0 \: \textrm{and} \: h_{\textbf{x}} \cdot f(\textbf{x}) > -\lambda_i h \}$, where $\lambda_i$ is either one of the two roots of $s^2 + l_1 s + l_0 = 0$.
A barrier function $h(\textbf{x})$ is a  CBF with respect to the admissible set $\SC$ if $h$ is differentiable in $\textrm{Int}(\SC)$, $h(\textbf{x}) \rightarrow 0$ as $\textbf{x} \rightarrow  \partial\SC$, and for $\textbf{x} \in \textrm{Int}(\SC)$
\begin{align}
h_{\textbf{x}} \cdot g(\textbf{x})\bar{u} = 0 \implies h_{\textbf{x}} \cdot f(\textbf{x}) + l_0 h(\textbf{x}) \ge 0
\end{align}
if $h(\textbf{x})$ is relative degree 1 with respect to the control inputs and
\begin{align}
\frac{\partial (h_{\textbf{x}} \cdot f(\textbf{x}))}{\partial \textbf{x}} \cdot g(\textbf{x})\bar{u}  = 0 \implies \frac{\partial (h_{\textbf{x}} \cdot f(\textbf{x}))}{\partial \textbf{x}} \cdot f(\textbf{x}) + l_1 (h_{\textbf{x}} \cdot f(\textbf{x})) + l_0 h(\textbf{x}) \ge 0
\end{align}
if $h(\textbf{x})$ is relative degree 2 with respect to the control inputs.
\subsubsection{Decoupled Barrier Functions for Autonomous Driving}
\begin{figure}[!ht]
\centering
\includegraphics[width = 1\linewidth]{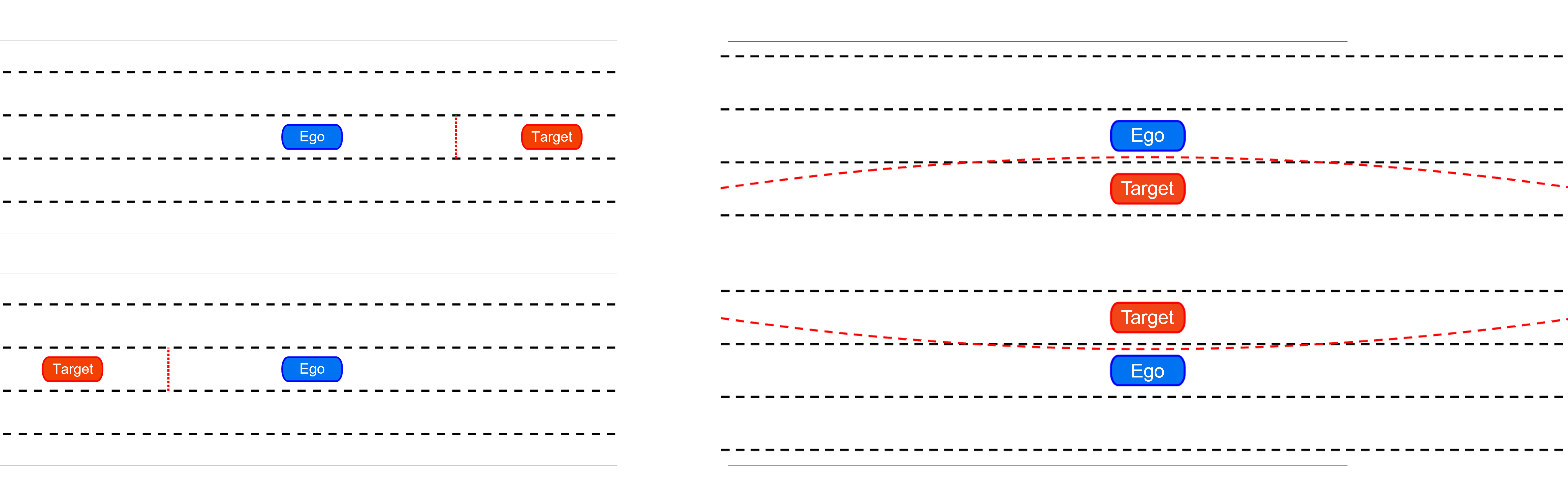}
\caption{Longitudinal and Lateral Barriers.}
\label{fig:longlatcbf}
\end{figure}

We define the longitudinal collision avoidance barrier functions as
\begin{equation}
\begin{aligned}
    h_{x,\textrm{F}} &= x_\rmT - k_v v_\rmH -d_{x,\textrm{min}} - \frac{L_\rmH}{2} - \frac{L_\rmT}{2}, \\
    h_{x,\textrm{R}} &= -x_\rmT - k_v v_\rmT -d_{x,\textrm{min}} - \frac{L_\rmH}{2} - \frac{L_\rmT}{2},\\
\end{aligned}
\label{eq:lon_CBF}
\end{equation}
where the subscript $\rmF$ is for the forward barrier and the subscript $\rmR$ is for the rearward barrier. Similarly, we define the lateral collision avoidance barrier functions as
\begin{equation}
\begin{aligned}
    h_{y,\textrm{L}} &= -y_\rmT - d_{y,\textrm{min}} + c_\rmb x_\rmT^2, \quad
    h_{y,\textrm{R}} &= y_\rmT - d_{y,\textrm{min}} + c_\rmb x_\rmT^2,
\end{aligned}
\end{equation}
where the subscript $\rmL$ is for the barrier to the left of the ego, the subscript $\rmR$ is for the barrier to the right of the ego vehicle, $L_\rmT$ is the length of the target vehicle, $d_{x,\textrm{min}}$ is the minimum longitudinal distance allowed between vehicle bumpers, $d_{y,\textrm{min}}$ is the minimum lateral distance allowed between vehicle centers, and $c_\rmb$ is a coefficient that determines the amount of bowing in the lateral barrier. The bowing is meant to reduce the steering effort required to satisfy the collision avoidance constraint when the target is far away from the ego. The minimum lateral distance of the barrier $d_{y,\textrm{min}}$ is only enforced when the ego is driving alongside the ego (i.e $x_\rmT = 0$). The possible longitudinal and lateral barriers are illustrated in Fig. \ref{fig:longlatcbf}. The dotted red line shows the position which the front/rear/left/right side of the ego vehicle must maintain in order to satisfy the barrier constraint.
\vspace{-2em}
\vspace{-5em}
\subsection{Calculation of Barrier Constraints}
\subsubsection{Longitudinal Barrier}
We use the simplified bicycle model (\ref{eq:simpdyn}) to compute $\dot{h}_{x,F}$, $\dot{h}_{x,R}$, $\ddot{h}_{x,R}$, which are used to compute the constraints for the longitudinal barrier. Since $h_{x,\rmF}$ is relative degree 1 with respect to the control input $\alpha$, only $\dot{h}_{x,F}$ is required to calculate the constraints for vehicles ahead of the ego vehicle.
\begin{equation}
\begin{aligned}
    h_{x,\rmF} &= x_\rmT - k_v v_{\rmH} - d_{x,\textrm{min}} - \frac{L_\rmH}{2} - \frac{L_\rmT}{2},\\
    \dot{h}_{x,\rmF}(\alpha) &= -gk_v\alpha + \Big(v_\rmT \textrm{cos}(\phi_\rmT) - v_\rmH \textrm{cos}(\phi_\rmH)\Big)\\
    h_{x,\rmR} &= -x_\rmT - k_v v_{\rmT} - d_{x,\textrm{min}} - \frac{L_\rmH}{2} - \frac{L_\rmT}{2},\\
    \dot{h}_{x,\rmR} &= \Big(v_\rmT \textrm{cos}(\phi_\rmT) - v_\rmH \textrm{cos}(\phi_\rmH)\Big)\\
    \ddot{h}_{x,\rmR}(\alpha) &= g\textrm{cos}(\phi_\rmH)\alpha \\
\end{aligned}
\end{equation}

\subsubsection{Lateral Barriers}
We use (\ref{eq:simpdyn}) to compute $\dot{h}_\rmL$, $\ddot{h}_\rmL$, $\dot{h}_\rmR$, and $\ddot{h}_\rmR$ which are used to compute the constraints for the lateral barriers. We treat the nominal acceleration $\alpha_0$ as a disturbance into the system.
\begin{equation}
\begin{aligned}
    h_{y,\textrm{L}}(c_\rmb) &= -y_\rmT - d_{y,\textrm{min}} + c_\rmb x_\rmT^2,\\
    \dot{h}_{y,\textrm{L}}(c_\rmb) &= v_\rmH \textrm{sin}(\phi_\rmH) - v_\rmT \textrm{sin}(\phi_\rmT) + 2c_\rmb x_\rmT(v_\rmT \textrm{cos}(\phi_\rmT) - v_\rmH \textrm{cos}(\phi_\rmH))\\
    \ddot{h}_{y,\textrm{L}}(\alpha_0,\delta,c_\rmb) &= \Big(\textrm{cos}(\phi_\rmH)+2c_\rmb x_\rmT \textrm{sin}(\phi_\rmH)\Big)\frac{v_\rmH^2\delta}{L_\rmH} + \Big(\textrm{sin}(\phi_\rmH)-2c_\rmb x_\rmT \textrm{cos}(\phi_\rmH)\Big)g\alpha_0 \\& \quad + 2c_\rmb\Big(v_\rmT \textrm{cos}(\phi_\rmT) - v_\rmH \textrm{cos}(\phi_\rmH)\Big)^2,
\end{aligned}
\end{equation}
and
\begin{equation}
\begin{aligned}
    h_{y,\textrm{R}}(c_\rmb) &= y_\rmT - d_{y,\textrm{min}} + c_\rmb x_\rmT^2, \\
    \dot{h}_{y,\textrm{R}}(c_\rmb) &= v_\rmT \textrm{sin}(\phi_\rmT) - v_\rmH \textrm{sin}(\phi_\rmH) + 2c_\rmb x_\rmT(v_\rmT \textrm{cos}(\phi_\rmT) - v_\rmH \textrm{cos}(\phi_\rmH)), \\
    \ddot{h}_{y,\textrm{R}}(\alpha_0,\delta,c_\rmb) &= \Big(2c_\rmb x_\rmT \textrm{sin}(\phi_\rmH) -\textrm{cos}(\phi_\rmH)\Big)\frac{v_\rmH^2\delta}{L_\rmH}+\Big(-\textrm{sin}(\phi_\rmH)-2c_\rmb x_\rmT \textrm{cos}(\phi_\rmH)\Big)g\alpha_0 \\& \quad + 2c_\rmb\Big(v_\rmT \textrm{cos}(\phi_\rmT) - v_\rmH \textrm{cos}(\phi_\rmH)\Big)^2
\end{aligned}
\end{equation}
\subsubsection{Road Keeping Barriers}
We also use road keeping barriers that prevent the vehicle from veering off the drivable area while avoiding a collision whenever possible
\begin{equation}
\begin{aligned}
    h_{\textrm{RK}} = \matl{c} 3 w_\rml - \frac{W_\rmH}{2} - y_\rmH \\ y_\rmH - \frac{W_\rmH}{2} \matr, \quad
    \dot{h}_{\textrm{RK}} = \matl{c} -v_\rmH \phi_\rmH \\ v_\rmH \phi_\rmH \matr, \quad
    \ddot{h}_{\textrm{RK}}(\delta) = \matl{c} \frac{-v_\rmH^2\textrm{cos}(\phi_\rmH)\delta}{L_\rmH} \\ \frac{v_\rmH^2\textrm{cos}(\phi_\rmH)\delta}{L_\rmH} \matr
\end{aligned}
\label{eq:LKBarrier}
\end{equation}
where $y_\rmH$ is the $y$-coordinate of the ego vehicle w.r.t. a lane attached frame where the y-coordinate of the right most lane line is $0$, $L_\rmH$ is the length of the ego vehicle, and $w_\rml$ is the lane width.

\subsection{Contextual Selection of Decoupled CBF}
In this section, we describe a method that takes advantage of the structure and rules of the road and knowledge of steering/braking efficacy for different situations to select which barrier constraints to enforce. We assume $\alpha_\rmS = \alpha_{0} + \alpha_{\rmC\rmB\rmF}$ where $\alpha_{0}$ is the nominal longitudinal acceleration that control provided by the virtual driver and $\alpha_{\rmC\rmB\rmF}$ is the correction computed by the collision avoidance algorithm. Similarly $\delta_\rmS = \delta_{0} + \delta_{\rmC\rmB\rmF}$ where $\delta_{0}$ is the nominal front wheel angle and $\delta_{\rmC\rmB\rmF}$ is the correction computed by the collision avoidance algorithm. The goal is then to calculate $\alpha_{\rmC\rmB\rmF}$ and $\delta_{\rmC\rmB\rmF}$ given $\alpha_0$ and $\delta_0$ so that the constraints are not violated.

\subsubsection{Provisional Selection of Decoupled CBF}
The algorithm has two primary mechanisms for determining whether to use CBF constraints for a target vehicle. The first mechanism is to use the below constraints to determine whether a target vehicle is a threat to ego when only using the nominal acceleration and front wheel angle proposed by the DRL Agent
\begin{equation}
\begin{aligned}
C_x &= \begin{cases}
    \dot{h}_{x,\rmF}(\alpha_0) + l_{0,x} h_{x,\rmF} \ge 0,& \text{if } x_\rmT\geq 0\\
    \ddot{h}_{x,\rmR}(\alpha_0) + l_{1,x}\dot{h}_{x,\rmR} + l_{0,x} h_{x,\rmR} \ge 0,              & \text{otherwise}
\end{cases}\\
C_y &= \ddot{h}_{y}(\alpha_0,\delta_0,0) + l_{1,y}\dot{h}_{y}(0) + l_{0,y} h_{y}(0) \ge 0.
\end{aligned}
\label{eq:barrierconstraints}
\end{equation}
These constraints arise out of the longitudinal and lateral CBF. We use the first order longitudinal barrier constraint if $x_\rmT \ge 0$ and the second order barrier constraint if $x_\rmT < 0$ because $h_{x,\rmF}$ is relative degree 1 with respect to $\alpha$ and $h_{x,\rmR}$ is relative degree 2. In (\ref{eq:barrierconstraints}), the threat assessment is performed with $\alpha_0$, $\delta_0$, and $c_b=0$.

\begin{figure}[!ht]
\centering
\includegraphics[width = 0.75\linewidth]{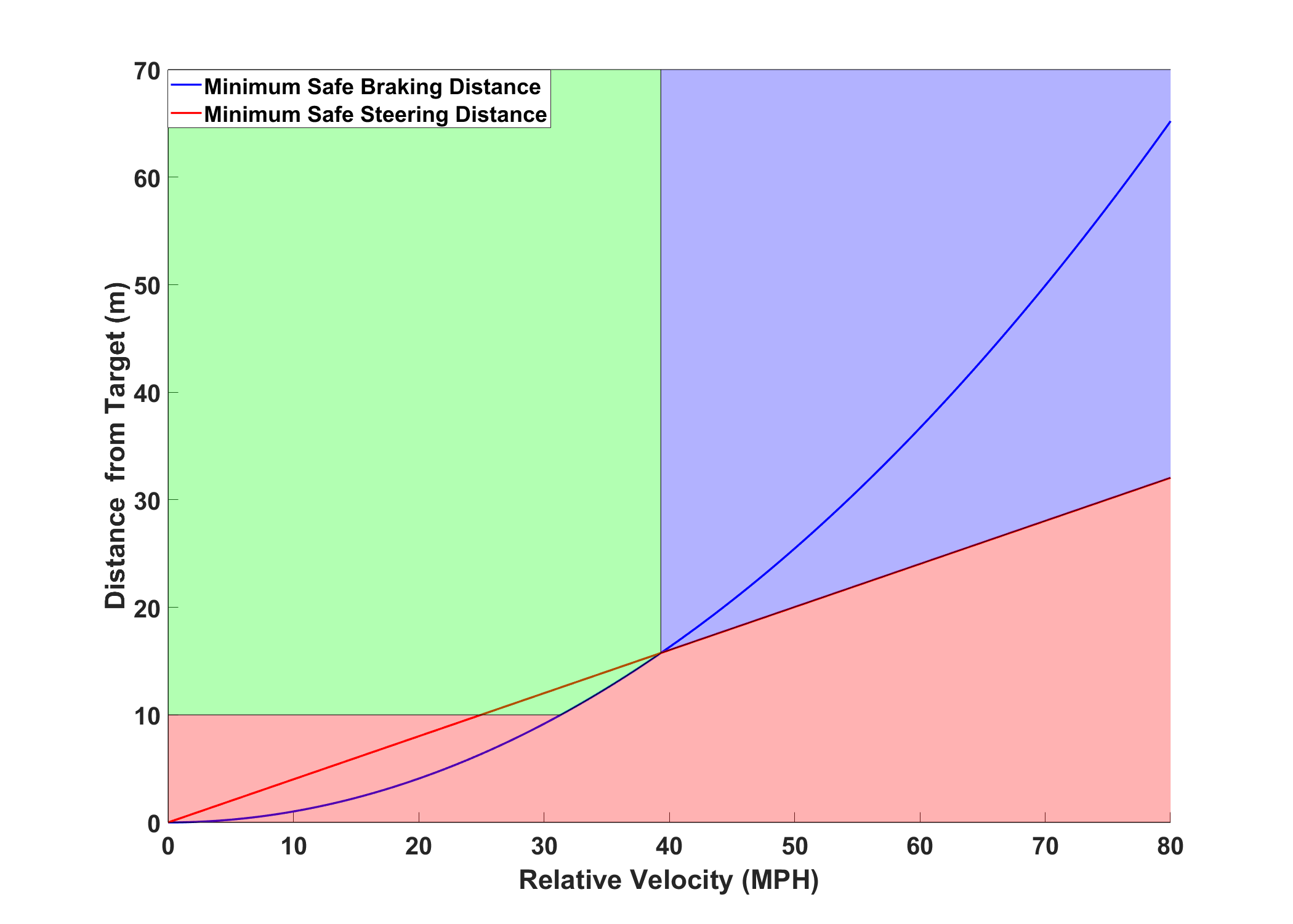}
\caption{Initial Selection of Decoupled CBF.}
\label{fig:csd_decision_plot_OR}
\end{figure}

\begin{figure}[!ht]
\centering
\includegraphics[width = 0.7\linewidth]{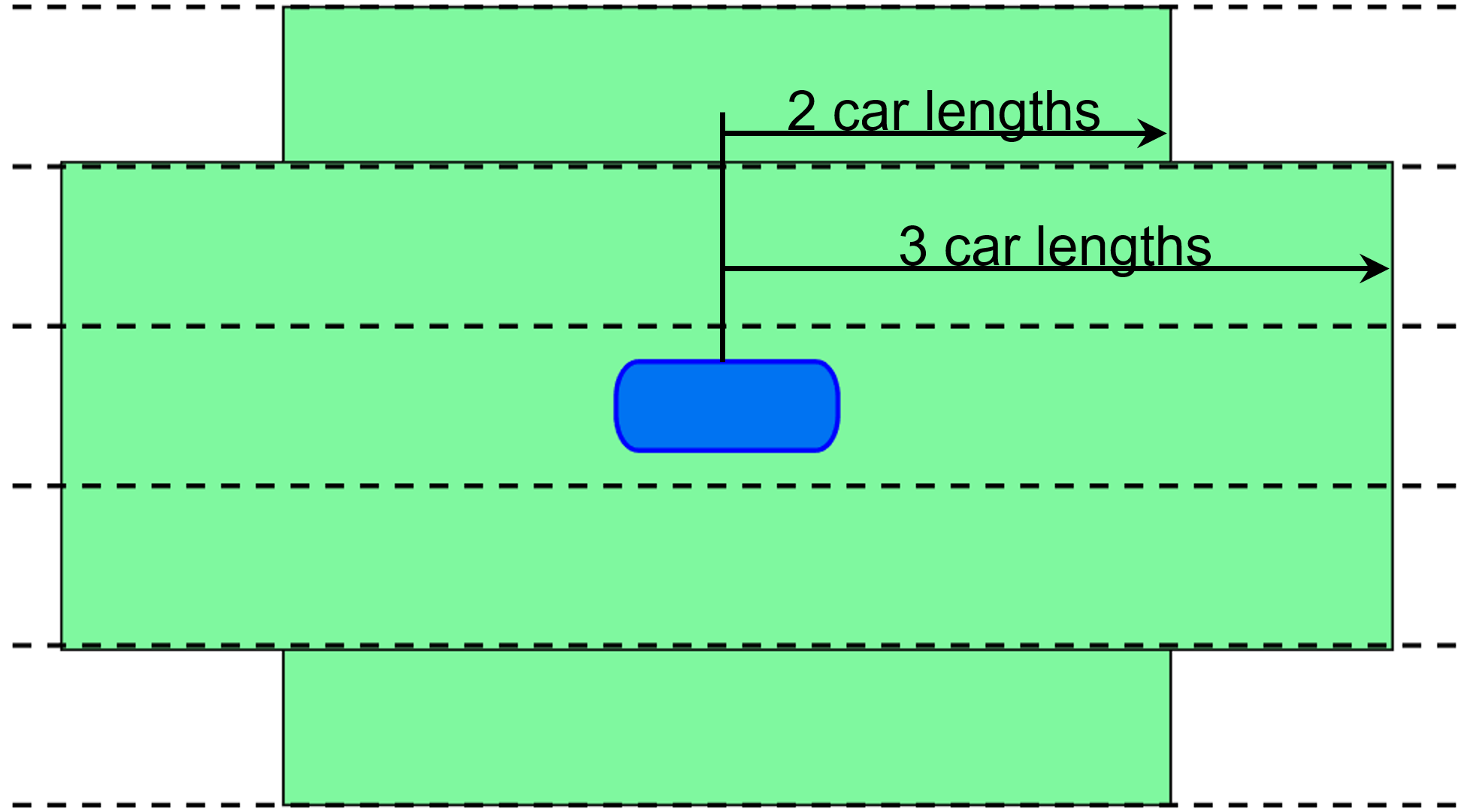}
\caption{Geometric interpretation of preemptive lateral barriers.}
\label{fig:dec2}
\end{figure}

Once a vehicle has been determined as a threat, the next step is to determine whether to brake, steer, or to do both. This is determined by the relative distance and velocity of the obstacle as shown in Fig. \ref{fig:csd_decision_plot_OR}. The $x$-axis shows the relative velocity and the $y$-axis shows the distance from the target. To arbitrate between steering and braking to avoid a collision, we use the the physical limits of the vehicle to determine whether to steer around and obstacle or to brake ahead of it. The minimum distances at which the ego vehicle can brake ahead of a target $d_\rmb$ and at which the ego vehicle can steer around a target $d_\rms$ are given by
\begin{align}
d_\rmb = -\frac{v_\rmR^2}{2\textrm{dec}_{\textrm{max}}}, \quad d_\rms = \sqrt{\frac{2d_{y,\textrm{min}}}{a_{y,\textrm{max}}}}v_\rmR, \notag
\end{align}
where $v_\rmR = v_\rmH - v_\rmT$, $\textrm{dec}_{\textrm{max}}$ is the maximum deceleration and $a_{y,\textrm{max}}$ is the maximum lateral acceleration of the ego vehicle. We define $v_{\textrm{crit}}$ as the relative velocity where $d_\rmb = d_\rms$. In the green shaded area we choose to brake, in the blue shaded area we choose to steer, and in the red shaded area we choose to do both. The boundary between the green and blue regions is determined by $v_{\textrm{crit}}$.
\vspace{-0em}
\vspace{-0em}
The second mechanism that determines whether to use CBF constraints is based on whether target vehicle is in a lane adjacent to the ego vehicle and within a predefined longitudinal threshold distance. We preemptively apply the lateral barriers for vehicles that are in either:
\begin{itemize}
\item the ego lane,
\item the lane immediately adjacent to the ego lane and within 3 car lengths from the center of the ego vehicle,
\item the lane two lanes adjacent to the ego lane and within 2 car lengths from the center of the ego vehicle.
\end{itemize}
Fig. \ref{fig:dec2} shows a geometric interpretation of the logic. If the center of a target vehicle falls within the shaded area, the appropriate lateral barrier is enforced.

\begin{figure}[!ht]
\centering
\includegraphics[width = 1\linewidth]{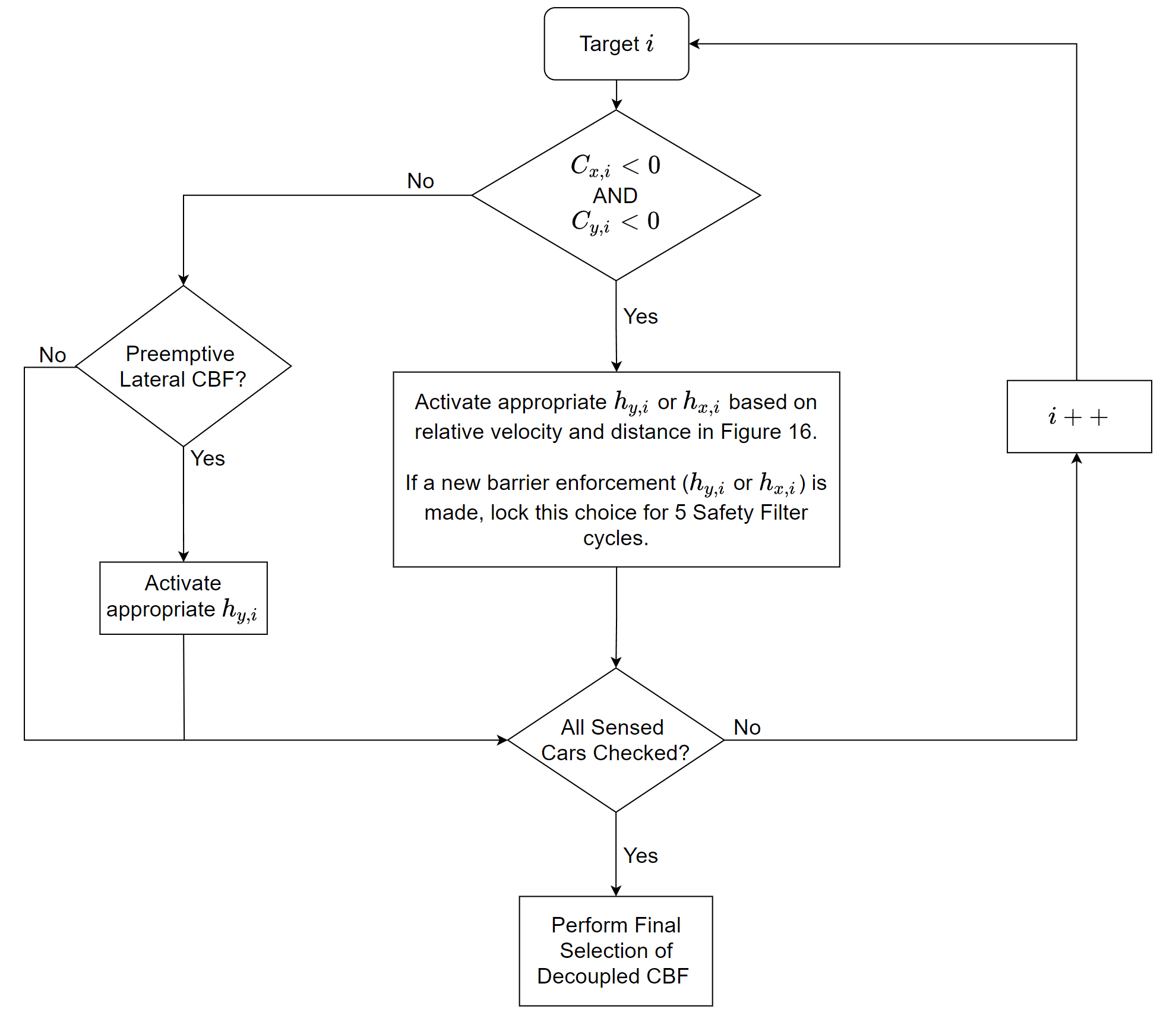}
\caption{Initial Selection of Decoupled CBF.}
\label{fig:flowchartA}
\end{figure}

Using these two mechanisms, the barriers to be enforced are provisionally selected, as shown in Fig. \ref{fig:flowchartA}. The process shown in Fig. \ref{fig:flowchartA} is performed for the closest front and rear targets in each lane. By considering all the vehicles nearby the ego, the collision avoidance algorithm can prevent a broad range of collisions and is not limited to forward collision avoidance. This barrier selection is provisional because in the case the selection cannot be enforced due to conflicting constraints, the algorithm systematically replaces lateral constraints with longitudinal constraints until a feasible solution is found.

Once the barriers have been provisionally selected, we designate the threat (i.e. barrier enforced through the threat assessment mechanism) with the minimum longitudinal distance as the \textit{primary obstacle} $i_{\rmP\rmO}$. For the primary obstacle, we provisionally activate both the left and right lateral barriers. This results in 2 distinct options, going to the right or the left of the primary obstacle. For all other lateral barriers, we choose the left or right barrier based on the location of the target w.r.t. the ego vehicle.

\subsubsection{Computing $\alpha_{_{\textrm{CBF}}}$ and $\delta_{_{\textrm{CBF}}}$ with Quadratic Programs}
\begin{figure}[!ht]
\centering
\includegraphics[width = 0.7\linewidth]{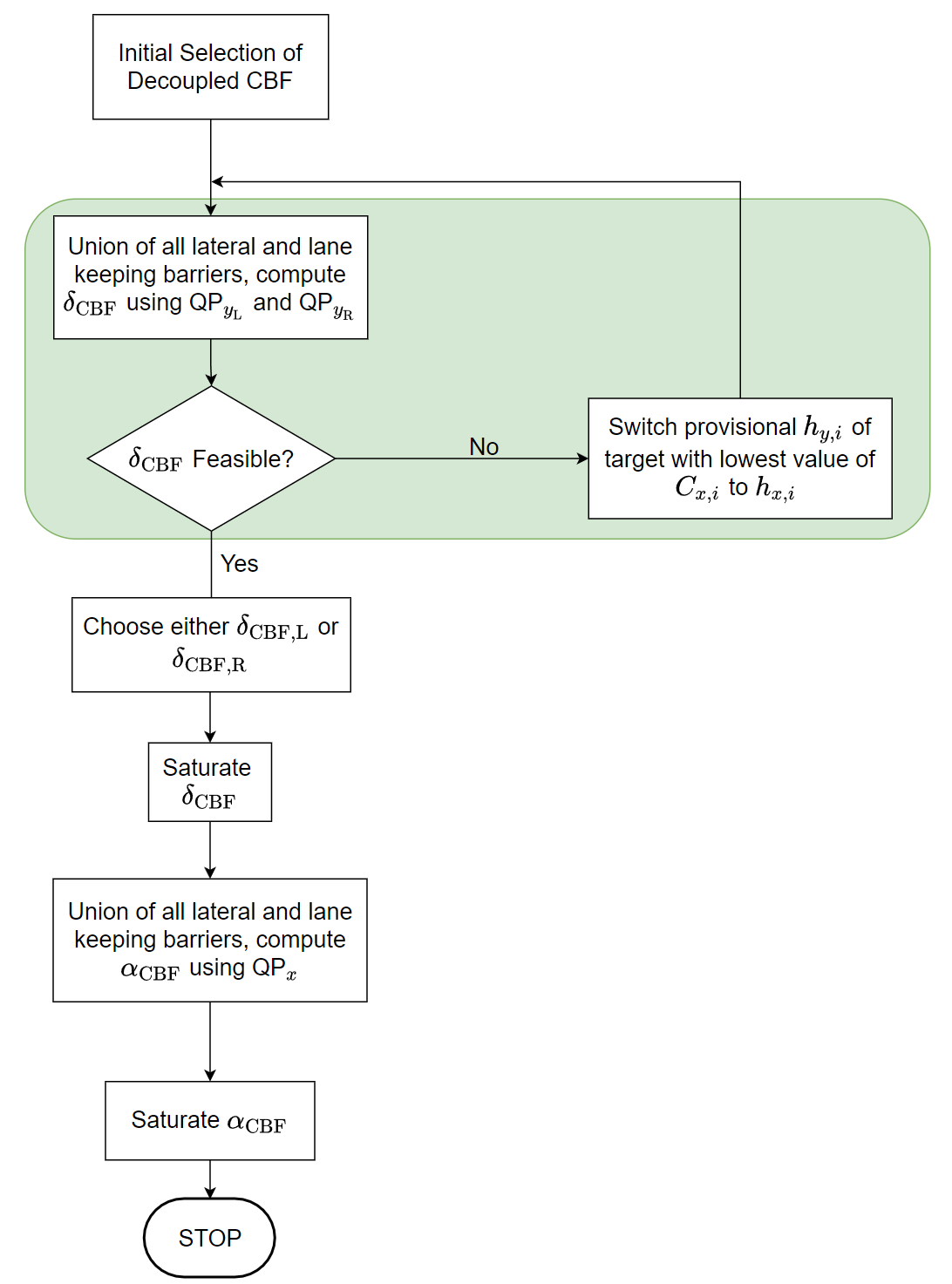}
\caption{Final Selection of Decoupled CBF. The steps in the shaded area are performed for both QP$_{y_\rmL}$ and QP$_{y_\rmR}$.}
\label{fig:flowchartD}
\end{figure}

We calculate the safe longitudinal acceleration $\alpha_\rmS = \alpha_{_{\textrm{CBF}}} + \alpha_{0}$ and the safe front wheel angle $\delta_\rmS = \delta_{_{\textrm{CBF}}} + \delta_{0}$ using 3 quadratic programs, QP$_{y_\rmL}$, QP$_{y_\rmR}$ and QP$_x$.
\paragraph{QP$_{y_\rmL}$ and QP$_{y_\rmR}$:}
We define QP$_{y_\rmL}$ as
\begin{gather}
\argmin_{\delta_{_{\textrm{CBF,L}}}, s_{\textrm{RK}}, s_{\rms\rma\rmt}} \matl{ccc} \delta_{_{\textrm{CBF,L}}}  & s_{\textrm{RK}} & s_{\rms\rma\rmt} \matr Q_y \matl{c} \delta_{_{\textrm{CBF,L}}}  \\ s_{\textrm{RK}} \\ s_{\rms\rma\rmt} \matr,
\end{gather}
subject to:
\begin{gather}
	\ddot{h}_{y,\rmL,i_{\rmP\rmO}}(\alpha_0,\delta_0+\delta_{_{\textrm{CBF,L}}}) + l_{1,y}\dot{h}_{y,\rmL,i_{\rmP\rmO}} + l_{0,y} h_{y,\rmL,i_{\rmP\rmO}} \ge 0, \notag \\
	\ddot{h}_{y,i}(\alpha_0,\delta_0+\delta_{_{\textrm{CBF,L}}}) + l_{1,y}\dot{h}_{y,i} + l_{0,y} h_{y,i} \ge 0, \quad  \forall i \in \SY, \\
	\ddot{h}_{\textrm{RK}}(\delta_0+\delta_{_{\textrm{CBF,L}}}) + l_{1,\textrm{RK}}\dot{h}_{\textrm{RK}} + l_{0,\textrm{RK}} h_{\textrm{RK}} + \matl{c} 1 \\ 1 \matr s_{\textrm{RK}} \ge 0, \notag \\	
    \delta_{\textrm{min}} - \delta_{0} \le \delta_{_{\textrm{CBF,L}}} + s_{\rms\rma\rmt}, \notag\\
	\delta_{_{\textrm{CBF,L}}} - s_{\rms\rma\rmt} \le \delta_{\textrm{max}} - \delta_{0}, \notag
\end{gather}
where $\SY$ is the set of target ID's for which the lateral barrier is enforced excluding the primary obstacle, and $l_{1,y}$, $l_{0,y}$, $l_{1,\textrm{RK}}$, and $l_{0,\textrm{RK}}$ are tunable parameters that determine the sensitivity of the barriers. We define $J_{y_\rmL}$ as the optimal quadratic cost of QP$_{y_\rmL}$. Similarly, we define QP$_{y_\rmR}$ as
\begin{gather}
\argmin_{\delta_{_{\textrm{CBF,R}}}, s_{\textrm{RK}}, s_{\rms\rma\rmt}} = \matl{ccc} \delta_{_{\textrm{CBF,R}}}  & s_{\rmR\rmK} & s_{\rms\rma\rmt} \matr Q_y \matl{c} \delta_{_{\textrm{CBF,R}}}  \\ s_{\rmR\rmK} \\ s_{\rms\rma\rmt} \matr,
\end{gather}
subject to:
\begin{gather}
	\ddot{h}_{y,\rmR,i_{\rmP\rmO}}(\alpha_0,\delta_0+\delta_{_{\textrm{CBF,R}}}) + l_{1,y}\dot{h}_{y,\rmR,i_{\rmP\rmO}} + l_{0,y} h_{y,\rmR,i_{\rmP\rmO}} \ge 0, \notag \\
	\ddot{h}_{y,i}(\alpha_0,\delta_0+\delta_{_{\textrm{CBF,R}}}) + l_{1,y}\dot{h}_{y,i} + l_{0,y} h_{y,i} \ge 0, \quad  \forall i \in \SY, \\
	\ddot{h}_{\textrm{RK}}(\delta_0+\delta_{_{\textrm{CBF,R}}}) + l_{1,\textrm{RK}}\dot{h}_{\textrm{RK}} + l_{0,\textrm{RK}} h_{\textrm{RK}} + \matl{c} 1 \\ 1 \matr s_{\rmR\rmK} \ge 0, \notag\\	
    \delta_{\textrm{min}} - \delta_{0} \le \delta_{_{\textrm{CBF,R}}} + s_{\rms\rma\rmt}, \notag\\
	\delta_{_{\textrm{CBF,R}}} - s_{\rms\rma\rmt} \le \delta_{\textrm{max}} - \delta_{0},\notag
\end{gather}
and we define $J_{y_\rmR}$ as the optimal quadratic cost of QP$_{y_\rmR}$. The distinction between QP$_{y_\rmL}$ and QP$_{y_\rmR}$ is that in QP$_{y_\rmL}$, the left barrier is enforced for the primary obstacle and in QP$_{y_\rmR}$ the right barrier is enforced for the primary obstacle. The slack variable $s_{\rmR\rmK}$ allows the algorithm to momentarily leave the road surface if such an action is necessary to avoid a collision and return when it is safe to do so. Similarly, $s_{\rms\rma\rmt}$ allows the algorithm to find a solution to the QPs even in the event of actuator saturation.
\paragraph{Final Selection of Lateral CBF:} \label{ProcessD}
The procedure used for the final selection of the CBFs is shown in Fig. \ref{fig:flowchartD}. After the provisional selection of the CBFs, we attempt to solve QP$_{y_\rmL}$ and QP$_{y_\rmR}$. We define one or both QP$_{y_\rmL}$ and QP$_{y_\rmR}$ as infeasible if any of the lateral constraints conflict (one lateral constraint requires $\delta > 0$ and another requires $\delta < 0$). If either QP$_{y_\rmL}$ or QP$_{y_\rmR}$ is infeasible, we replace the provisional $h_{y,i}$ of the target with the lowest value of $C_{x,i}$ with the longitudinal CBF $h_{x,i}$, and resolve. This procedure is repeated until either a solution is found or all the provisional lateral barriers have been replaced by a longitudinal barrier. Note that both QP$_{y_\rmL}$ and QP$_{y_\rmR}$ may still be infeasible after all provisional lateral barriers have been replaced. Once QP$_{y_\rmL}$ and QP$_{y_\rmR}$ have either been solved or deemed infeasible, we select $\delta_{_{\textrm{CBF}}}$ based on the following criteria.
\begin{itemize}
\item If QP$_{y_\rmL}$ is infeasible, $\delta_{_{\textrm{CBF}}} = \delta_{_{\textrm{CBF,R}}}$
\item If QP$_{y_\rmR}$ is infeasible, $\delta_{_{\textrm{CBF}}} = \delta_{_{\textrm{CBF,L}}}$
\item If QP$_{y_\rmL}$ and QP$_{y_\rmR}$ are both feasible,
\begin{itemize}
\item if $|J_{y_\rmL}-J_{y_\rmR}| < 1e-5$, $\delta_{_{\textrm{CBF}}}$ = $\delta_{_{\textrm{CBF,L}}}$,
\item else if $J_{y_\rmL} < J_{y_\rmR}$ AND ($\delta_{_{\textrm{CBF,R}}}$ was NOT selected in the last step OR $J_{y_\rmL} < \frac{J_{y_\rmR}}{2}$), $\delta_{_{\textrm{CBF}}}$ = $\delta_{_{\textrm{CBF,L}}}$,
\item else if $J_{y_\rmR} < J_{y_\rmL}$ AND ($\delta_{_{\textrm{CBF,L}}}$ was NOT selected in the last step OR $J_{y_\rmR} < \frac{J_{y_\rmL}}{2}$), $\delta_{_{\textrm{CBF}}}$ = $\delta_{_{\textrm{CBF,R}}}$,
\item else if $\delta_{_{\textrm{CBF,L}}}$ was selected in the last step, $\delta_{_{\textrm{CBF}}}$ = $\delta_{_{\textrm{CBF,L}}}$,
\item else if $\delta_{_{\textrm{CBF,R}}}$ was selected in the last step, $\delta_{_{\textrm{CBF}}}$ = $\delta_{_{\textrm{CBF,R}}}$,
\end{itemize}
\item If QP$_{y_\rmL}$ and QP$_{y_\rmR}$ are both infeasible, $\delta_{_{\textrm{CBF}}}$ is infeasible. Skip \ref{par:QPx} and perform max braking.
\end{itemize}
The logic above chooses the cheaper option and also prevents switching between choosing left or right in sequential steps. $\delta_{_{\textrm{CBF}}}$ is saturated after its computation by the dual QPs.

\paragraph{QP$_x$:} \label{par:QPx}
We solve QP$_x$ defined below, and saturate $\alpha_{_{\textrm{CBF}}}$ based on the actuation limits of the ego vehicle.
\begin{gather}
\alpha_{_{\textrm{CBF}}} = \argmin_{\alpha_{_{\textrm{CBF}}}} \alpha_{_{\textrm{CBF}}}^2,
\end{gather}
subject to:
\begin{gather}
\dot{h}_{x,i}(\alpha_0+\alpha_{_{\textrm{CBF}}}) + l_{0,x} h_{x,i}, \ge 0 \quad  \forall i \in \SX,	
\end{gather}
where $\SX$ is the set of target ID's for which the longitudinal barrier is enforced and $l_{0,x}$ is a tunable parameter that determine the sensitivity of the barriers. For each target $i$, we use $h_{x,\rmF}$ if $x_\rmT >= 0$ and $h_{x,\rmR}$ if $x_\rmT < 0$. If QP$_x$ is infeasible due to conflicting constraints, we remove the conflicting rear constraints and resolve.

\subsection{Examples}
In all examples shown in this section, the parameters are set to the values shown in Table \ref{tab:tuning}.


\begin{table}[!ht]
\centering
\caption{Tuning Parameters}
\begin{tabular}{SSSSSSSSSS} \toprule
    {Parameter} & {$\, k_v (\rms) \,$} & {$\,d_{x,\textrm{min}} (\rmm)\,$} & {$\,d_{y,\textrm{min}} (\rmm)\,$} & {$\,c_\rmb\,$} & {$\,l_{1,y}\,$} & {$\,l_{1,y}\,$} & {$\,l_{1,\textrm{RK}}\,$} & {$\,l_{0,\textrm{RK}}\,$} & {$\,l_{0,x}\,$} \\ \midrule
    {Value} & {1}     & {6}                    & {3.15}                 & {0.0025}   & {7}         & {10}        & {7}                   & {10}                  & {$2\sqrt{\frac{0.4g}{|x_\rmT|}}$} \\ \bottomrule
\end{tabular}
\label{tab:tuning}
\end{table}

In the position plots shown below (Figures \ref{fig:ex1hist}, \ref{fig:ex2hist}, and \ref{fig:ex3hist}), the green vehicle is the ego with the collision avoidance activated and the dotted pink is the ego vehicle without the collision avoidance algorithm. All target vehicles are shown in blue. The dotted red lines show the active constraints at that time step.
\subsubsection{Aggressive/Erratic Ego Cut-in Prevention} \label{ex1}
\begin{figure}[!ht]
\begin{subfigure}{1\textwidth}
  \centering
  \includegraphics[width=1\linewidth]{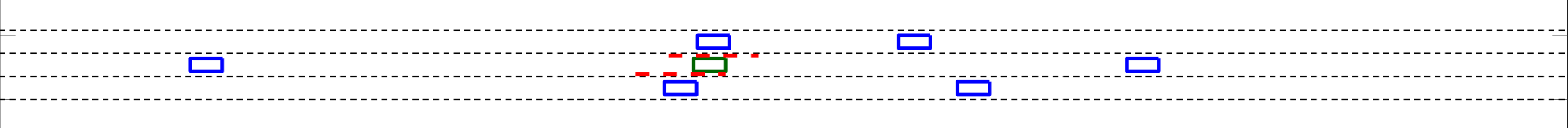}
  \caption{$t = 0$ sec}
  \label{fig:sfig11}
\end{subfigure}
\begin{subfigure}{1\textwidth}
  \centering
  \includegraphics[width=1\linewidth]{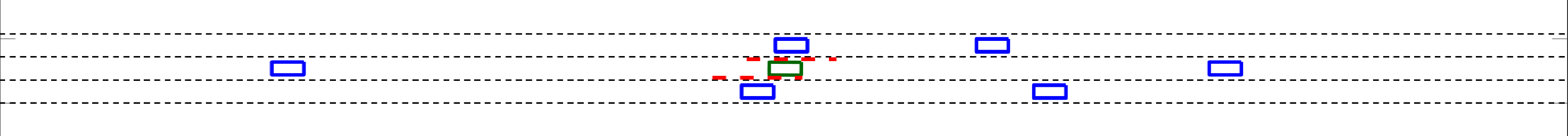}
  \caption{$t = 0.5$ sec}
  \label{fig:sfig12}
\end{subfigure}
\begin{subfigure}{1\textwidth}
  \centering
  \includegraphics[width=1\linewidth]{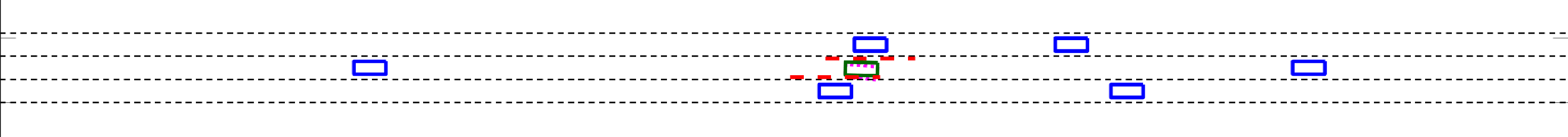}
  \caption{$t = 1$ sec}
  \label{fig:sfig13}
\end{subfigure}
\begin{subfigure}{1\textwidth}
  \centering
  \includegraphics[width=1\linewidth]{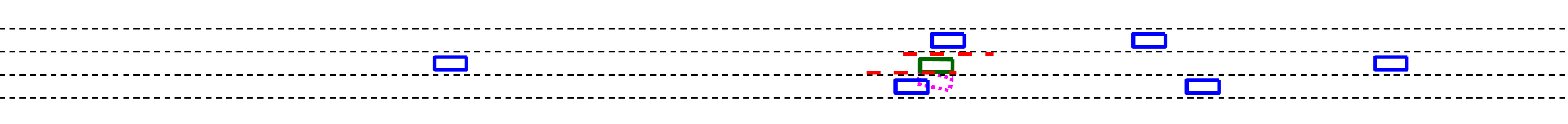}
  \caption{$t = 1.5$ sec}
  \label{fig:sfig14}
\end{subfigure}
\begin{subfigure}{1\textwidth}
  \centering
  \includegraphics[width=1\linewidth]{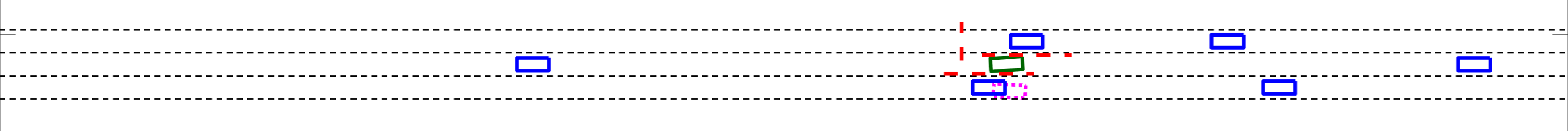}
  \caption{$t = 2$ sec}
  \label{fig:sfig15}
\end{subfigure}
\begin{subfigure}{1\textwidth}
  \centering
  \includegraphics[width=1\linewidth]{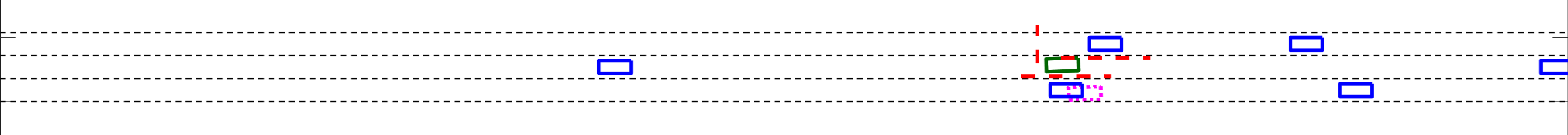}
  \caption{$t = 2.5$ sec}
  \label{fig:sfig16}
\end{subfigure}
\begin{subfigure}{1\textwidth}
  \centering
  \includegraphics[width=1\linewidth]{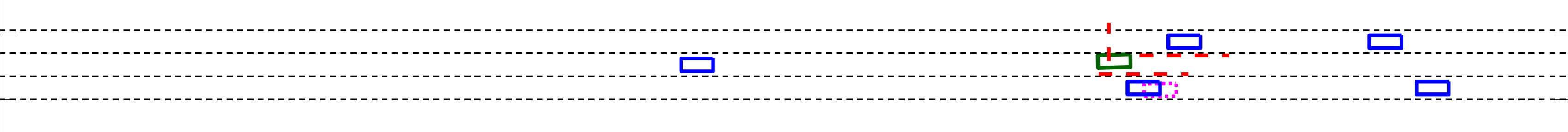}
  \caption{$t = 3$ sec}
  \label{fig:sfig17}
\end{subfigure}
\caption{Illustration of Example \ref{ex1} -- Position of the ego vehicle and the target vehicles at 0.5 sec intervals.}
\label{fig:ex1hist}
\end{figure}

\begin{figure}[!ht]
\begin{subfigure}{0.5\textwidth}
  \centering
  \includegraphics[width=1\linewidth]{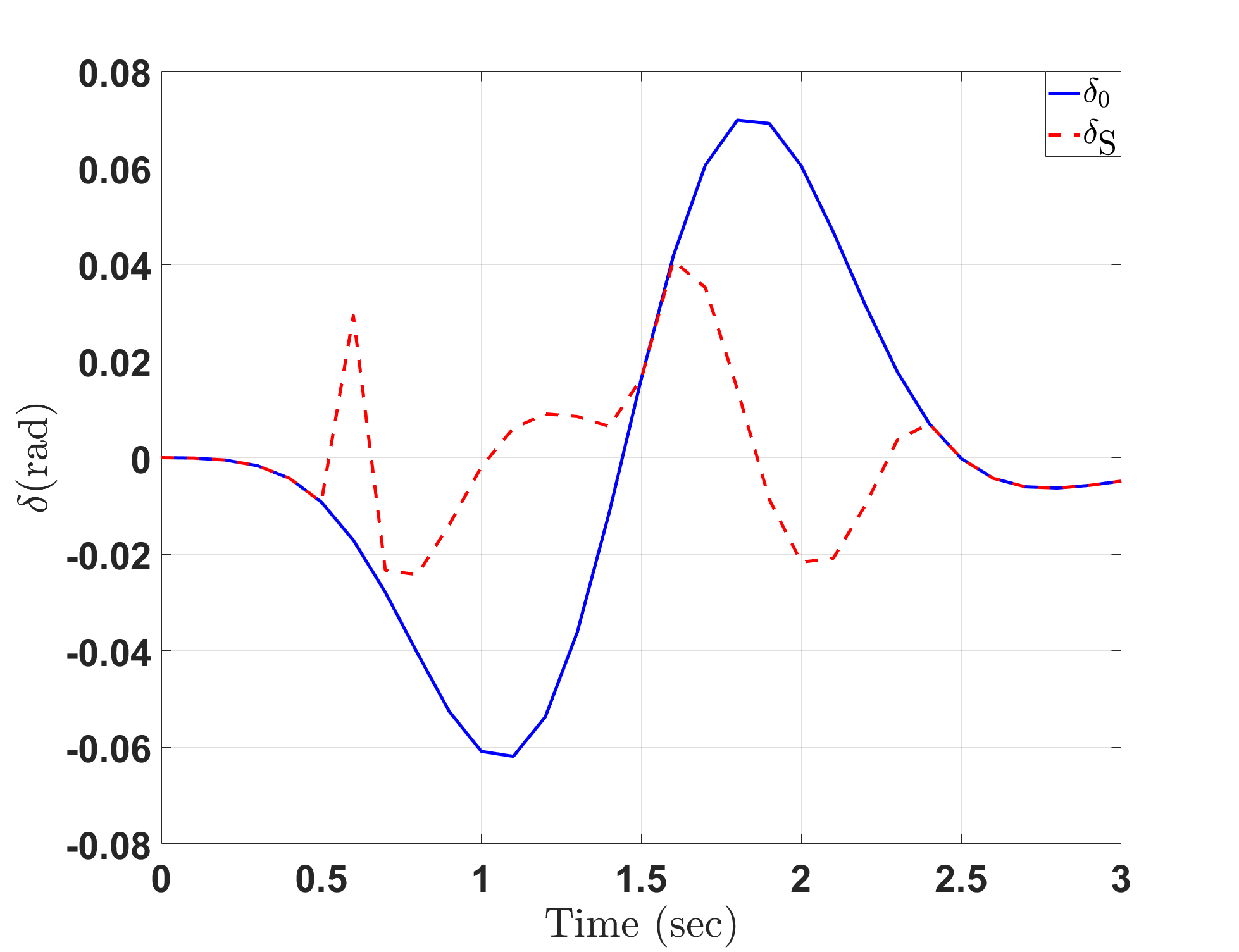}
  \caption{Front Wheel Angle.}
  \label{fig:ex1steer}
\end{subfigure}
\begin{subfigure}{0.5\textwidth}
  \centering
  \includegraphics[width=1\linewidth]{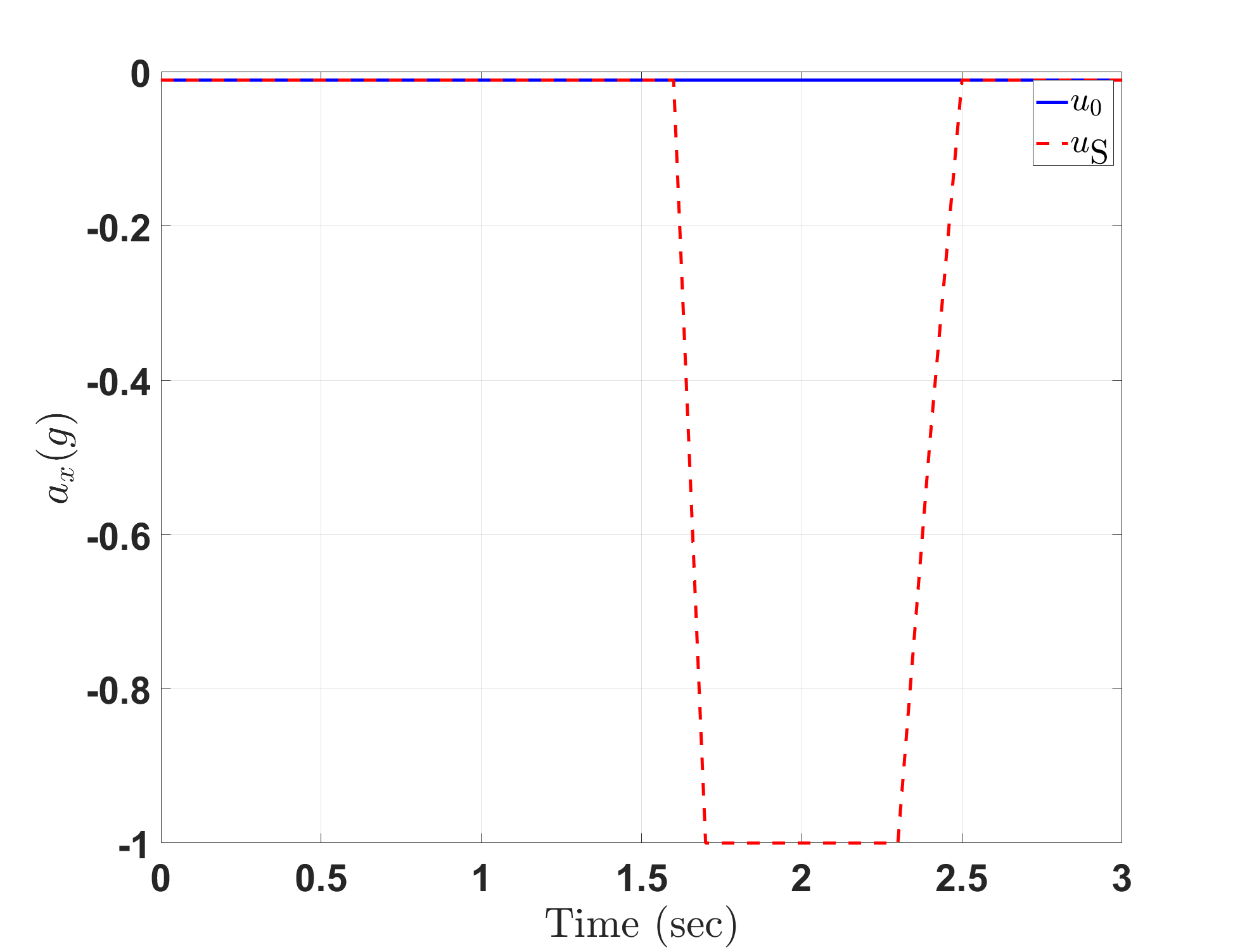}
  \caption{Longitudinal Acceleration.}
  \label{fig:ex1acc}
\end{subfigure}
\caption{Illustration of Example \ref{ex1} -- Front Wheel Angle and Longitudinal Acceleration.}
\label{fig:ex1control}
\end{figure}

\begin{figure}[!ht]
\centering
\includegraphics[width = 0.65\linewidth]{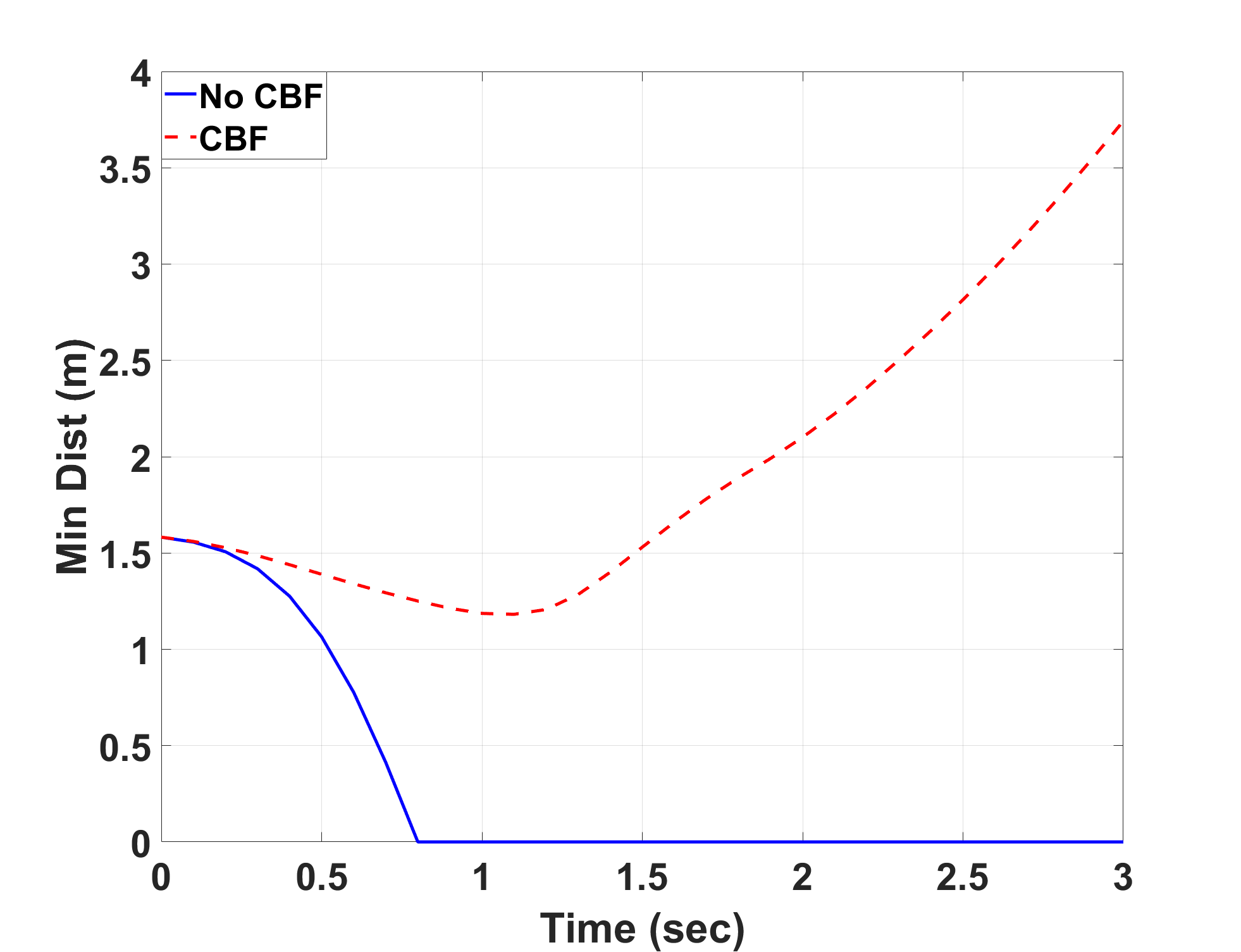}
\caption{Illustration of Example \ref{ex1} -- Minimum Euclidean distance between ego and target vehicles.}
\label{fig:ex1mindist}
\end{figure}

In this example, the ego vehicle attempts to change lanes while a vehicle is in its blind spot. Fig. \ref{fig:ex1hist} shows the position of the ego vehicle and the target vehicles at 0.5 sec intervals. Figures \ref{fig:ex1steer} and \ref{fig:ex1acc} show the front wheel angle and acceleration of the ego vehicle and Figure \ref{fig:ex1mindist} shows the minimum Euclidean distance between the ego and the critical target vehicle (i.e. the one threatening to collide with the ego vehicle) edges. The CBF safety filter corrected the nominal steering actuation and prevented a potentially erratic right lane change -- which is shown by the dashed-red rectangle in Fig. \ref{fig:ex1hist} (d)-(g).Similarly, it also prevented the ego vehicle from making a left lane change (shown by the dotted red line constraint).


\subsubsection{Target Cut-in Evasion}\label{ex2}
\begin{figure}[!ht]
\begin{subfigure}{1\textwidth}
  \centering
  \includegraphics[width=1\linewidth]{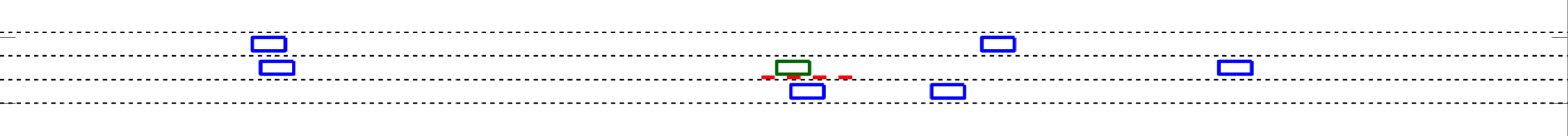}
  \caption{$t = 0$ sec}
  \label{fig:sfig21}
\end{subfigure}
\begin{subfigure}{1\textwidth}
  \centering
  \includegraphics[width=1\linewidth]{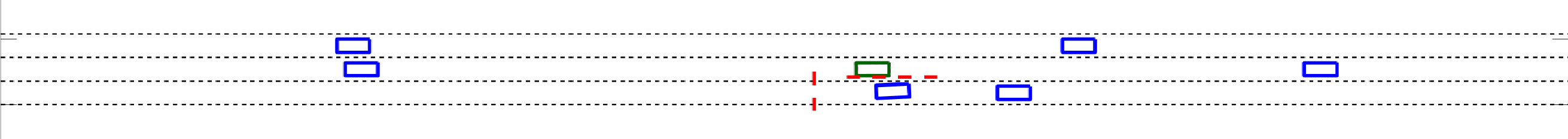}
  \caption{$t = 0.5$ sec}
  \label{fig:sfig22}
\end{subfigure}
\begin{subfigure}{1\textwidth}
  \centering
  \includegraphics[width=1\linewidth]{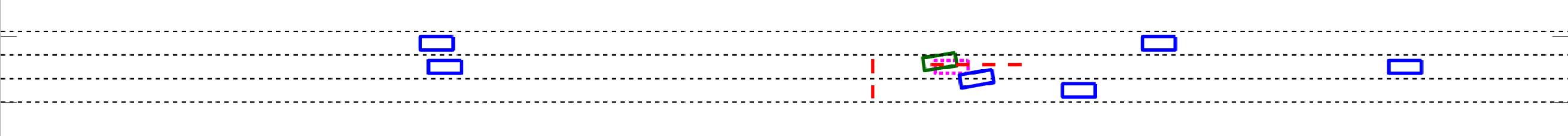}
  \caption{$t = 1$ sec}
  \label{fig:sfig23}
\end{subfigure}
\begin{subfigure}{1\textwidth}
  \centering
  \includegraphics[width=1\linewidth]{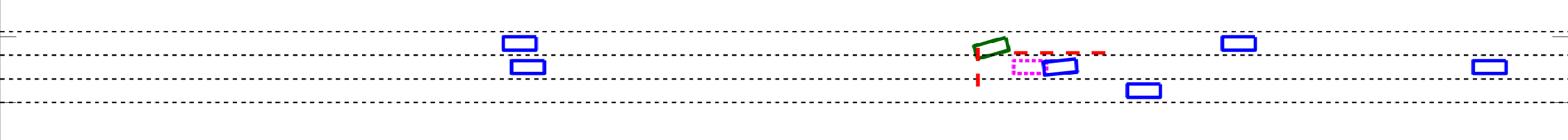}
  \caption{$t = 1.5$ sec}
  \label{fig:sfig24}
\end{subfigure}
\begin{subfigure}{1\textwidth}
  \centering
  \includegraphics[width=1\linewidth]{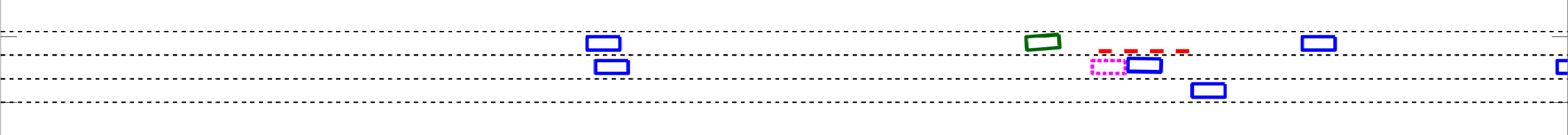}
  \caption{$t = 2$ sec}
  \label{fig:sfig25}
\end{subfigure}
\begin{subfigure}{1\textwidth}
  \centering
  \includegraphics[width=1\linewidth]{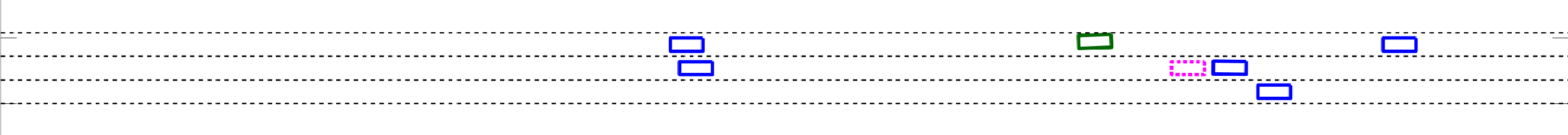}
  \caption{$t = 2.5$ sec}
  \label{fig:sfig26}
\end{subfigure}
\begin{subfigure}{1\textwidth}
  \centering
  \includegraphics[width=1\linewidth]{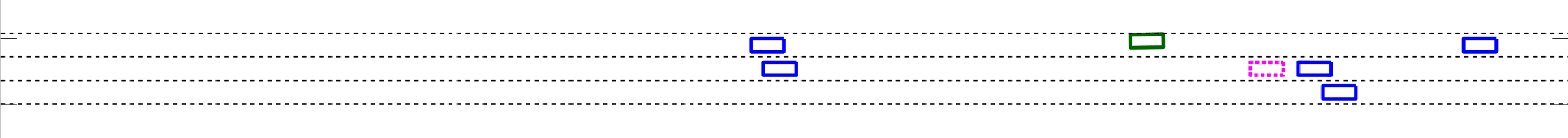}
  \caption{$t = 3$ sec}
  \label{fig:sfig27}
\end{subfigure}
\caption{Illustration of Example \ref{ex2} -- Position of the ego vehicle and the target vehicles at 0.5 sec intervals.}
\label{fig:ex2hist}
\end{figure}

\begin{figure}[!ht]
\begin{subfigure}{0.5\textwidth}
  \centering
  \includegraphics[width=1\linewidth]{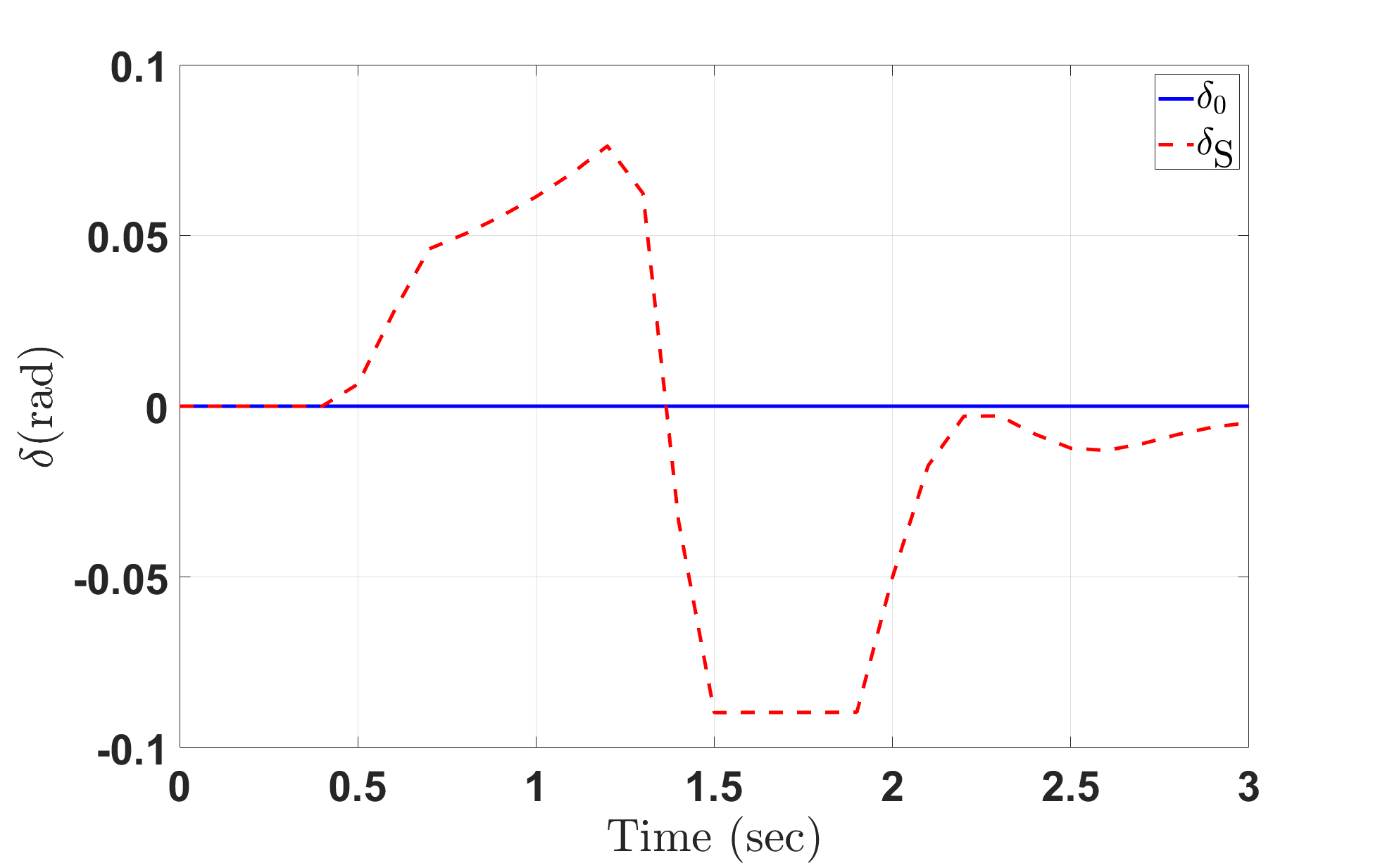}
  \caption{Front Wheel Angle.}
  \label{fig:ex2steer}
\end{subfigure}
\begin{subfigure}{0.5\textwidth}
  \centering
  \includegraphics[width=1\linewidth]{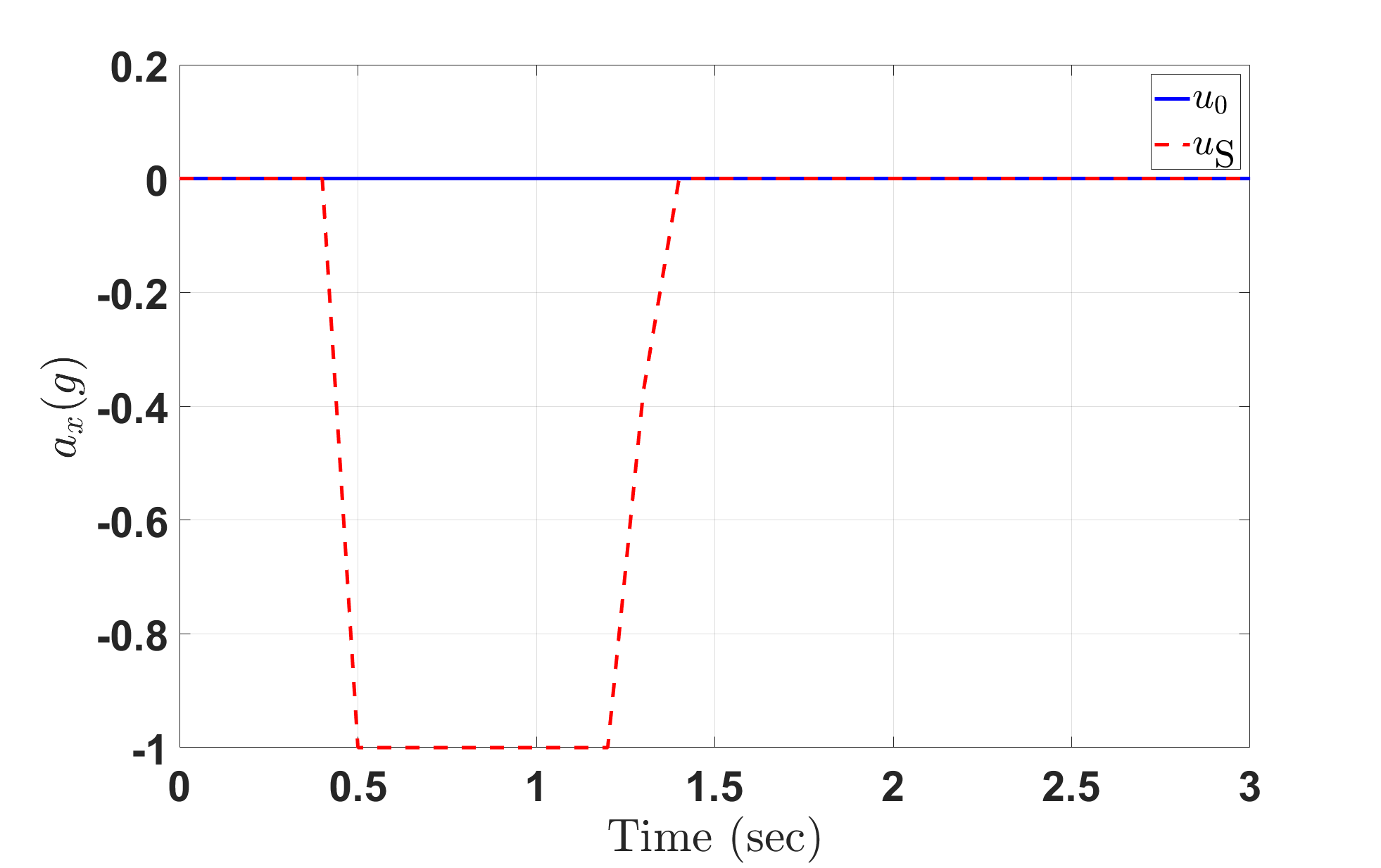}
  \caption{Longitudinal Acceleration.}
  \label{fig:ex2acc}
\end{subfigure}
\caption{Illustration of Example \ref{ex2} -- Front Wheel Angle and Longitudinal Acceleration.}
\label{fig:ex2control}
\end{figure}

\begin{figure}[!ht]
\centering
\includegraphics[width = 0.65\linewidth]{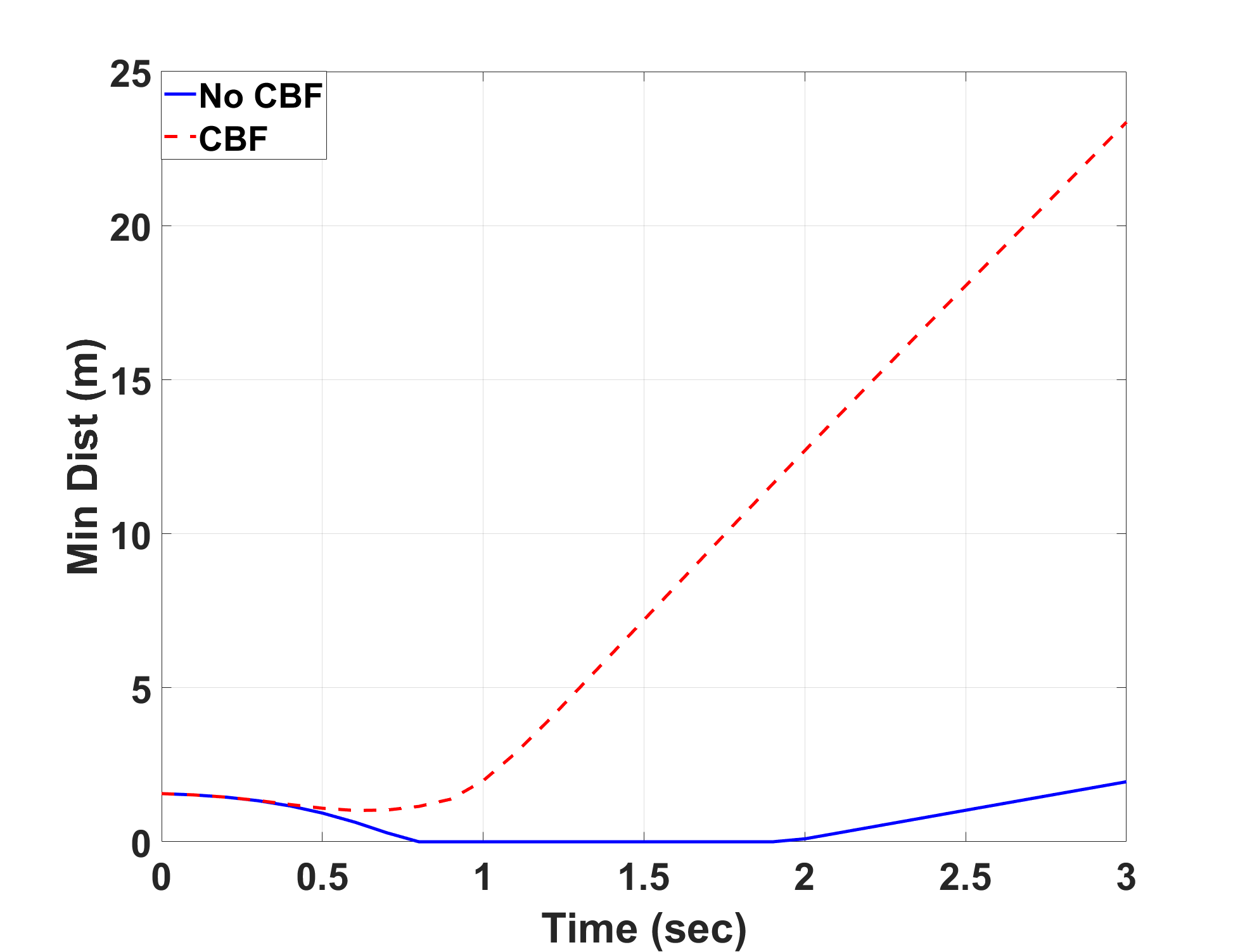}
\caption{Illustration of Example \ref{ex2} -- Minimum Euclidean distance between ego and target vehicles.}
\label{fig:ex2mindist}
\end{figure}

In this example, the ego vehicle evades a target vehicle that cuts-in ahead. Fig. \ref{fig:ex2hist} shows the position of the ego vehicle and the target vehicles at 0.5 sec intervals. Figures \ref{fig:ex2steer} and \ref{fig:ex2acc} show the front wheel angle and acceleration of the ego vehicle and Figure  \ref{fig:ex2mindist} shows the minimum Euclidean distance between the ego and the critical target vehicle edges. The safety filter avoids the vehicle cutting in by changing lanes, and also prevents the ego vehicle from driving off the highway due to the road keeping barriers.

\subsubsection{Avoiding a Stationary Vehicle Ahead} \label{ex3}
\begin{figure}[!ht]
\begin{subfigure}{1\textwidth}
  \centering
  \includegraphics[width=1\linewidth]{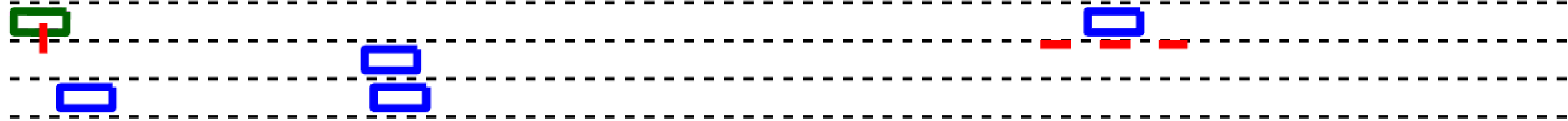}
  \caption{$t = 0$ sec}
  \label{fig:sfig31}
\end{subfigure}
\begin{subfigure}{1\textwidth}
  \centering
  \includegraphics[width=1\linewidth]{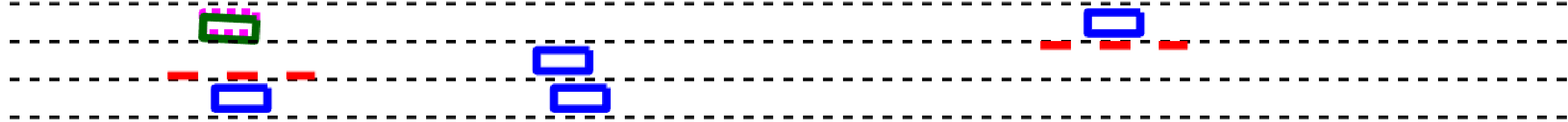}
  \caption{$t = 0.5$ sec}
  \label{fig:sfig32}
\end{subfigure}
\begin{subfigure}{1\textwidth}
  \centering
  \includegraphics[width=1\linewidth]{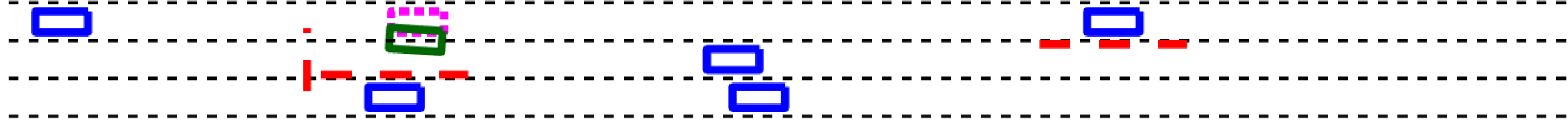}
  \caption{$t = 1$ sec}
  \label{fig:sfig33}
\end{subfigure}
\begin{subfigure}{1\textwidth}
  \centering
  \includegraphics[width=1\linewidth]{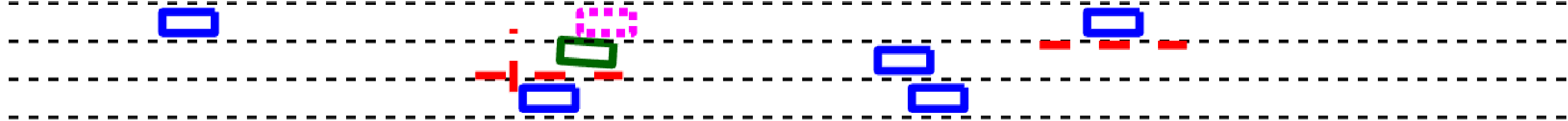}
  \caption{$t = 1.5$ sec}
  \label{fig:sfig34}
\end{subfigure}
\begin{subfigure}{1\textwidth}
  \centering
  \includegraphics[width=1\linewidth]{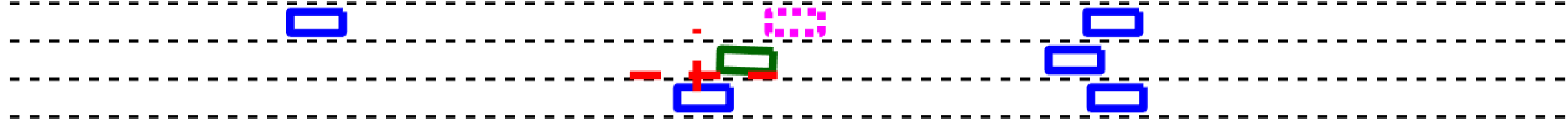}
  \caption{$t = 2$ sec}
  \label{fig:sfig35}
\end{subfigure}
\begin{subfigure}{1\textwidth}
  \centering
  \includegraphics[width=1\linewidth]{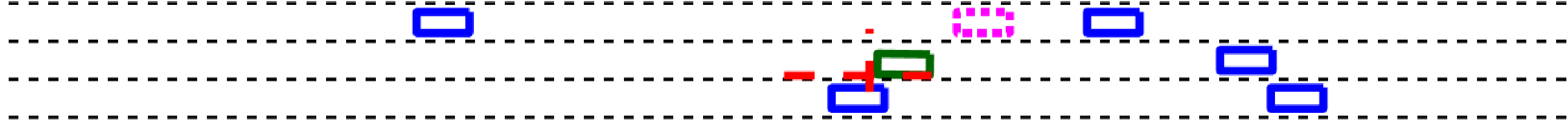}
  \caption{$t = 2.5$ sec}
  \label{fig:sfig36}
\end{subfigure}
\begin{subfigure}{1\textwidth}
  \centering
  \includegraphics[width=1\linewidth]{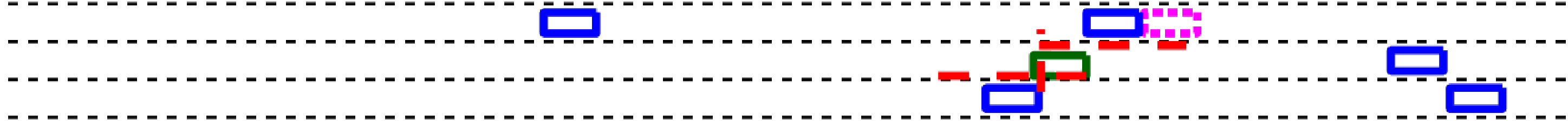}
  \caption{$t = 3$ sec}
  \label{fig:sfig37}
\end{subfigure}
\caption{Example \ref{ex3} -- Position of the ego vehicle and the target vehicles at 0.5 sec intervals.}
\label{fig:ex3hist}
\end{figure}

\begin{figure}[!ht]
\begin{subfigure}{0.5\textwidth}
  \centering
  \includegraphics[width=1\linewidth]{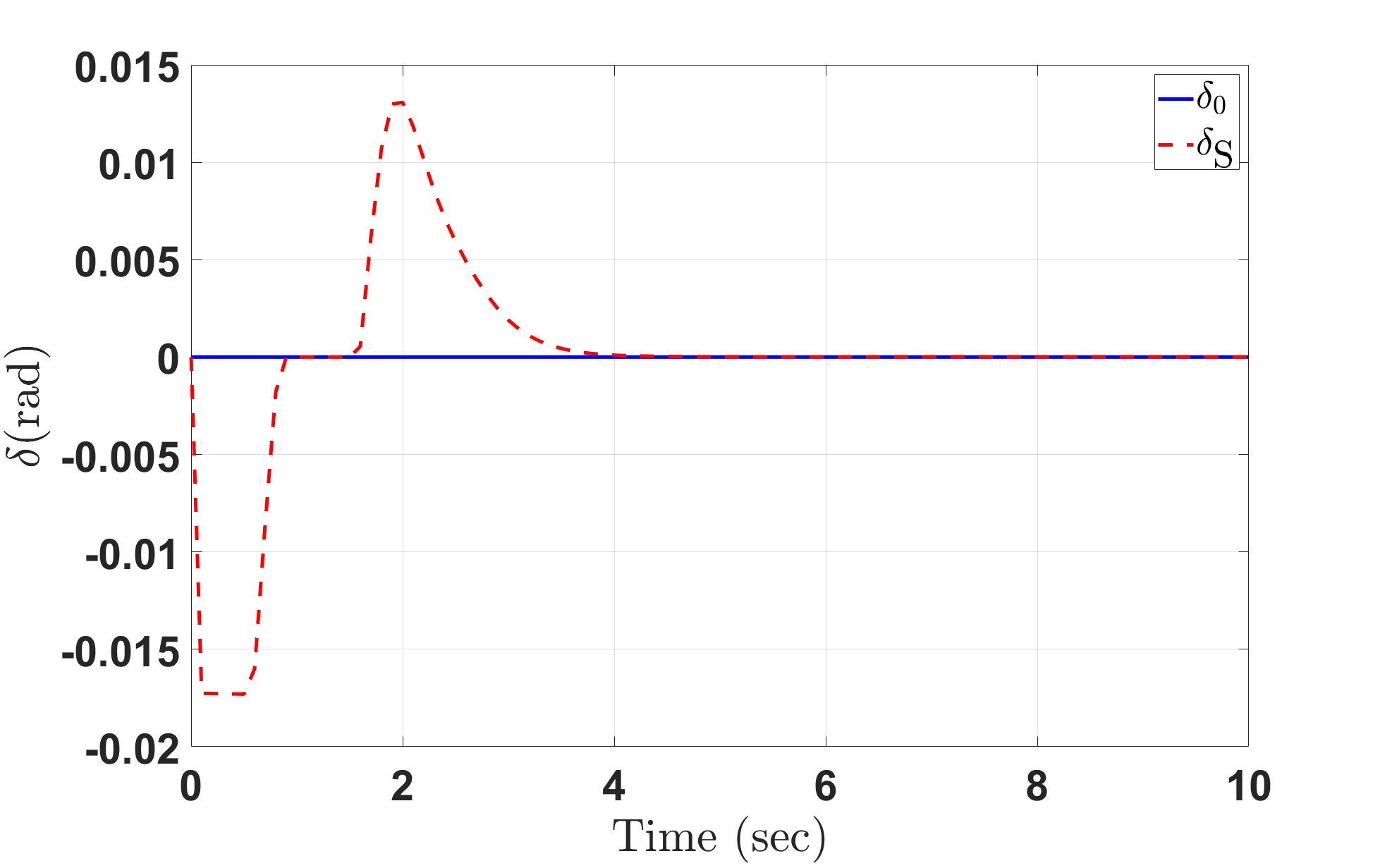}
  \caption{Front Wheel Angle.}
  \label{fig:ex3steer}
\end{subfigure}
\begin{subfigure}{0.5\textwidth}
  \centering
  \includegraphics[width=1\linewidth]{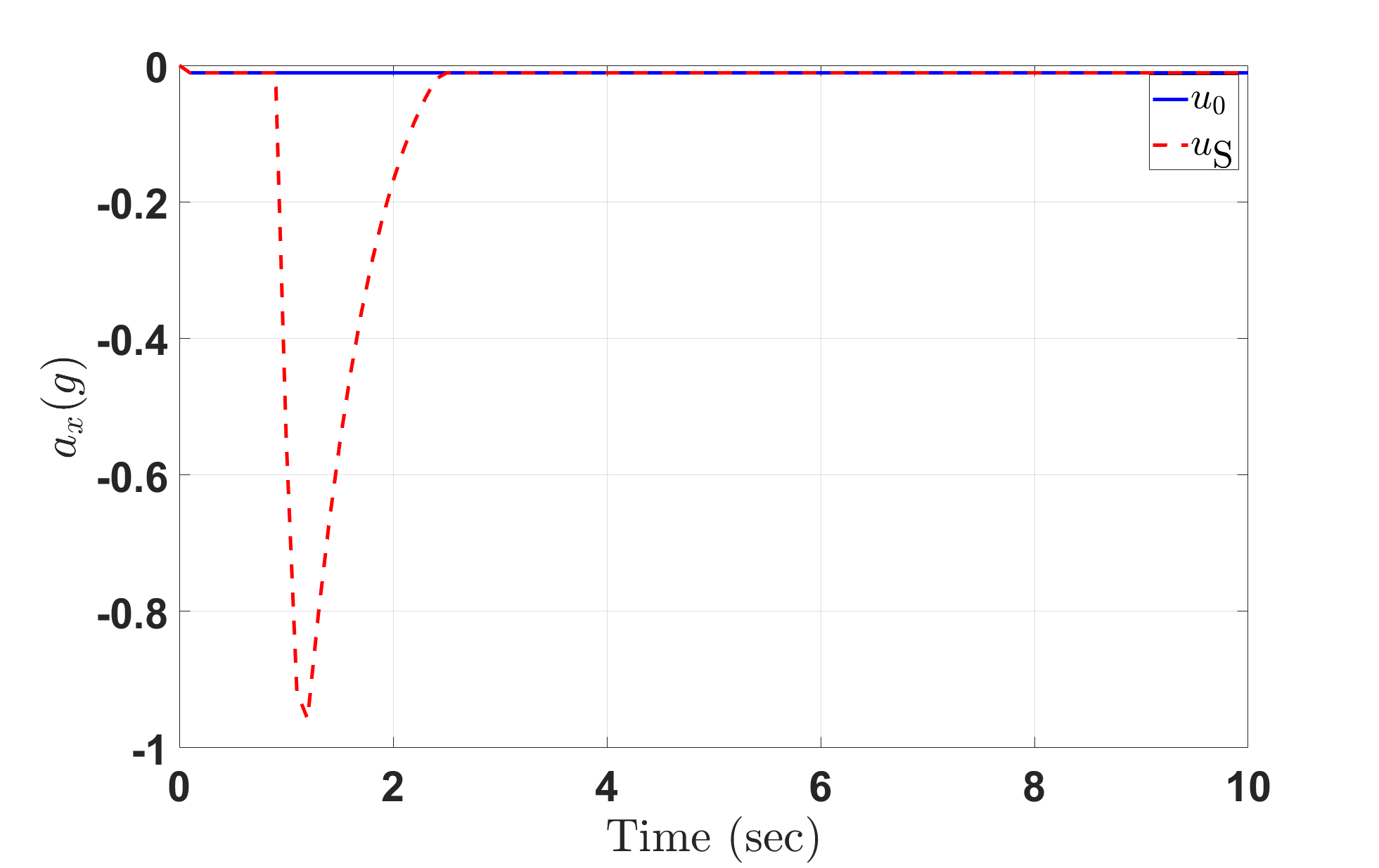}
  \caption{Longitudinal Acceleration.}
  \label{fig:ex3acc}
\end{subfigure}
\caption{Example \ref{ex3} -- Front Wheel Angleand Longitudinal Acceleration.}
\label{fig:ex3control}
\end{figure}

\begin{figure}[!ht]
\centering
\includegraphics[width = 0.60\linewidth]{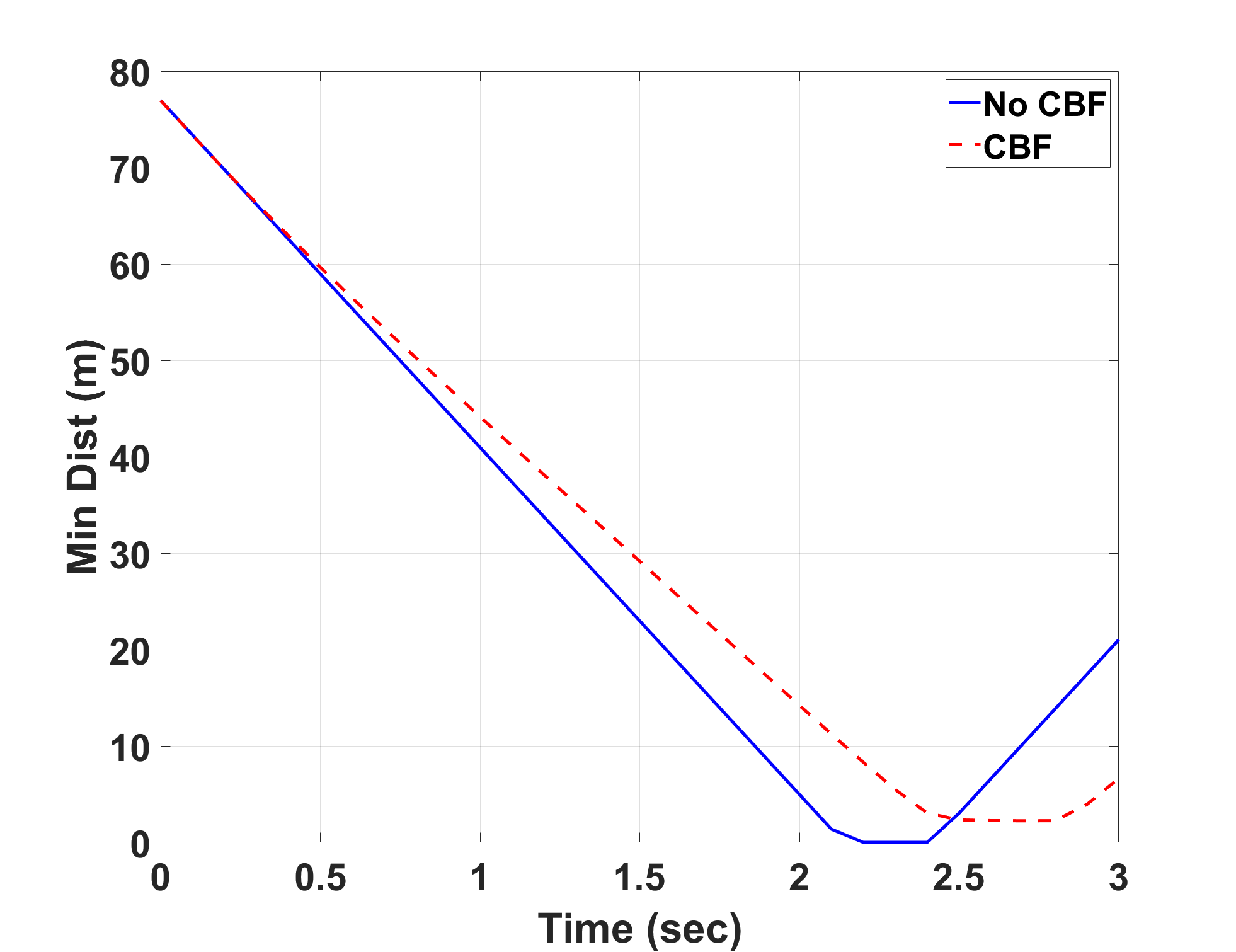}
\caption{Example \ref{ex3} -- Minimum Euclidean distance between ego and target vehicles.}
\label{fig:ex3mindist}
\end{figure}

In this example, the ego vehicle evades a potential collision with an obstacle that is stationary ahead. Fig. \ref{fig:ex3hist} shows the position of the ego vehicle and the target vehicles (including the stationary obstacle/vehicle on the left-most lane) at 0.5 sec intervals. Figures \ref{fig:ex3steer} and \ref{fig:ex3acc} show the front wheel angle and acceleration of the ego vehicle and Figure \ref{fig:ex3mindist} shows the minimum Euclidean distance between the ego and the critical target vehicle edges. The CBF safety filter changes lane to avoid the stationary car ahead, while slowing down to avoid colliding with the slow travelling lead car in the center lane. (Note that it further prevents erratic cut-in into the right most lane.)


\subsection{Summary}

In this section, we introduced decoupled longitudinal and lateral Control Barrier Functions for collision avoidance, and a safety filter that utilizes contextual knowledge of steering/braking efficacy in different situations to arbitrate between the two options to avoid collisions. We also present road keeping barrier functions that are used to prevent driving off the roadway when attempting to avoid collisions. The safety filter checks if the nominal steering and throttle/brake commands are safe by evaluating for CBF constraint violation. If the nominal control is safe, the safety filter does not modify them. If they are unsafe the algorithm overrides the nominal control with a minimally intrusive safe control that prevents the constraint violation. This safe control is calculated by solving constrained quadratic optimization problems for the front wheel angle and the longitudinal acceleration. Several examples demonstrating collision avoidance in common highway situations were presented including prevention of erratic cut in, evasion upon target cut in, and avoidance of stationary obstacles.

\section{Integrated Driving Policy with DRL, Motion Control, and CBF Safety Filter} \label{SEC:RL_CBF}
In this section we replace the rule-based safety filter from Section \ref{DRL+RB} with the CBF based collision avoidance algorithm from Section \ref{Rahman2021CSD}. Using the CBF safety filter has two primary advantages. First, it can be used to provide continuous feedback to the Agent, as opposed to binary safe/unsafe feedback by rule-based safety filter. This helps the Agent to assess the severity of its mistake and prevents it from repeating the same. Secondly, with a CBF safety filter the Agent can adapt to different environments after its initial training without endangering the occupants of the AV or nearby motorists.

\subsection{Training Architecture with CBF Safety Filter}
\begin{figure}[!ht]
\centering
\includegraphics[width = 1\linewidth]{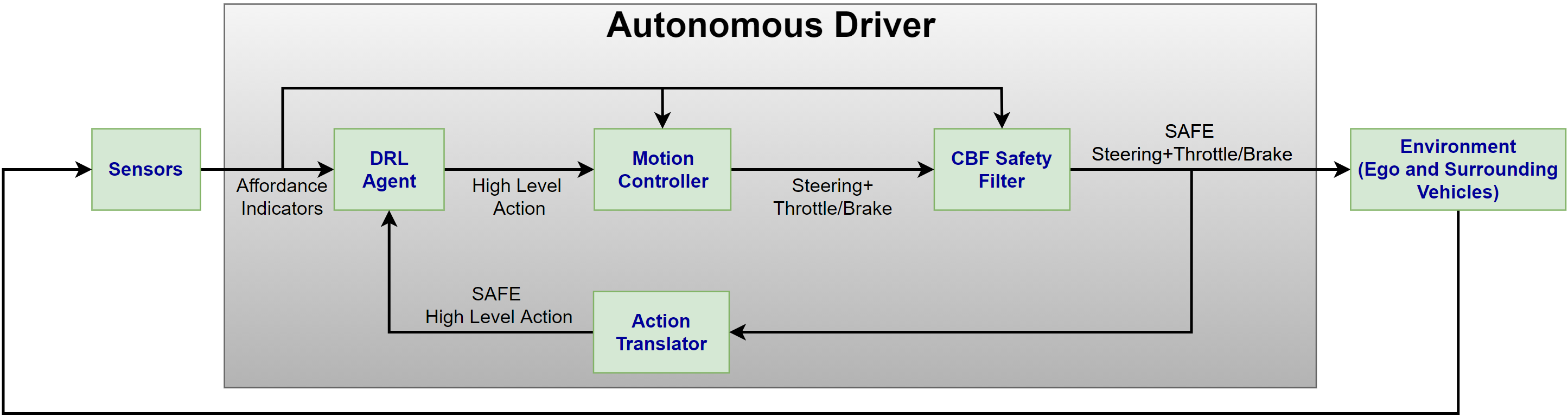}
\caption{Training Architecture.}
\label{fig:TrainingArch}
\end{figure}

The high level decisions made by the Agent are executed by a low level motion controller algorithm (Section \ref{sec:Motion Control}). These \textit{nominal} controls and the positions and velocities of surrounding vehicles are fed to the CBF based collision avoidance system. If the control action given the situation is deemed safe, it is passed through without modification. However, if the control action is unsafe, it is overridden with a safe control calculated by the CBF safety filter. The combination of the RL Agent and the CBF based collision avoidance system is referred to as the Autonomous Driver.
Fig. \ref{fig:TrainingArch} shows the system used for training the Agent. Similar to Section \ref{DRL+RB}, sensors provide affordance indicators to the Agent, which then makes a decision in the form of a high level action. The high level action is then translated to a front wheel angle, throttle and brake command by a motion controller algorithm. These low level commands and affordance indicators are then provided to the CBF Safety Filter. The safety filter computes the safe front wheel angle and throttle/brake commands. If the original path is safe, these commands are unchanged. The safe front wheel angle and throttle/brake commands are then applied to the ego vehicle. The Action Translator block translates the safe low level commands back into a safe high level action. This is used to provide feedback to the Agent about the action it has taken. If the safe action differs from the action selected by the Agent, it is penalized.
In this work, the Agent makes a decision every second, and the motion controller and CBF updates the front wheel angle and throttle/brake every 10\textsuperscript{th} of a second.
\subsubsection{Action Translation and Penalizing Unsafe Actions}
To provide effective feedback to the DRL Agent, the continuous safe front wheel angle and longitudinal acceleration signals must be translated back into the high level action space. For the longitudinal acceleration, this is straightforward, since both the high level action and the continuous signal have the same units (acceleration in $g$'s). For the front wheel angle $\delta$, we use a Barrier Function based intent prediction \citep{rahmanACC2021} to determine if the steering action is trying to prevent a lane change or triggering one.

\begin{figure}[!ht]
\begin{subfigure}{0.5\textwidth}
  \centering
  \includegraphics[width=1\linewidth]{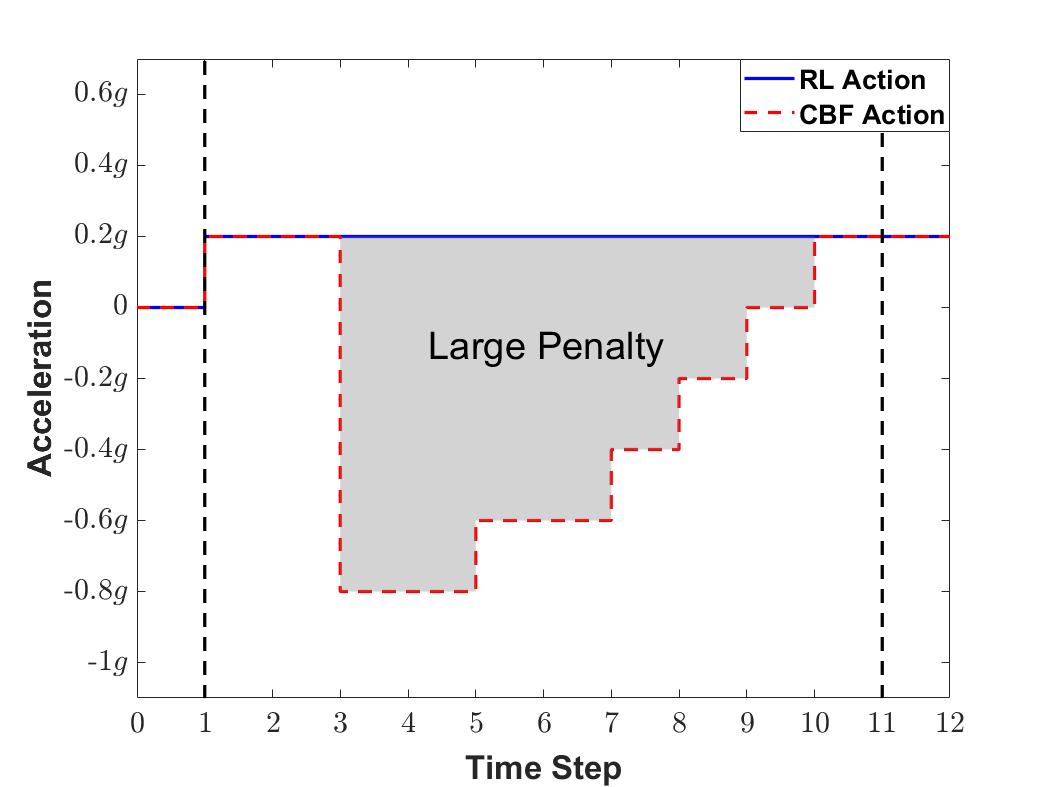}
  \caption{Example large longitudinal penalty.}
  \label{fig:lon_large}
\end{subfigure}
\begin{subfigure}{0.5\textwidth}
  \centering
  \includegraphics[width=1\linewidth]{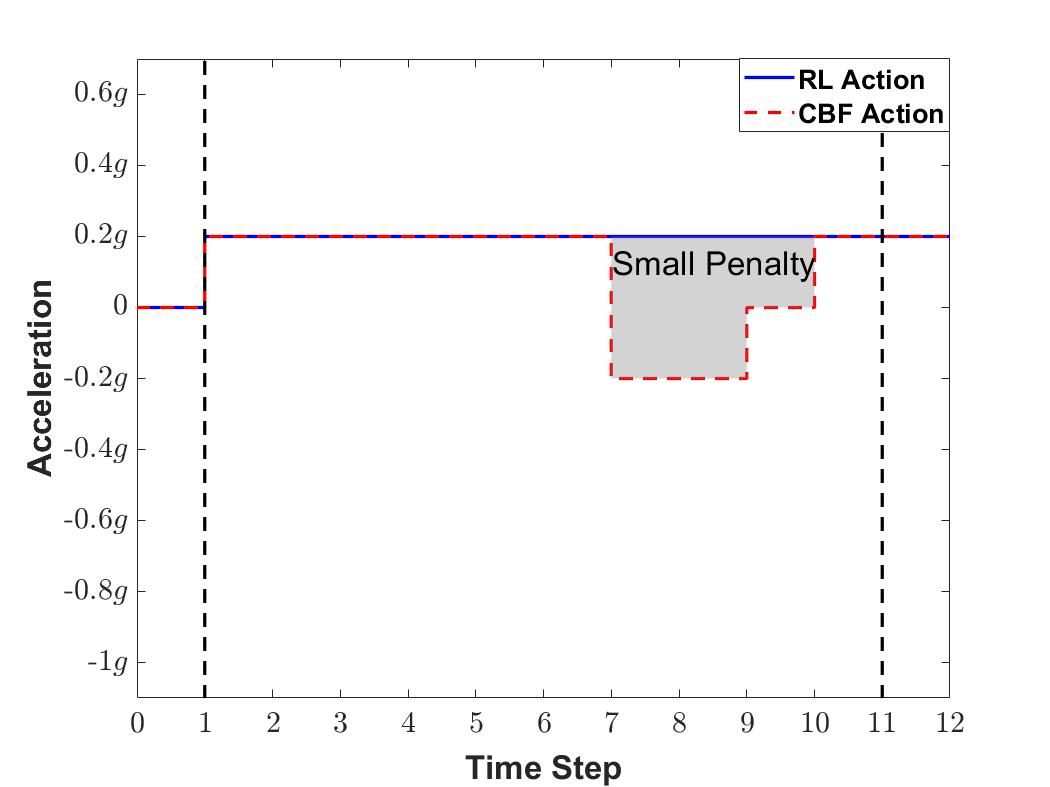}
  \caption{Example small longitudinal penalty.}
  \label{fig:lon_small}
\end{subfigure}
\caption{Determination of Penalty for Unsafe Longitudinal Actions.}
\label{fig:lon_penalty}
\end{figure}

\begin{figure}[!ht]
\begin{subfigure}{0.5\textwidth}
  \centering
  \includegraphics[width=1\linewidth]{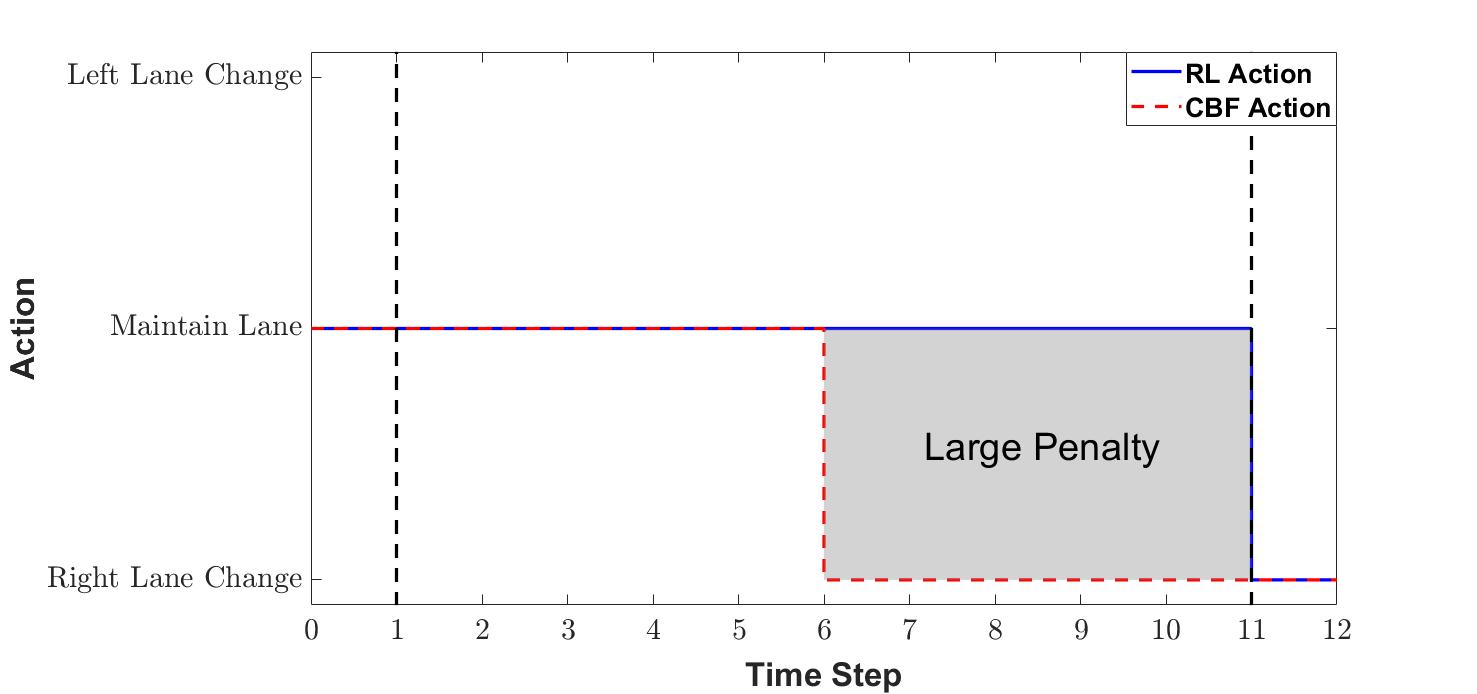}
  \caption{Example large lateral penalty.}
  \label{fig:lat_large}
\end{subfigure}
\begin{subfigure}{0.5\textwidth}
  \centering
  \includegraphics[width=1\linewidth]{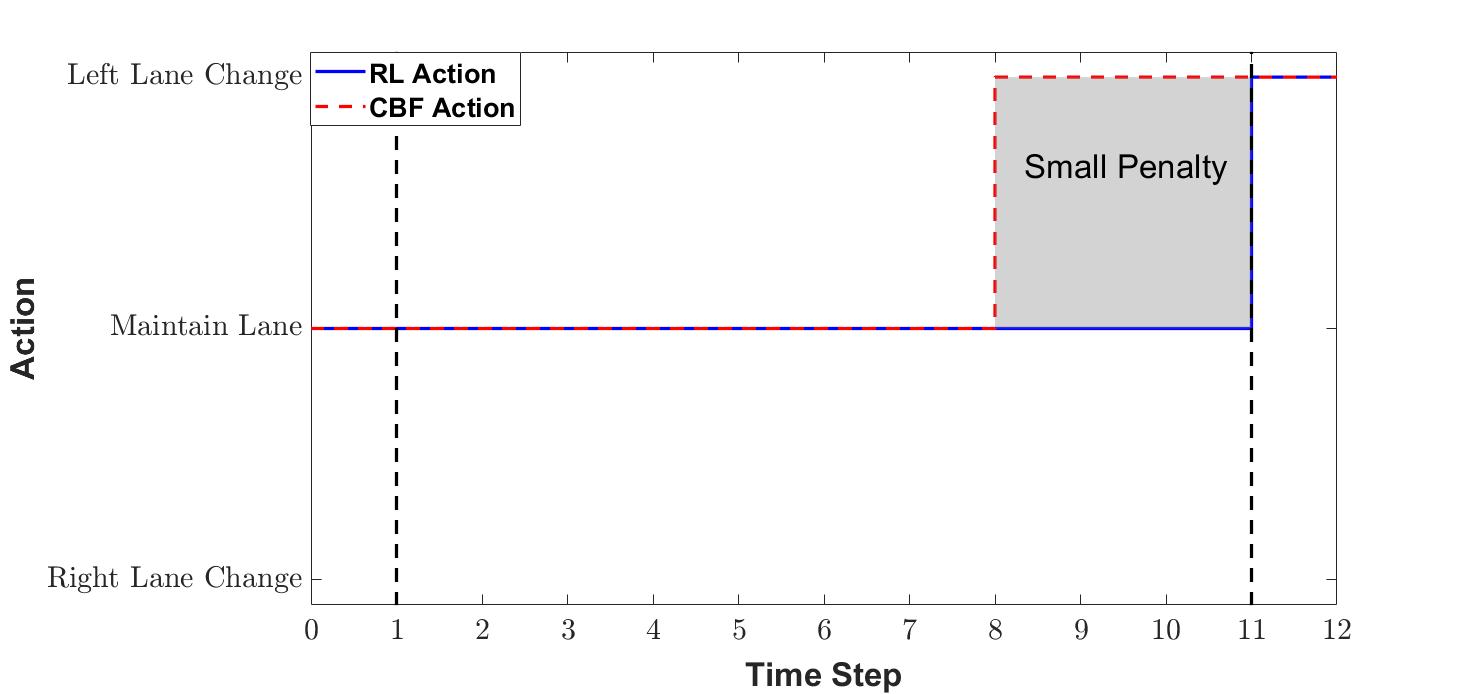}
  \caption{Example small lateral penalty.}
  \label{fig:lat_small}
\end{subfigure}
\caption{Determination of Penalty for Unsafe Lateral Actions.}
\label{fig:lat_penalty}
\end{figure}

We use the same reward function components (\ref{eq:rewv})-(\ref{eq:rewx}) as in Section \ref{DRL+RB}, with an additional component. The addition is a safety component that determines how safe the action is by comparing it to the safe action output by the CBF
\begin{align}
r_\rms = f_x(a_x,\bar{a}_x) + f_y(a_y,\bar{a}_y),
\end{align}
where $a_x$ is the longitudinal action selected by the Agent, $\bar{a}_x$ is the safe longitudinal action by the CBF filter, $a_y$ is the lateral action selected by the Agent, $\bar{a}_y$ is the safe lateral action by the CBF filter, and $f_x$ and $f_y$ are functions that determine the size of the penalty for unsafe longitudinal and lateral actions respectively. The size penalty on unsafe actions depends on how different the unsafe action is compared to the safe action, and also how long the decision chosen by the Agent was deemed unsafe. Figures \ref{fig:lon_penalty} and \ref{fig:lat_penalty} show qualitatively how the size of the penalty is determined. In Fig. \ref{fig:lon_large}, the RL action is to accelerate at 0.2$g$, but the safe (determined by CBF Safety filter) action is to brake hard for nearly the entire duration of the RL Action. Such a case incurs a large penalty. In Fig. \ref{fig:lon_small}, the RL action is again to accelerate at 0.2$g$. In this case, the safe action is to brake near the end of the RL Action, and so this decision incurs a smaller penalty.
Similarly, in Fig. \ref{fig:lat_large}, the RL action is to maintain the lane. However, at the 6$^{\rmt\rmh}$ time step, the safety filter must perform a left lane change in order to avoid a collision. The time duration at which RL action becomes unsafe determines how large the penalty is.

\subsubsection{Training Results}
\begin{figure}[!ht]
\centering
\includegraphics[width = 0.9\linewidth]{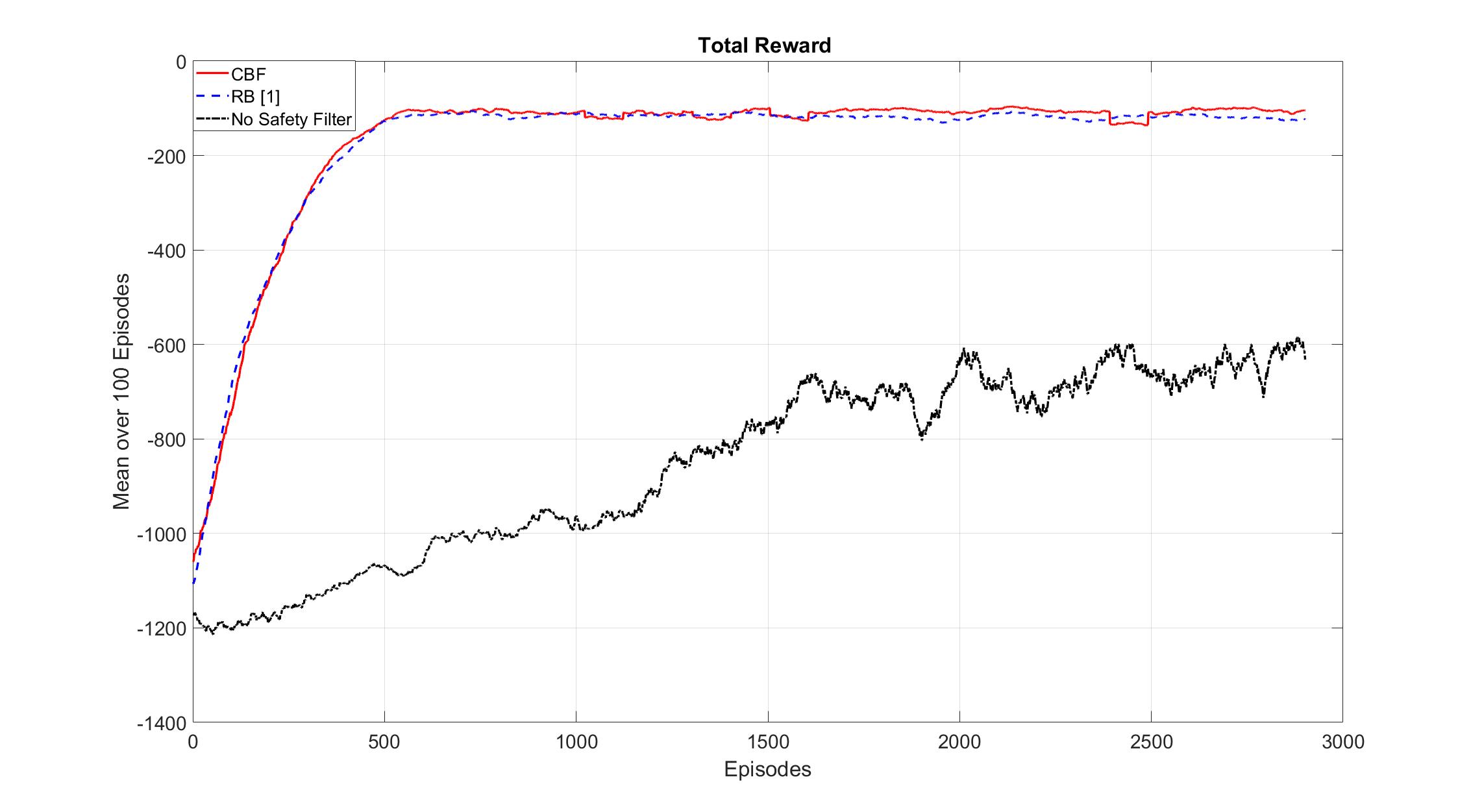}
\caption{Training with and without the CBF Filter.}
\label{fig:trainingCBF}
\end{figure}

To train the Agent, we use episodes consisting of 200 seconds of highway driving. In each episode, the surrounding environment, i.e. the density, speed, location of the traffic is randomized. Fig. \ref{fig:trainingCBF} shows the training performance with and without CBF, and also using only the rule based safety filter from Section \ref{DRL+RB}.
Whilst learning how to drive without any safety filter, the Agent has many collisions initially, and does not learn how to drive safely. With the CBF, the time taken to learn to drive acceptable driving behavior is reduced significantly, the driving behavior is better as shown by the higher reward, and the driving behavior is safer, since CBF prevents collisions in case the Agent makes an unsafe decision. Without a safety filter, it is difficult to structure the reward function in a way that guides the Agent to drive safely and in a manner that is not exceedingly conservative.

\begin{figure}[!ht]
\centering
\includegraphics[width = 0.8\linewidth]{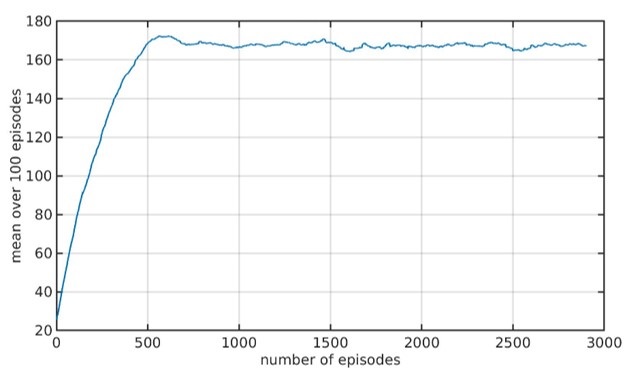}
\caption{Number of Safe Agent Actions in Each Episode.}
\label{fig:CBFIntervention}
\end{figure}

Fig. \ref{fig:CBFIntervention} shows the mean over 100 episodes of the number of safe Agent actions in each episode. In each episode, 20 actions are random exploration, so the maximum number of safe actions selected by the Agent is 180. The also shows that the Agent learns to drive safer as time progresses.

\begin{figure}[!ht]
\centering
 \includegraphics[width=0.9\linewidth]{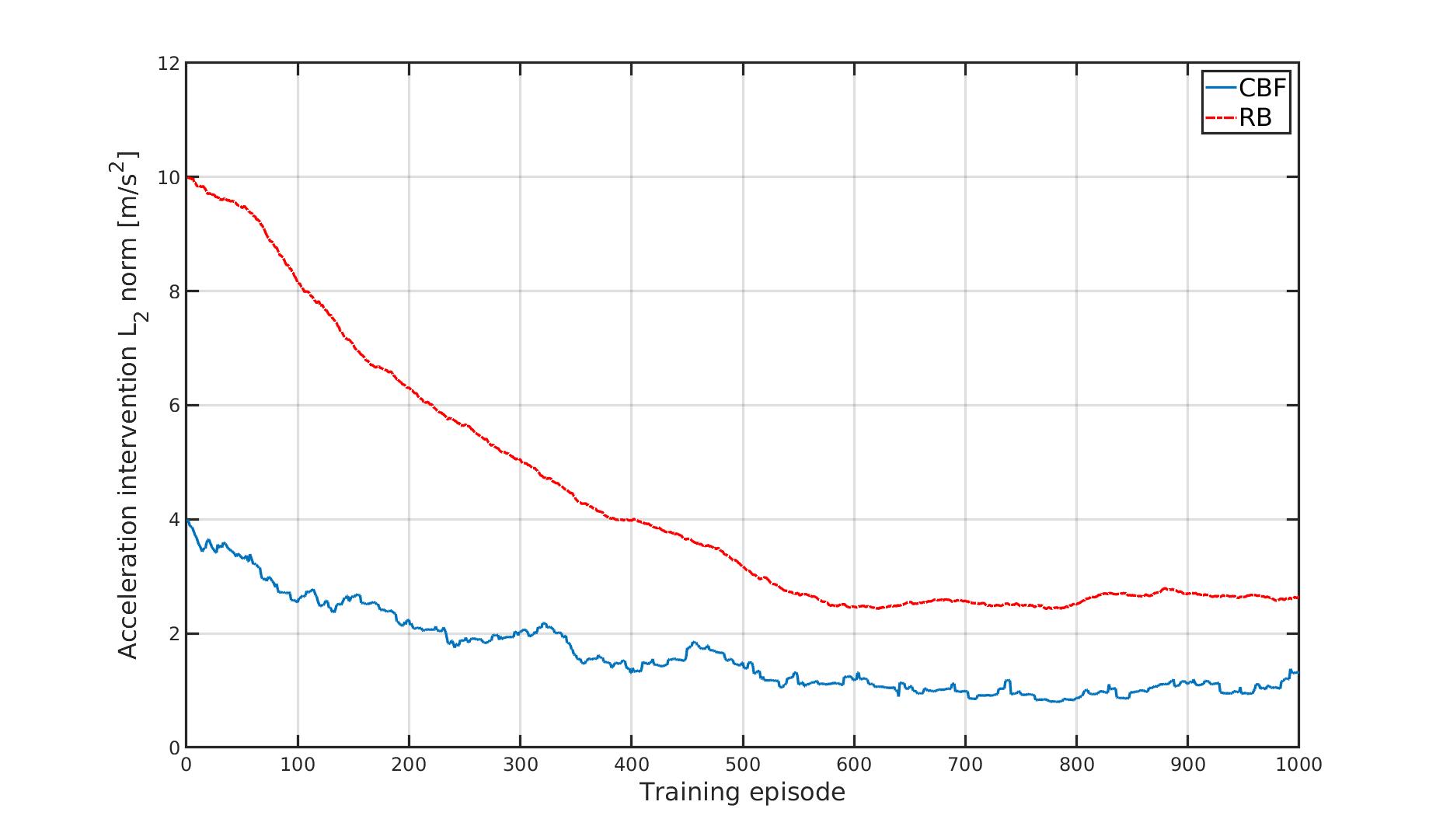}
\caption{Comparison of CBF  Longitudinal Intervention with RB Safety Filter.}
\label{fig:CBFIntervention1}
\end{figure}

\begin{figure}[!ht]
\centering
\includegraphics[width=0.9\linewidth]{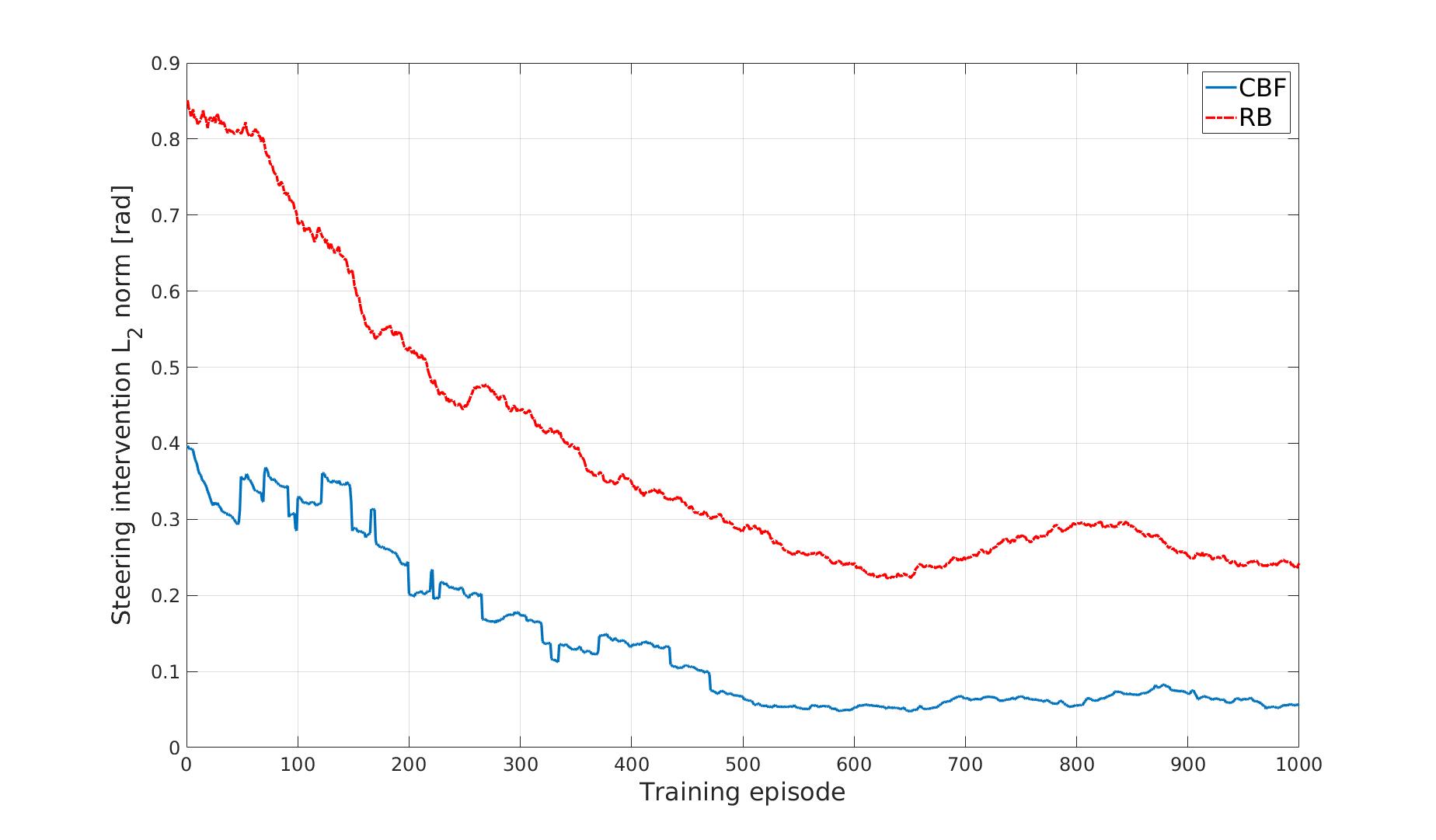}
\caption{Comparison of CBF Lateral Intervention with RB Safety Filter.}
\label{fig:CBFIntervention2}
\end{figure}

\paragraph{Comparison of CBF Safety Filter Intervention with Rule-Based Safety Filter:}
Figures \ref{fig:CBFIntervention1} and \ref{fig:CBFIntervention2} show the longitudinal and lateral interventions of the CBF and RB safety filters as the agent is trained. For both safety filters, the number of interventions and the severity of those interventions decrease over time, implying that the Agent is learning to be a safer driver, and since the reward is higher with the safety filter, this implies that the addition of the safety filter does not cause the Agent to become too conservative. The plots show that the longitudinal safety filter is more intrusive with its intervention, but this does not have an adverse effect on the reward. An advantage of using the CBF safety filter over the rule-based safety filter from Section \ref{DRL+RB} is that the CBF can be tuned to a desired degree of conservative behavior, whereas the rule-based safety filter can only choose between several predefined actions that may be either insufficient or unnecessarily excessive, which can reduce passenger confidence and also result in an uncomfortable ride.

\begin{figure}[!ht]
\begin{subfigure}{1\textwidth}
  \centering
  \includegraphics[width=0.9\linewidth]{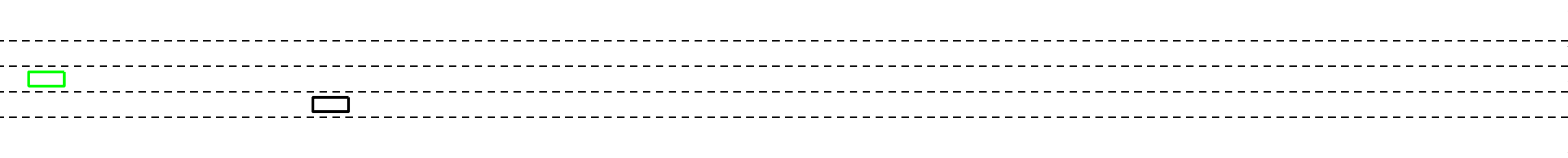}
  \caption{$t = 0$ sec}
\end{subfigure}
\begin{subfigure}{1\textwidth}
  \centering
  \includegraphics[width=0.9\linewidth]{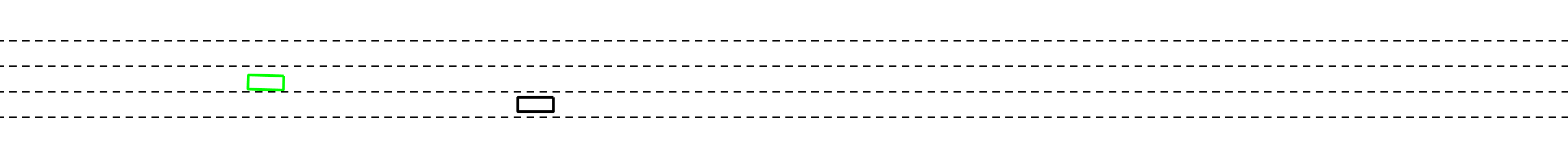}
  \caption{$t = 1$ sec}
\end{subfigure}
\begin{subfigure}{1\textwidth}
  \centering
  \includegraphics[width=0.9\linewidth]{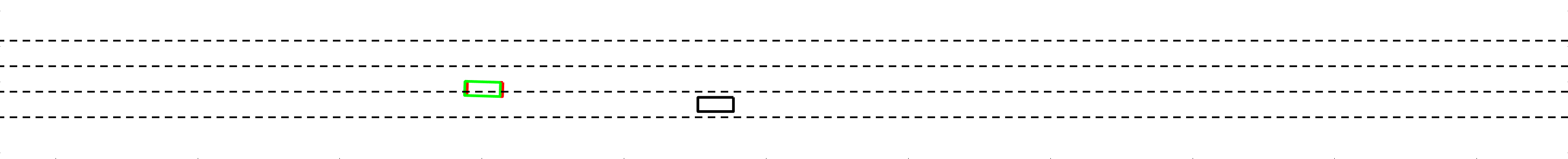}
  \caption{$t = 2$ sec}
\end{subfigure}
\begin{subfigure}{1\textwidth}
  \centering
  \includegraphics[width=0.9\linewidth]{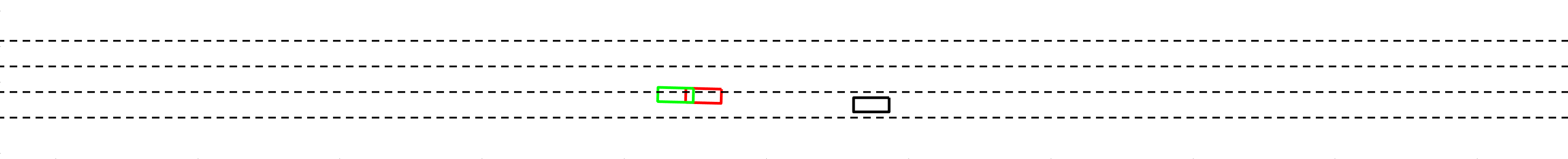}
  \caption{$t = 3$ sec}
\end{subfigure}
\begin{subfigure}{1\textwidth}
  \centering
  \includegraphics[width=0.9\linewidth]{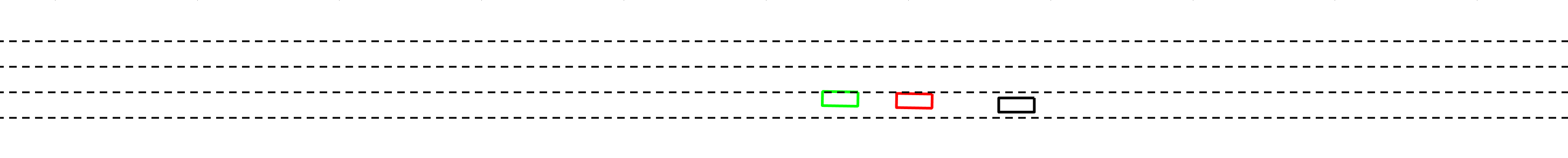}
  \caption{$t = 4$ sec}
\end{subfigure}
\begin{subfigure}{1\textwidth}
  \centering
  \includegraphics[width=0.9\linewidth]{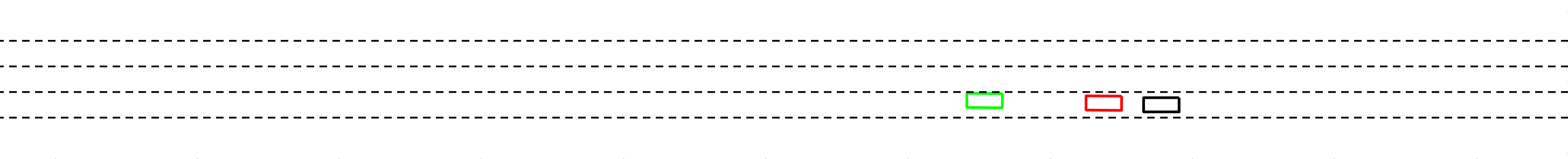}
  \caption{$t = 5$ sec}
\end{subfigure}
\begin{subfigure}{1\textwidth}
  \centering
  \includegraphics[width=0.9\linewidth]{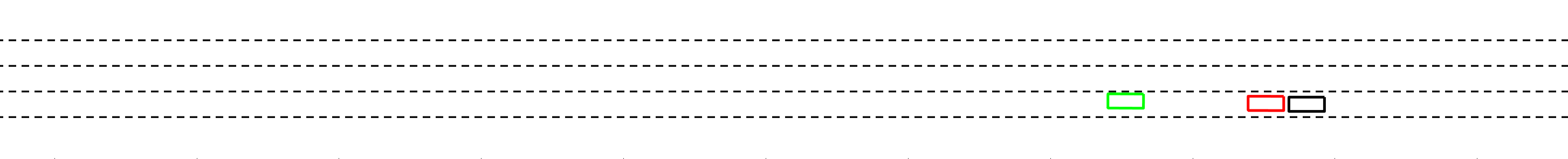}
  \caption{$t = 6$ sec}
\end{subfigure}
\begin{subfigure}{1\textwidth}
  \centering
  \includegraphics[width=0.9\linewidth]{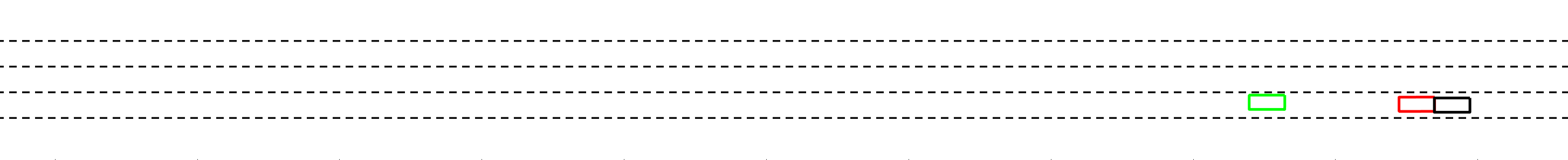}
  \caption{$t = 7$ sec}
\end{subfigure}
\caption{Position of the ego vehicle and the target vehicles at 1 sec intervals.}
\label{fig:cbf_v_rb}
\end{figure}

\paragraph{Collision avoidance with CBF Safety Filter Intervention with Rule-Based Safety Filter} \label{ex4}
In this example, we show an use case where the intervention by the RB safety filter is insufficient. The AV and a target vehicle are driving at 70 mph. The RL policy initiates a lane change at $t=0$ sec. At $t = 2$ sec, the black target vehicle decelerates at $0.35g$ for 3 seconds. Fig. \ref{fig:cbf_v_rb} shows the response of the RB safety filter and the CBF Safety Filter. The green vehicle is the AV with the CBF safety filter and the red vehicle is the AV with the RB safety filter. The RB safety filter reacts later to the decelerating target vehicle and is unable to avoid a collision.

\paragraph{Retraining:}
One of the key advantages of the proposed approach is that it can allow an already trained Agent to continue to adapt its driving policy once on the road. For example, acceptable driving behavior in different areas of a country are often slightly different. An Agent can potentially adapt in the field and optimize to its local conditions by retraining itself. The actual adaptation algorithm is a subject of ongoing research and is not discussed in this paper. However, without a safety filter, an Agent making exploratory decisions may get itself into an unsafe situation. The benefit of the safety filter is twofold. If the Agent makes an unsafe decision during retraining, the decision is overridden and the safety of the AV is not compromised. Secondly, the Agent is provided immediate feedback that the decision was unsafe, therefore making it unlikely to make the same mistake again.
\subsection{Summary}

In this section we replaced the rule-based safety filter from Section \ref{DRL+RB} with the CBF based safety filter from Section \ref{Rahman2021CSD}. The amount of time taken for the Agent to learn a safe driving behavior is significantly reduced thanks to the efficient use of the Safety Filter. It overrides the agent's unsafe action and provides a safe alternative which acts as an additional feedback to the agent. To this end we modified the reward function from Section~\ref{DRL+RB}  to include an additional penalty based on the difference between the agent’s nominal action and the safe action by the CBF. Training results show that the agent learns to drive safely and that the number of unsafe decisions reduces with training. The severity of the safety filter interventions also decrease as the training proceeds. The addition of the CBF safety filter helps to make the Autonomous Driver more robust to unsafe high level decisions and to aggressive traffic vehicles. In addition, the Autonomous Driver can adapt to new environments by learning online without forgoing on safety and comfort.

\section{Summary and Conclusion}
In this chapter we provide a practical approach to the design of RL-based driving policy for highway autonomous driving. Proposed driving policy system integrates a high level DRL-driven decision-making module with a path planning and path following strategy transforming the discrete DRL action space into a sequence of motion control commands.  The benefits of the hierarchical decoupling of the RL decision logic from the algorithms for path formation and execution are the simplification of the RL algorithm design and training, and the opportunity to use versatile motion control algorithms. The latter facilitates the incorporation of engineering knowledge and experience in designing smooth human-like driving and safe lane change maneuvering.  The engineers’ field  experience and knowledge includes the selection of lead target vehicle for longitudinal speed profile, the calibration of lateral acceleration profile and corresponding feedback control gains, the compliance of actuation constraints during lane change maneuvers, and the time required to perform a lane change. The chapter elaborates on the practical aspects of the design of the DRL decision-making strategy and motion control as critical components for real-world implementation of AI based decision policy for autonomous driving.

The chapter further focuses on addressing the overall robustness and safety of the output produced by the decision-making and motion control layers of the driving policy. We discuss the concept of a safety filter as an important means in autonomous driving applications of RL.  The safety filter overrides the nominal front wheel angle and throttle/brake commands in case of imminent/unexpected threats while allowing the RL optimizer to focus on longer term goal. Two alternative safety filters defining the safety boundary of the produced control output are discussed.  The first one uses handcrafted rules to constrain the DRL/motion control output within a predefined safety boundary. It accepts or rejects the RL decisions by enforcing traffic regulations and handcrafted rules for time headway and gap size requirement. The second solution introduces a CBF-based filter that provides a dynamic safety envelope safeguarding the control output along the vehicle trajectory. It adjusts the control actuation corresponding to RL decisions by leveraging the concept of invariant set to assess imminent threats and calculate minimal required adjustment (in steering/braking) for collision avoidance. The design and vehicle implementation of CBFs as a tool for collision avoidance and following road geometry constraints are discussed. The chapter concludes with a discussion on the pros and cons of both safety filters and their impact on the training and overall performance of the DRL algorithm, and the open opportunity of DRL – integrated with motion control and a safety filter – to adapt to new environments by learning online with corresponding safety and comfort assurance.

\bibliographystyle{spbasic}
\bibliography{bib}

\end{document}